\documentclass[a4paper,12pt]{report}
\pdfoutput=1
\usepackage{setspace}
\usepackage{graphics}
\usepackage{thesis}
\usepackage{latexsym}
\usepackage[dvips]{epsfig}
\usepackage{graphicx}
\usepackage{amsmath}
\usepackage{amsthm} 
\usepackage{amssymb}  
\usepackage{amsfonts}  
\usepackage{mathrsfs}
\usepackage{mathtools}
\usepackage{txfonts}
\usepackage{autobreak} 
\usepackage{subfigure} 
\usepackage{bm}
\usepackage{enumitem}
\newtheorem{coro}{Corollary}
\newtheorem{theorem}{Theorem}
\newtheorem{property}{Property}
\newtheorem{defn}{Definition}
\usepackage[bottom]{footmisc}
\usepackage[linesnumbered,ruled]{algorithm2e} 
\usepackage{microtype}
\usepackage{multirow}
\usepackage{hyperref}
\newcommand\sbullet[1][.5]{\mathbin{\vcenter{\hbox{\scalebox{#1}{$\bullet$}}}}}
\usepackage{CJKutf8}
\usepackage{color,soul}

\newcommand*{\blankpage}{%
\newpage
\vspace*{\fill}
{\centering (This page is intentionally left blank.)\par}
\vspace{\fill}
\thispagestyle{plain} 
\mbox{}
}

\begin{document}

\thesistitle{Ray-based Interference Free Workspace Analysis and Path Planning for Cable-Driven Robots}
\authorname{ZHANG, Zeqing}
\degree{Master of Philosophy}
\programme{Mechanical and Automation Engineering}
\supervisor{}
\submitdate{September 2019}

\coverpage


\pagenumbering{roman}


    \addcontentsline{toc}{chapter}{Abstract}
	\vspace*{2cm}
	\large \noindent
	Abstract of thesis entitled: \\
	\indent \thesistitle \\
	Submitted by \authorname \\
	for the degree of \degree \\
	at \institution~in \submitdate

	\vskip 1cm \noindent
	\pdfoutput=1

\doublespacing
This thesis studies the interference free workspace (IFW) of arbitrary cable-driven robots (CDRs) for both cables and obstacles using the ray-based method.
Continuing from this, the point to point path planning and verification technique to ensure that the resulting motion is within the workspace is also studied.

CDRs are a type of parallel mechanism where cables, actuated by motorized winches, are connected from a base platform to either a single end-effector (also called cable-driven parallel robots, or CDPRs) or multiple links that are connected in series through passive joints (also called multi-link cable-driven Robots or MCDRs). CDRs have attracted much attention over years due to their promising advantages, such as high payload capacity, large potential workspace, and ease of reconfiguration. 

However, since utilizing cables as actuators, the collision between cables and environment would affect the workspace of CDRs. Therefore, the interference free workspace (IFW) between cable themselves and with environmental objects for CDRs must be well determined.
Existing IFW studies lack a generalized method to determine the IFW of arbitrary CDRs with high degrees of freedom (DoFs) and few works can handle different shaped obstacles. The proposed thesis tackle with these problems using a ray-based approach to transform the cable interference conditions into a set of univariate polynomial equations. In addition, the obstacles approximated by numerous triangles or composed of primitive geometric elements: segments, points and cones can be treated by the ray-based method to detect the potential interference between cables.
 
Based on the connectivity information from the ray based work-space, a novel path planning and verification approach between two given points is proposed. This work expands the ray-based method from a lattice grid to generate a polynomial curve and plan paths for CDRs on translation and orientation spaces simultaneously. Using the generated  path, the verification procedure can validate the feasibility of planned paths. 

The proposed thesis contributes to the workspace analysis and path study on planning and verification for CDRs. It can be used for multiple applications, such as the design of geometry of CDRs to satisfy the specific working area, comparison of CDRs depending on the working performance and path determination in the obstacle-strewn environment. 

	\newpage

\blankpage
\newpage








\tableofcontents

\listoffigures

\blankpage

\listoftables

\blankpage


\newpage
\pagenumbering{arabic}

\pagestyle{headings}

\doublespacing


\pdfoutput=1

\chapter{Introduction}


\section{Motivation}
The cable-driven robots (CDRs) are a subclass of parallel robots where cables, actuated by motorized winches, are connected from a base platform to either a single end-effector (cable-driven parallel robots or CDPRs) or multiple links that are connected in series through passive joints (multi-link cable-driven Robots or MCDRs).
In the last few decades, the studies on CDRs have been conducted for a large variety of applications. For example, 
building construction of brick structures \cite{ifw_cuhk}, large-scale laser cutting \cite{crolla2018inflatable} and large translational motion simulator \cite{miermeister2016cablerobot}. 

One fundamentally important problem related to applications of CDRs is the \textit{workspace determination}, which needs to determine the set of poses (both position and orientation) under different working requirements, such as wrench-feasibility, wrench-closure and interference-free conditions. In practice, it is obvious that the collision of cable-cable and cable-obstacle may affect the valid working area of CDRs. Thus, many papers focus on the interference free workspace (IFW) determination \cite{ifw_ptwise,ifw_iabm,ifw_jp,ifw-pt-opti,random,ifw_app2,ifw_app,ifw_Go}. However, three challenges remain in the study IFW analysis of CDRs.

First, the continuity of workspace is useful, since most robots should work in several continuous areas. One widely-used numerical method, that is the point-wise approach, will check the interference free condition at finitely many points, which lacks of continuity of the workspace.

In addition, the efficiency is a significant issue for IFW determination, especially for on-line applications. For point-wise method, the finer discretization the more accuracy about the resulting workspace. However, this would significantly increase the computation time required.

Finally, a general method to IFW determination for any CDRs is needed. Some existing methods to IFW analysis are analytical approaches, determining the closed-form boundaries of the IFW and hence the workspace region \cite{ifw_jp, ifw_Go}. Although it guarantees the workspace continuity, analytical techniques have only been used to investigate the IFW of particular types of CDRs, such as single link CDPRs. 

In summary, a general and efficient method to continuous IFW determination for any CDRs is desired.

In addition, finding available trajectories for CDRs is important since most practical tasks are defined as different trajectories in aforementioned applications. It contains two problems, that is \textit{path planning} and \textit{path verification}, respectively. 

The \textit{path planning} aims to generate a sequence of feasible poses that satisfy one or more workspace conditions, such as wrench closure \cite{pathplanning2005} and cable interference-free \cite{opt}. On the other hand, 
\textit{path verification} problem is to check whether the path connecting two given poses always satisfies particular workspace conditions or other constraints.

Therefore, the approach to finding a valid path under certain workspace conditions based on the continuity information from ray-based workspace is needed. Moreover, the verification technique to check the feasibility of the proposed paths is also desired.

\section{Research Questions}
This thesis is mainly trying to address the following questions:
\begin{itemize}
    \item A general method to determine the interference free workspace including cable-cable and cable-obstacle for any configured CDRs (including CDPRs and MCDRs).
    \item How to generate a path for CDRs satisfying certain conditions such as interference free conditions, etc.
    \item Can we guarantee the feasibility of the given path? If not, can we point out which parts along the given path are feasible or infeasible?
\end{itemize}

\section{Contributions}
In this thesis, the method of the interference free workspace (IFW) analysis for any configured CDRs is proposed by means of the ray-based method. It is shown that the proposed methodology has efficient computational advantage compared withe the point-wise evaluation. Aside from that, interference between different shaped objects and cable segments can be detected by the proposed method. The objects could be approximated as polyhedra by a set of triangles, or modeled as the combination of cylinders, ellipsoids, spheres and cones.

Furthermore, based on the benefits of the ray-based method, the continuity information of the interested workspace can be obtained, which will be used for the path planning. Based on the ray-based workspace, one novel point to point path planning method is proposed. Here paths on the translation and orientation space can be generated simultaneously, followed by a verification technique utilized for checking the feasibility of the obtained paths under given workspace conditions.

\section{Structure of the Thesis}
This thesis includes 5 chapters and the remainder are organized as follows.

First of all, Chapter 2 reviews the background and some relevant works that are related to the contributions of the thesis.

Then, Chapter 3 presents a general and efficient approach to determine the continuous IFW for any CDRs. The interference conditions are formulated and the interference cases between cables and different types of obstacles are considered. In addition, a generalized method to determine the coefficients of interference conditions is reported. Finally, the time costs between the point-wise method and the proposed method are compared as well.

In addition, Chapter 4 states a novel method to path planning on the basis of ray-based workspace. Furthermore, the ray-based path verification is proposed to figure out the feasibility of given paths under interference free conditions. It is shown that the valid and invalid parts along the given paths can be exactly determined, which provides an alternative way to representation the workspace between start and end points.

Finally, Chapter 5 concludes the thesis and presents future works.

\section{Related Publications}
The following publication is related to this thesis.
\begin{itemize}
    \item Zeqing Zhang, Hung Hon Cheng and Darwin Lau*, Efficient Wrench-Closure and Interference Free Conditions Verification for Cable-Driven Parallel Robot Trajectories Using Ray-Based Method, IEEE Robot. Auto. Let. (RA-L) (under review)
\end{itemize}

\chapterend
\pdfoutput=1
\chapter{Literature Review of CDRs}


\section{History and Background}\label{ssec:history}
\begin{figure}[htbp]
	\centering
	\subfigure[High complexity KUKA TITAN 1000 Robot]{
		\label{ffig:kuka}
		\includegraphics[width=0.35 \textwidth]{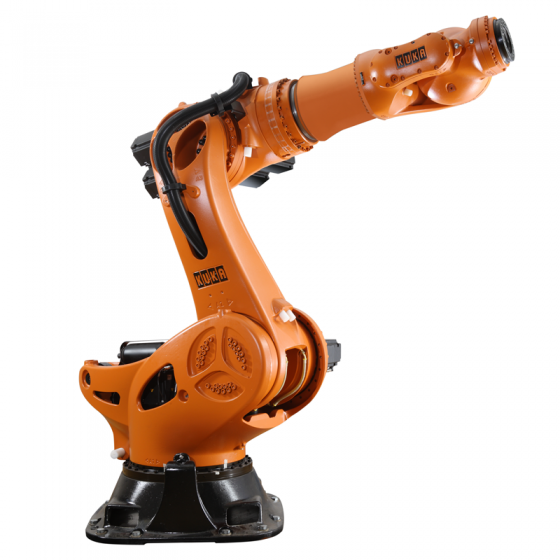}}~\hfil
	\subfigure[Universal Robots UR3]{
		\label{ffig:ur3}
		\includegraphics[width=0.35\textwidth]{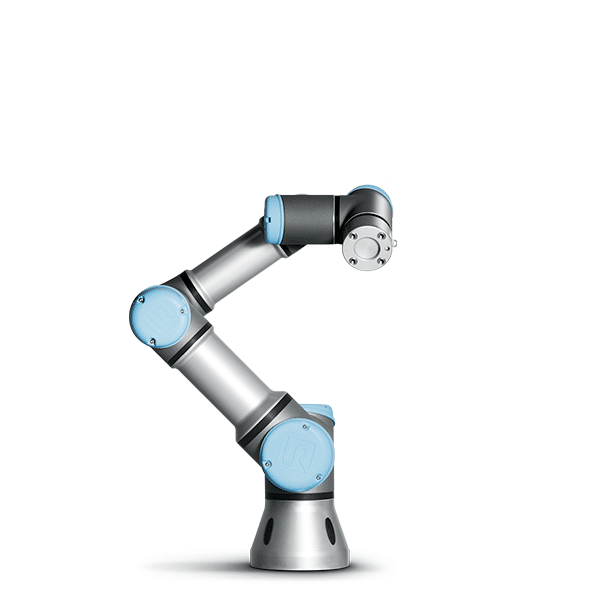}}~\vfil
	\subfigure[Lufthansa using Stewart platform]{
		\label{ffig:stewart}
		\includegraphics[width=0.35\textwidth]{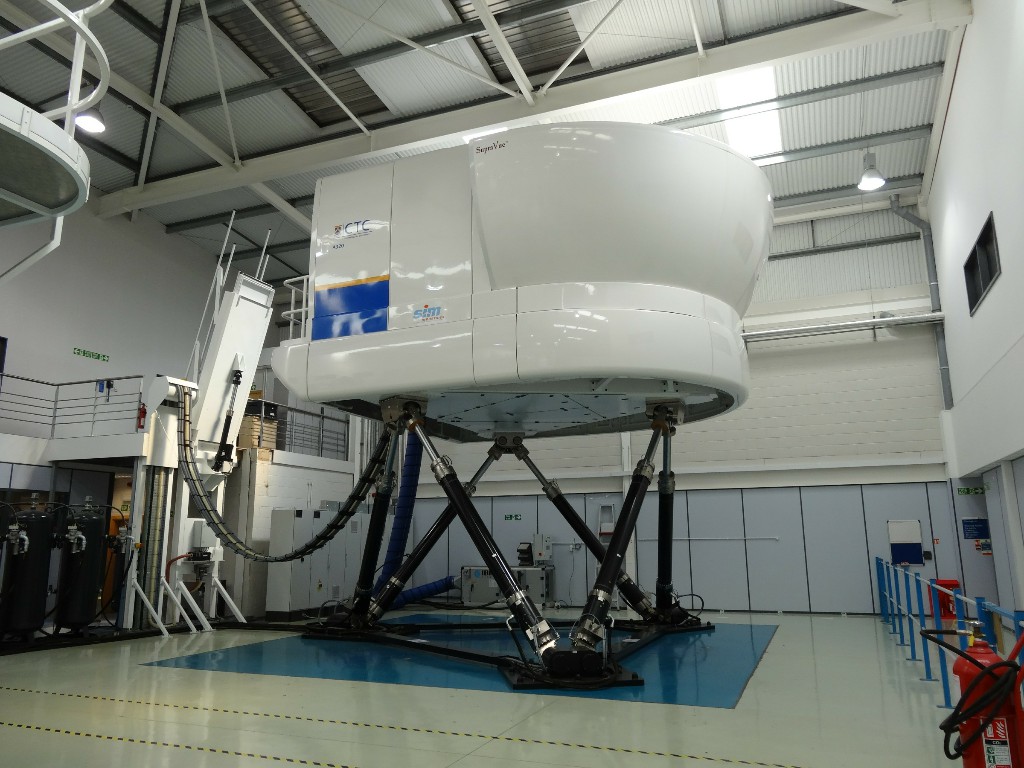}}~\hfil
	\subfigure[ABB IRB 340 Flexpicker]{
		\label{ffig:delta}
		\includegraphics[width=0.35\textwidth]{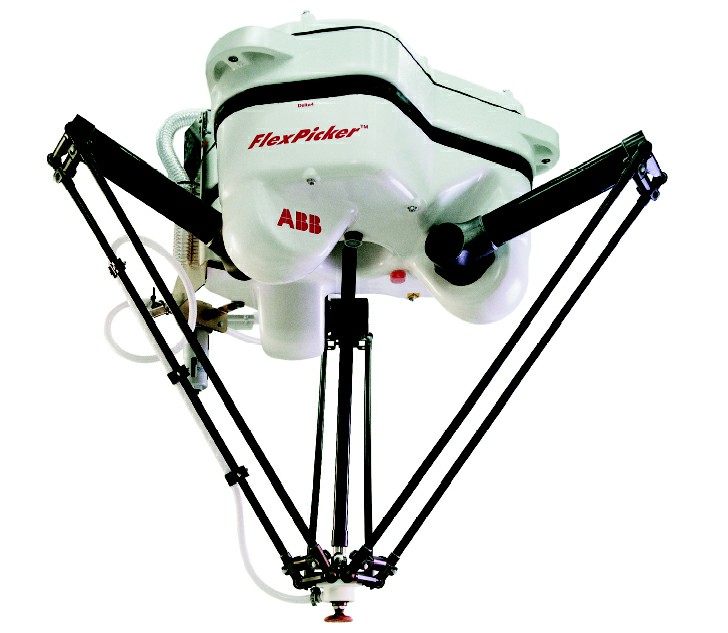}}~\hfil
	\caption{Robots}
	\label{fig:robot}
\end{figure}

Robots are designed for helping people to accomplish desired tasks with low human power and more efficiency. In general, there are many ways to classify robots, such as depending on the functions, applications, forms, etc. Based on the form of robots, they can be categorized into {\it serial robots} and {\it parallel robots}. Here the main distinction between them, which can be analogy to that between the serial circuit and parallel circuit in electrical networks, lies in their topology rather than their geometry \cite{pott2018}.
As shown in Fig.~\ref{fig:robot}, some robots are demonstrated, including serial robots (Fig.~\ref{ffig:kuka} and Fig.~\ref{ffig:ur3}) and parallel robots (Fig.~\ref{ffig:stewart} and Fig.~\ref{ffig:delta}).

Among parallel robots, one unique subclass was proposed in 1980s, that is {\it cable-driven parallel robot} or CDPR in short.
In 1984, Landsberger first came up with the design of a cable-controlled parallel manipulator in his master thesis \cite{land1984}, which is generally accepted as the first prototype of CDPRs. Later on, Higuchi \cite{higuchi1988} implemented the multi-cable cranes for construction purpose in 1988.

\begin{figure}[htbp]
	\centering
	\subfigure[Cable-driven parallel robot \cite{pott2015cdpr}]{
		\label{ffig:realcdpr}
		\includegraphics[width=0.4 \textwidth]{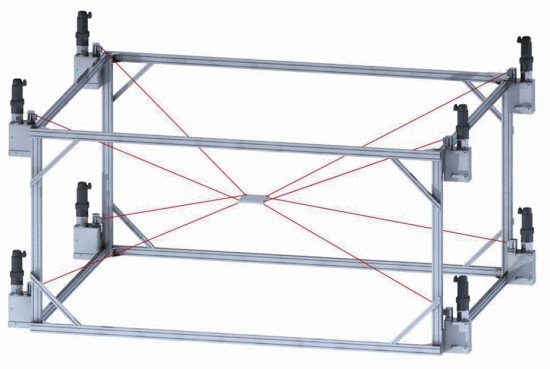}}~\hfil
	\subfigure[Multi-link cable-driven robot \cite{jonathan2018thesis}]{
		\label{ffig:realmcdr}
		\includegraphics[width=0.4\textwidth]{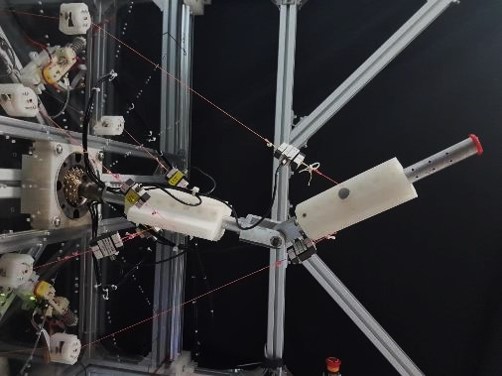}}
	\caption{Cable-driven robots}
	\label{fig:realCDRs}
\end{figure}

As shown in Fig.~\ref{ffig:realcdpr}, the CDPRs can be seen as one branch of the family of parallel robots, where rigid links in conventional parallel robots are replaced by cables. By changing the length of cables simultaneously, the end-effector connected by cables can achieve translation and orientation motions. 

Beside CDPR, another type of cable-actuated system is {\it multi-link cable-driven robots} or called as MCDR \cite{salisbury1982hands,ma1992arm}, in which multiple links are connected in series through passive joints and driven by cables, as depicted in Fig.~\ref{ffig:realmcdr}. Note that cables can pass through several links, which can be referred to as {\it cable routing}. Hence the arbitrary cable routing provides some difficulties and challenges for modelling \cite{darwin2013}, control\cite{rezazadeh2007tensionability,mustafa2012force}, workspace analysis \cite{rezazadeh2011ws,ghasem}, etc. 

In summary, the CDPR and MCDR are considered together and called as {\it cable-driven robots} or CDRs.

\section{Properties and Applications}\label{ssec:propertyApp}
CDRs have attracted much attention due to their promising advantages in comparison with serial robots and rigid-link parallel robots. As a result, numerous prototypes of CDRs were developed around the world and applied to accomplish their specific tasks in practice. In this section, the characteristics of CDRs and associated applications are briefly recalled.

\begin{figure}
    \centering
    \includegraphics[width=0.7 \textwidth]{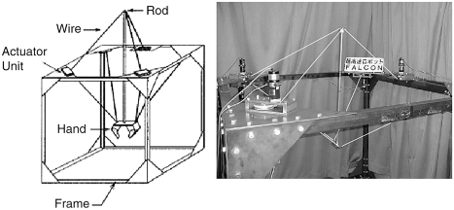}
    \caption{The ultrahigh speed robot FALCON \cite{kawamura1997falcon}}
    \label{fig:falcon}
\end{figure}

First of all, reduced weight and inertia is one of properties for CDRs. Similar to rigid-link parallel robots, the actuators can be located at the base and control each cable individually, which reduces the workload of actuators compared with serial robots. Moreover, the use of lightweight cable decreases the whole inertia of system further. As such, CDRs could be designed for some high or ultra-high speed applications, like IPAnema-Falcon \cite{pott2013ipanema}, the robot family Marionet \cite{merlet2008marionet} and Falcon \cite{kawamura1997falcon} (Fig.~\ref{fig:falcon}). 
%

\begin{figure}
    \centering
    \subfigure[LaserSnake for nuclear decommissioing]{
		\label{ffig:ocRotNuclear}
		\includegraphics[width=0.4 \textwidth]{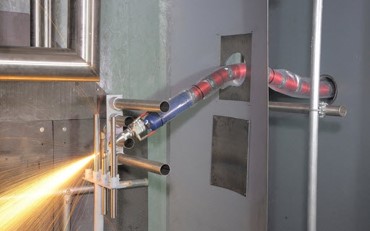}}~\hfil
	\subfigure[Hong Kong JetSnake cleaning and inspection of TBMs]{
		\label{ffig:ocRotClean}
		\includegraphics[width=0.34 \textwidth]{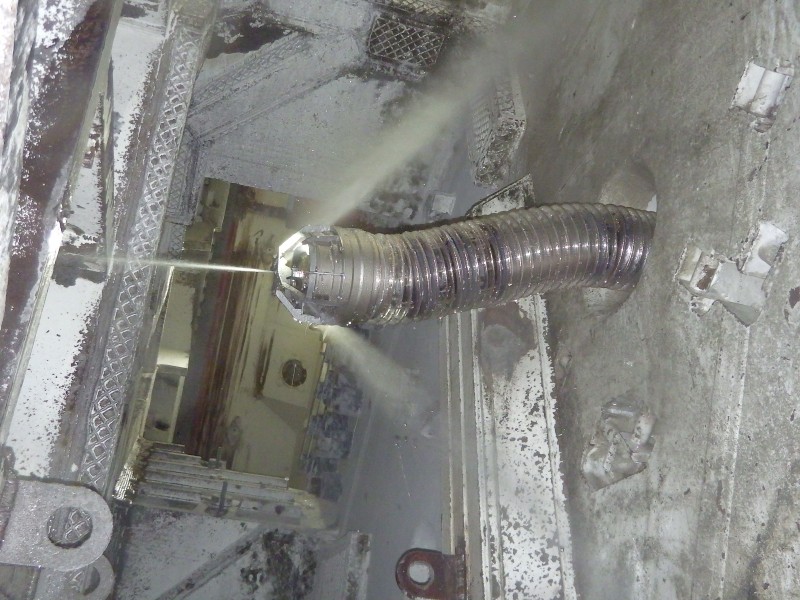}}~\hfil
    \caption{CDRs with high redundancy from OC Robotics Company}
    \label{fig:ocRobsnake}
\end{figure}

In addition, the high redundancy could be achieved in CDRs, especially for MCDR. Therefore, they are dedicated to finish some complicated tasks, like assisting doctors in surgeries \cite{li2013surgery}. As shown in Fig.~\ref{fig:ocRobsnake}, a company called \emph{OC Robotics} aims to employ CDRs to different applications, such as inspection, fastening and cleaning in cluttered environments or hazardous working area.
%

\begin{figure}
    \centering
    \subfigure[Arecibo observatory radio telescope]{
		\label{ffig:arecibo}
		\includegraphics[width=0.45 \textwidth]{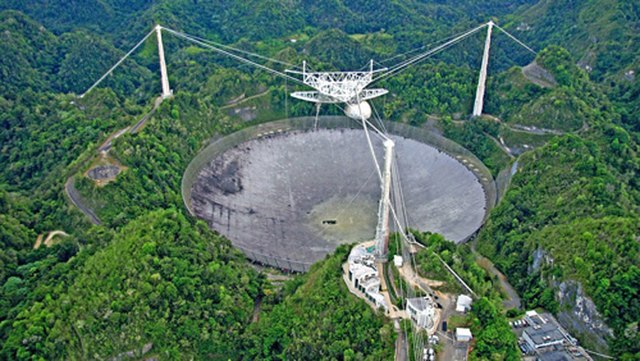}}~\hfil
	\subfigure[FAST telescope]{
		\label{ffig:fast}
		\includegraphics[width=0.4 \textwidth]{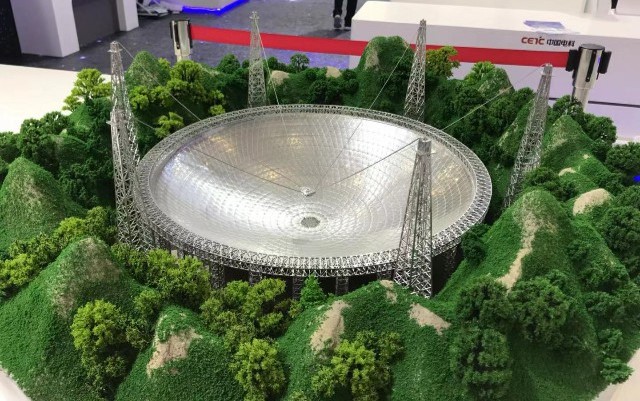}}~\hfil
    \caption{Telescopes with high payload capacity}
    \label{fig:payload}
\end{figure}

Alternatively, the high payload capacity is another benefit for CDRs. For example, in Fig.~\ref{ffig:arecibo} the Arecibo observatory radio telescope has a reflector with a weight of more than 800 tons, similar to the FAST telescope (Five hundred meter Aperture Spherical Telescope) \cite{duan2008fast}, as shown in Fig.~\ref{ffig:fast}. The huge loads can be sustained due to the pulley structures and the internal property of cables, which have been widely-used for construction in ancient age. 

\begin{figure}
    \centering
    \includegraphics[width=0.5 \textwidth]{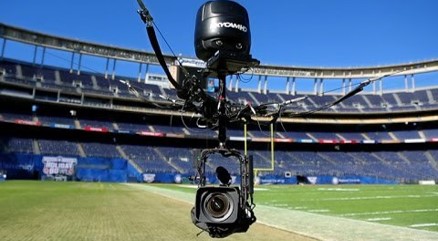}
    \caption{Skycam system over the stadium}
    \label{fig:skycam}
\end{figure}

Last but not least, CDRs are capable of working in the large workspace. Apart from the aforementioned Arecibo and FAST telescope with a size of hundreds of meter, the Skycam is one of successful commercial products benefit from the large workspace of CDRs \cite{cone1985skycam,brown1987skycam}. The Skycam, as depicted in Fig.~\ref{fig:skycam}, can work over the stadium or theater with a camera as the end-effector to capture the splendid moment.

\section{Modelling}\label{ssec:modelling}
\subsection{Kinematics}
\begin{figure}[htbp]
	\centering
	\subfigure[MCDR model with $ p $ links, $n$ DoF and $ m $ cable segments]{
		\label{mod_mcdr}
		\includegraphics[width=0.59 \textwidth]{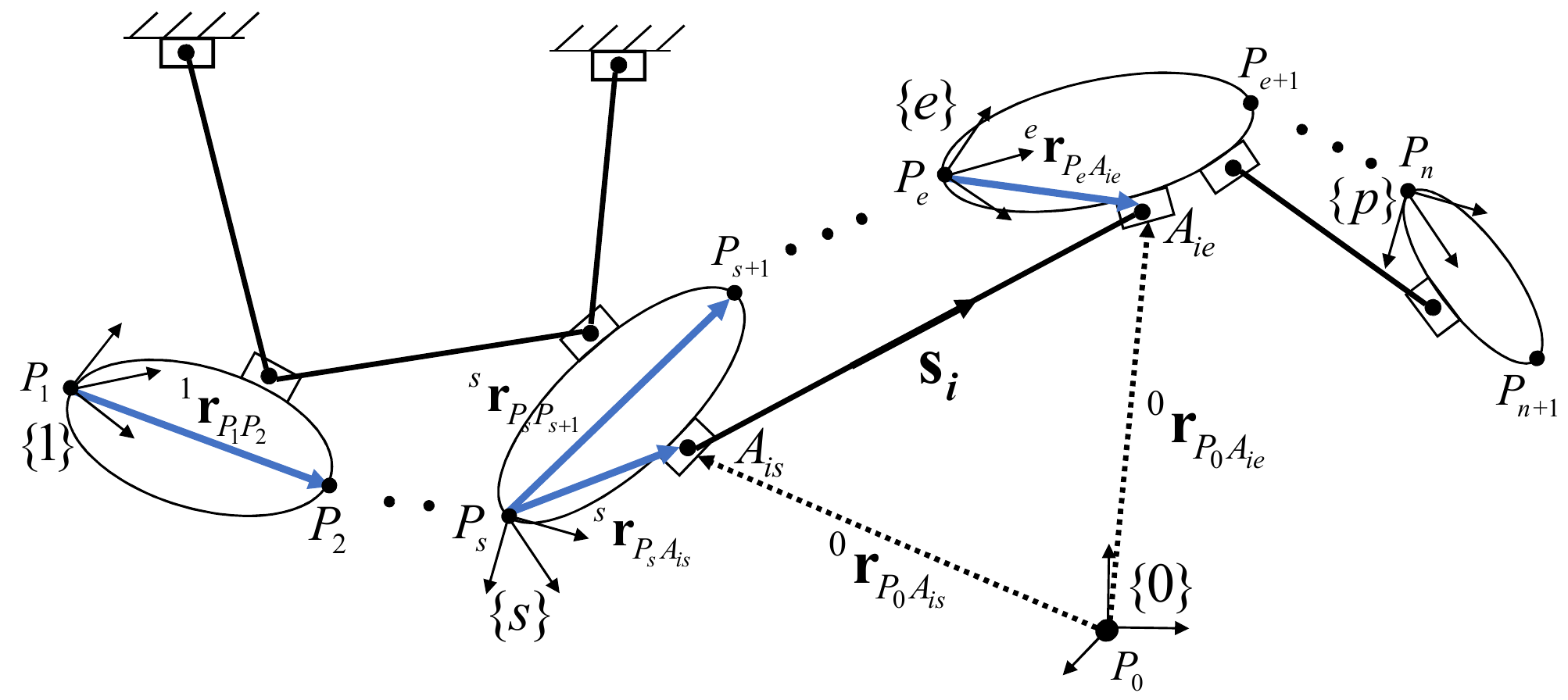}}~\hfil
	\subfigure[CDPR model with single link, $6$ DoF and $m$ cable segments]{
		\label{mod_cdpr}
		\includegraphics[width=0.4\textwidth]{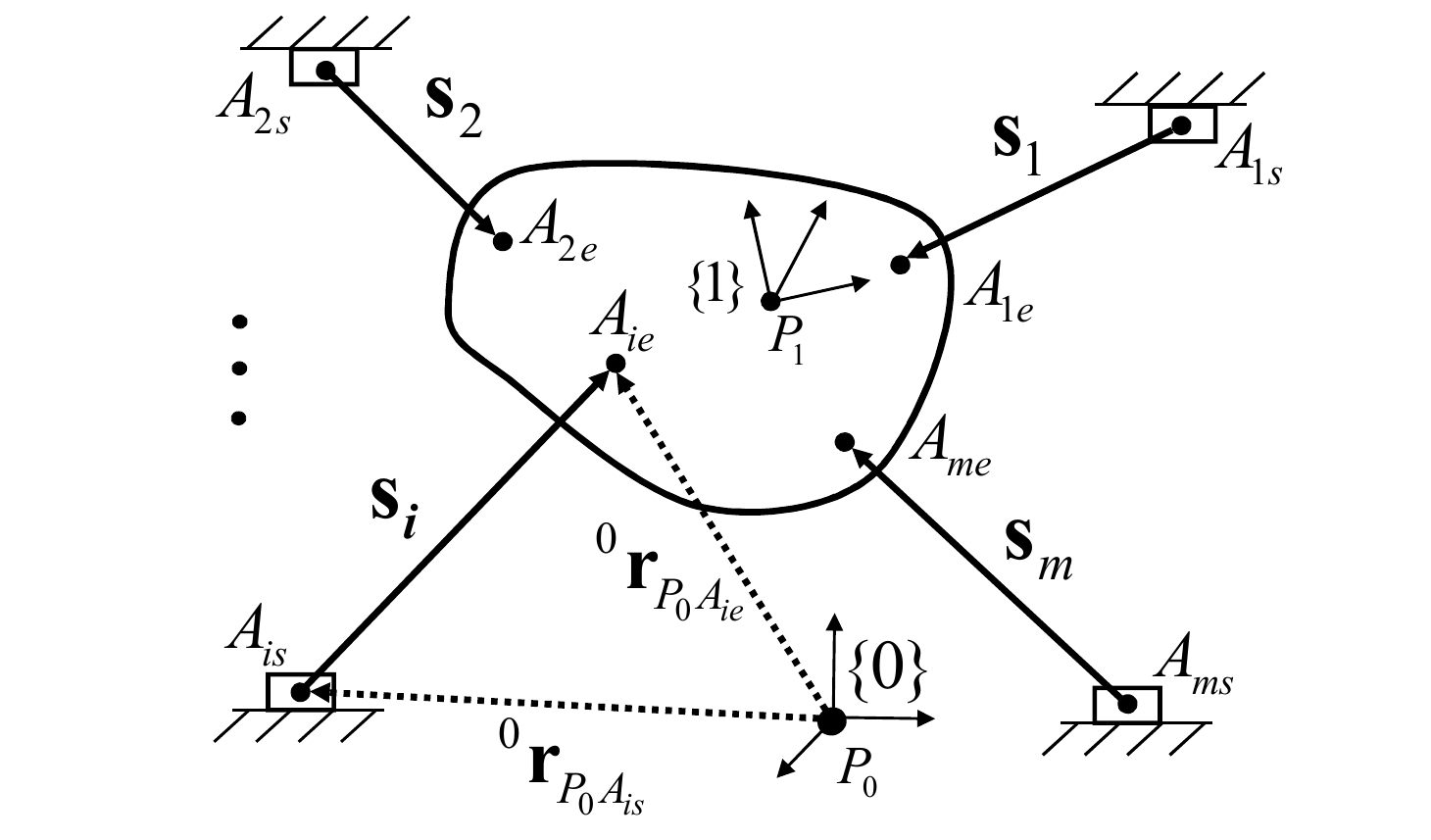}}
	\caption{The models of CDRs}
	\label{kin_model}
\end{figure}

The schematics of CDRs are shown in Fig. \ref{kin_model}, where an $n$-DoF MCDR (in Fig. \ref{mod_mcdr}) has $ p $ rigid links connected by passive joints in a serial open chain. In particular, if $p = 1$, then the robot can be referred to as a CDPR (Fig. \ref{mod_cdpr}). 

The generalized coordinates of the $ n $-DoF $p$-link CDR can be described as $ \mathbf{q} = [\mathbf{q}^T_1, \mathbf{q}^T_2, \mathbf{q}^T_3, \cdots, \mathbf{q}^T_p ]^T \in \mathbb{R}^n$, where $ \mathbf{q}_w $ refers to the generalized coordinates of the $ w $-th link ($1 \leqslant w \leqslant p$). 

In general, a single cable can be routed through one or more links, where the section of cable between two points is defined as a \emph{segment}. Suppose that segment $ i $ of one cable is attached from the link $ s $ to the link $ e $. Without loss of generality, it shall be assumed that $s < e\in \{0, 1, \cdots, p\}$, where the base is defined as link $ 0 $. The following notations will be used in this thesis:
\begin{itemize}
	\item $ {}^aR_b $ defines the rotation matrix from frames $ \{b\} $ to $\{a\}$.
	\item $ {}^s\textbf{r}_{P_{s}A_{is}}$ and $ {}^e\textbf{r}_{P_{e}A_{ie}}$ are the \emph{relative} position vectors of starting point $A_{is}$ and end point $A_{ie}$ of the \emph{segment} $ i $ of a cable expressed in the local frames $ \{s\} $ and $ \{e\} $, respectively.
    \item $ {}^0\textbf{r}_{P_{0}A_{is}}$ and $ {}^0\textbf{r}_{P_{0}A_{ie}} $ are the \emph{absolute} position vectors of starting point $A_{is}$ and end point $A_{ie}$ of the \emph{segment} $ i $ of a cable expressed in the global frame $\{0\}$, respectively.
	\item $ {}^c\textbf{r}_{P_{c}P_{c+1}}$ refers to the vector from the start of link $c$ to the start of link $c+1$ expressed in the local frame $\{c\}$.
\end{itemize}
\begin{property} \label{P1}
	The vectors $ {}^s\textbf{r}_{P_{s}A_{is}}$, $ {}^e\textbf{r}_{P_{e}A_{ie}}$ and $ {}^c\textbf{r}_{P_{c}P_{c+1}} $ are the geometrical parameters $\mathcal{G}$ of the CDR, say 
	$
	    \mathcal{G} = \{{}^s\textbf{r}_{P_{s}A_{is}}, {}^e\textbf{r}_{P_{e}A_{ie}}, {}^c\textbf{r}_{P_{c}P_{c+1}}\}
	$.
\end{property}

The positions of $A_{is}$ and $A_{ie}$ can be formulated in the global frame $\{0\}$ as
\begin{align}
    A_{is}:~ {}^0\textbf{r}_{P_{0}A_{is}} = \sum_{j = 0}^{s-1} [ {}^0R_j {}^j\textbf{r}_{P_{j}P_{j+1}} ] + {}^0R_s {}^s\textbf{r}_{P_{s}A_{is}} \label{mcdr_s_e}\\
    A_{ie}:~ {}^0\textbf{r}_{P_{0}A_{ie}} = \sum_{j = 0}^{e-1} [ {}^0R_j {}^j\textbf{r}_{P_{j}P_{j+1}} ] + {}^0R_e {}^e\textbf{r}_{P_{e}A_{ie}} \label{mcdr_s_ee}
\end{align}
%
where ${}^0R_0 = I $ is the $3$ by $3$ identity matrix. So the vector for segment $i$, denoted as $\textbf{s}_i$, can be defined by
\begin{align}\label{kin_mcdr}
    \textbf{s}_i =  {}^0\textbf{r}_{P_{0}A_{ie}} - {}^0\textbf{r}_{P_{0}A_{is}}
\end{align}

Here {\it Inverse kinematics} (IK) problem is to compute the cable length given the desired position and orientation of manipulator. In return, {\it forward kinematics} (FK) problems refer to the determination of the manipulator pose \textbf{q} for a given set of cable lengths $\|\mathbf{s}_i\|$.

\subsection{Dynamics}
In the dynamics model, the equation of motion can be expressed in the general form
\begin{align}\label{eq:eom}
    M(\mathbf{q})\ddot{\mathbf{q}}+\mathbf{C}(\mathbf{q},\dot{\mathbf{q}})+&\mathbf{G}(\mathbf{q})+\Gamma_{ext}=-\mathbf{J}^T(\mathbf{q})\mathbf{f} \\
    & \underline{\mathbf{f}} \preceq \mathbf{f} \preceq \Bar{\mathbf{f}} \label{forceBound}
\end{align}
where $ M \in \mathbb{R} ^ {m \times n}, \mathbf{C} \in \mathbb{R} ^{n}, \mathbf{G} \in \mathbb{R} ^{n} $ and $\Gamma_{ext} \in \mathbb{R} ^ {n} $ represent the mass-inertia matrix, centrifugal and Coriolis force vector and external force vector, respectively. Here $J$ refers to the Jacobian matrix mapping from the cable forces, i.e., $\mathbf{f} \in \mathbb{R} ^{m} $, to the wrench space in the left hand side of the \eqref{eq:eom}. Moreover, \eqref{forceBound} implies that cable forces are bounded between minimum and maximum values. 

{\it Inverse dynamics} (ID) problem is to compute the cable forces for given the desired position and orientation of manipulator. Besides that, the solution of forces should satisfy the some constrains of the system, i.e. the positive cable forces, the maximum or minimum that cables can provide. Since the number of cables, i.e., $m$ is more than the DoF of manipulator, i.e., $m$, for completely restrained systems, there are more likely to exit an infinite number of cable force solutions in FD. Hence, the optimization approach could be used, such as \cite{lau2014modelling}.

As regards the {\it forward dynamics} (FD) problems, it refer to the determination of the manipulator pose $\mathbf{q}$ for a given set of cable forces $\mathbf{f}$. Due to non-linearity of the equation of motion, the FD usually use numerical techniques. In order to decrease dynamic effect of cables on the dynamic system and to formulate a constraint to assure that the dynamic model is valid, \cite{pham2005dynamic} presented the dynamic analysis of cable-driven parallel mechanisms. In \cite{ma2005dynamics}, in order to study the microgravity dynamics and contact-dynamics, the ID, which is required for real-time control of the cable manipulator to mimic the dynamic motion under the specific situation, is proposed.

\section{Workspace Determination}\label{sec:wsAnalysis}
One fundamentally important problem related to the design and application of CDRs is \textit{workspace analysis}, referring to the determination of the set of feasible poses (positions and orientations) of the robot. 

\subsection{Dynamic Conditions}
The criteria about dynamics consider the fact that the CDR can produce limited wrenches resulting from cables only applying tension forces. Different types of dynamics workspace include the wrench feasible workspace (WFW) \cite{WFW_1, WFW_2, WFW_3}, wrench closure workspace (WCW) \cite{WCW_1,WCW_2,WCW_3,WCW_4} and static workspace \cite{SW}.

Specifically, {\it wrench closure workspace} (WCW) refers to the poses where there exists at least one set of cable forces that can balance the any wrenches applied on the CDRs. Mathematically, if denote the left hand side of \eqref{eq:eom} as $\mathbf{w}_p$, then WCW can be written as
%
\begin{align}
    \mathcal{W}_{wcw} = \{\mathbf{q}: \exists~\mathbf{f} \succ 0, \forall \mathbf{w}_p \in \mathbb{R}^n, \mathbf{w}_p = -\mathbf{J}^T(\mathbf{q})~\mathbf{f}\}
\end{align}
where $\succ$ indicates $\forall f_i \in \mathbf{f}$ such that $f_i > 0$

Furthermore, {\it wrench feasible workspace} (WFW) can be seen as a subset of WCW. Because it determines the poses in which at least one set of cable forces within some boundary exists to maintain any wrench in a given set, i.e.,
%
\begin{align}\label{cond_wfw}
    \mathcal{W}_{wfw} = \{\mathbf{q}: \exists~\mathbf{f} \in [\underline{\mathbf{f}},\Bar{\mathbf{f}}], \mathbf{w}_p = -\mathbf{J}^T(\mathbf{q})~\mathbf{f}, \forall \mathbf{w}_p \in \mathcal{W}_0\}
\end{align}
in which $\mathcal{W}_0$ is the given set of wrenches and $[\underline{\mathbf{f}},\Bar{\mathbf{f}}]$ refers to the lower bound and upper bound of cable forces.
 
In particular, when the given set of wrenches results from the gravity of end-effector and the load, then {\it static workspace} (SW) can be defined by 
%
\begin{align}
    \mathcal{W}_{sw} = \{\mathbf{q}:  \exists~\mathbf{f} \in [\underline{\mathbf{f}},\Bar{\mathbf{f}}], \mathbf{w}_G = -\mathbf{J}^T(\mathbf{q})~\mathbf{f}\}
\end{align}
where $\mathbf{w}_G$ is the external wrench due to the weight of both the end-effector and the load. From \eqref{cond_wfw}, it implies that the given set of wrenches contains only one vector, i.e., $\mathcal{W}_0 = \{\mathbf{w}_G\}$.

\subsection{Kinematic Conditions}
\emph{Kinematics workspace} depends on the robot's kinematic pose for checking whether the CDRs get into a singular pose and if the interference occurs resulting from cables. 

\begin{figure}[htbp]
	\centering
	\subfigure{
		\includegraphics[width=0.35 \textwidth]{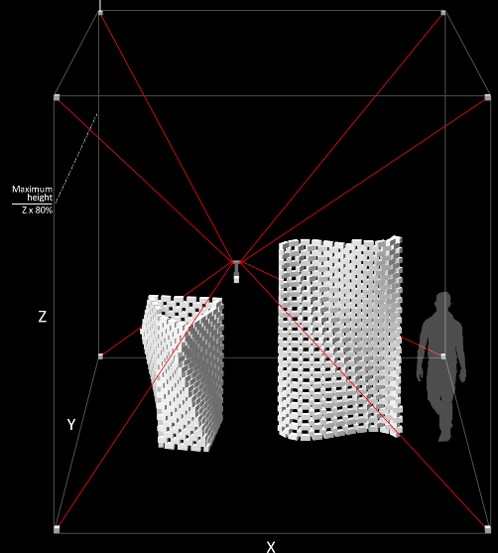}}
	\subfigure{
		\includegraphics[width=0.4\textwidth]{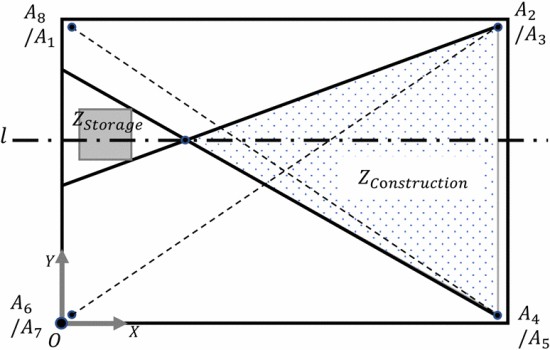}}
	\caption{The interference issues in the CU-Brick system}
	\label{fig:cubrick}
\end{figure}

The workspace avoiding the collisions is called as the interference free workspace (IFW) \cite{ifw_jp,ifw_Go,ifw_app,ifw_app2,ifw_ptwise,ifw-pt-opti,random,ifw_iabm}. The IFW is important as the collision between the cables themselves or with the environment are generally undesired and hence significantly restrict the overall CDR workspace. 
In \cite{ifw_cuhk}, it is observable that the interference problems, including the collision of cables against the brick walls and people, reduce a large proportion of available working area as shown in Fig.~\ref{fig:cubrick}.

\subsubsection{Existing Methods for IFW Determination}
The methods used to determine the IFW can be broadly categorized into analytical and numerical approaches. Analytical approaches aim to generate the closed-form boundaries of the IFW in order to determine the workspace region \cite{ifw_jp, ifw_Go}. These closed-form solutions guarantee high accuracy and provide geometric information about the workspace continuity. However, analytical techniques have only been used to investigate the IFW of particular types of CDRs. Furthermore, existing works make some notable simplifications and assumptions, such as the robot in constant orientation or ignoring the diameter of cables. Under these conditions, the extreme points of the cables and/or the edges of the end-effector all lie in the same plane. These simplify the equations that are required to be solved and thus makes it easier to analytically determine the closed-form IFW boundaries.

Numerical methods are an alternative type of approach to study the IFW in a more general manner, and can be applied on all types of robots. The point-wise approach is a widely-used numerical method where the workspace is determined by checking whether there are any interference (interference free conditions) for a set of poses, sampled discretely by methods such as grid sampling \cite{ifw_ptwise,ifw-pt-opti} or random sampling\cite{random}, to form point cloud. Although straightforward to implement, interference free conditions are only evaluated at finitely many points. This results in the artifact of discretization, where there is no guarantee of workspace continuity between valid points. This becomes an issue when using the workspace for applications that require workspace continuity, such as path planning. One method to limit the impact of this artifact is to use a finer discretization and sample more points. However, this would significantly increase the computation time required. And also, it does not solve the fundamental issue of a lack of workspace continuity.

An alternative to the point-wise approach is interval-analysis-based methods. Merlet \cite{jp-1,jp-2,jp-3} applied interval-analysis to study the workspace of traditional parallel mechanisms. In \cite{ifw_iabm}, the interval-analysis-based method was applied to analyze the IFW of CDRs. For interval-analysis methods, instead of a single pose, a box volume defined by intervals in all of the degrees-of-freedom (DoFs) is considered and the entire box is checked if it is completely within or outside of the IFW. If it is neither scenario, then the box is bisected and re-evaluated again, unless the box is within a predefined size tolerance. Interval-analysis methods provide entire regions and also continuity information compared with the individual poses from point-wise techniques. However, the main drawback is its high computational cost due to the numerous recursive bisections, especially at the workspace boundary.

\begin{figure}[htbp]
	\centering
	\subfigure[Analytical method \cite{ifw_Go} ]{
		\label{ffig:analytical}
		\includegraphics[width=0.45 \textwidth]{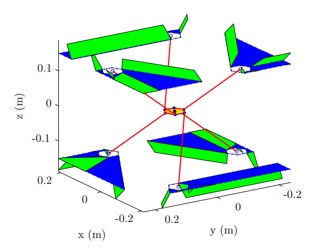}}~\hfil
	\subfigure[Point-wise method \cite{ifw_ptwise}]{
	    \label{ffig:pt_wise}
		\includegraphics[width=0.48\textwidth]{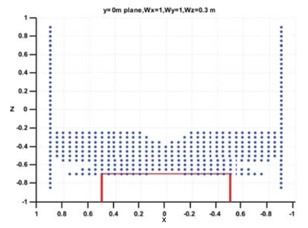}}~\vfil
	\subfigure[Interval analysis \cite{interval}]{
	    \label{ffig:interval}
		\includegraphics[width=0.5\textwidth]{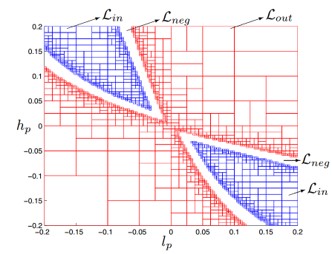}}~\hfil
	\subfigure[Topological approach \cite{topo}]{
	    \label{ffig:topological}
		\includegraphics[width=0.4\textwidth]{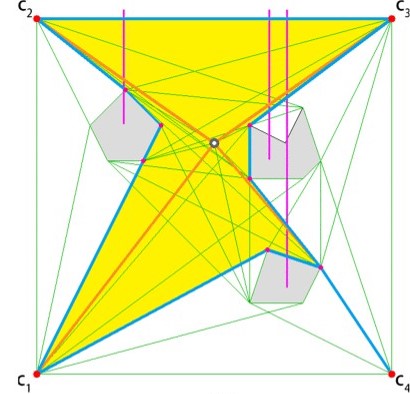}}
	\caption{The methods of the workspace determination for CDRs}
	\label{fig:IFW_methods}
\end{figure}

\subsubsection{Ray-based Method (RBM)}
The ray-based method (RBM) is another numerical workspace analysis method that has attracted attention recently. This concept was first proposed in \cite{linebase} for traditional parallel manipulators to verify whether linear trajectories between two given positions can be executed. Subsequently, \cite{hull,hull_1,hull_2} used the RBM to solve the workspace of CDRs with general conditions by iterative subdivisions. In \cite{DL_wcw}, a RBM dedicated specifically for the analysis of the WCW was proposed, where it was shown that the wrench-closure condition can be formulated as a set of univariate polynomials with respect to one DoF (denoted as the \textit{unknown variable}) while keeping other DoFs constant. Subsequently, \cite{ghasem} extended the RBM to generate a WCW lattice grid that provided additional continuity information between different DoFs. 

\begin{figure}
    \centering
    \includegraphics[width=0.9 \textwidth]{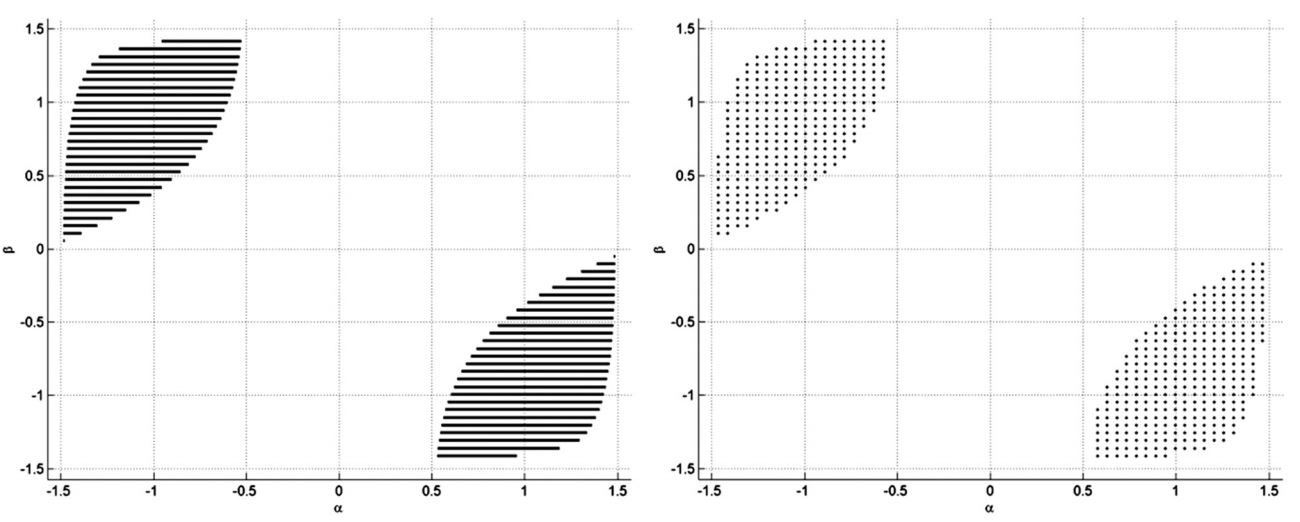}
    \caption{$ \alpha-\beta $ cross sections of the WCW \cite{DL_wcw}}
    \label{fig:wcw_ray_pt}
\end{figure}

For example, Fig.~\ref{fig:wcw_ray_pt} shows the $ \alpha-\beta $ cross sections of the WCW for a 3 DoF manipulator with some certain configurations for the ray-based and point-wise approach in \cite{DL_wcw}.

\section{Path Analysis}
\label{sec:pathstudy}
In all of these applications as introduced in \ref{ssec:propertyApp}, it is important that different tasks can be completed by executing a wide range of paths within the workspace, such as pick-and-place \cite{ifw_cuhk}, cutting of different shapes \cite{crolla2018inflatable} and execution of simulator motion \cite{miermeister2016cablerobot}. 
As such, in this section, \emph{path planning} and \emph{path verification} are introduced, respectively.

\subsection{Path Planning}\label{sssec:astar}
After workspace of CDRs has been determined, the next question may be finding a feasible path that robots can follow it to finish specific tasks. Therefore, in this section, 
one of well known method on path planning, i.e., {\it A-star} algorithm, is introduced.

\begin{figure}[htbp]
	\centering
	\subfigure[The map about parts of Hong Kong]{
		\includegraphics[width=0.45 \textwidth]{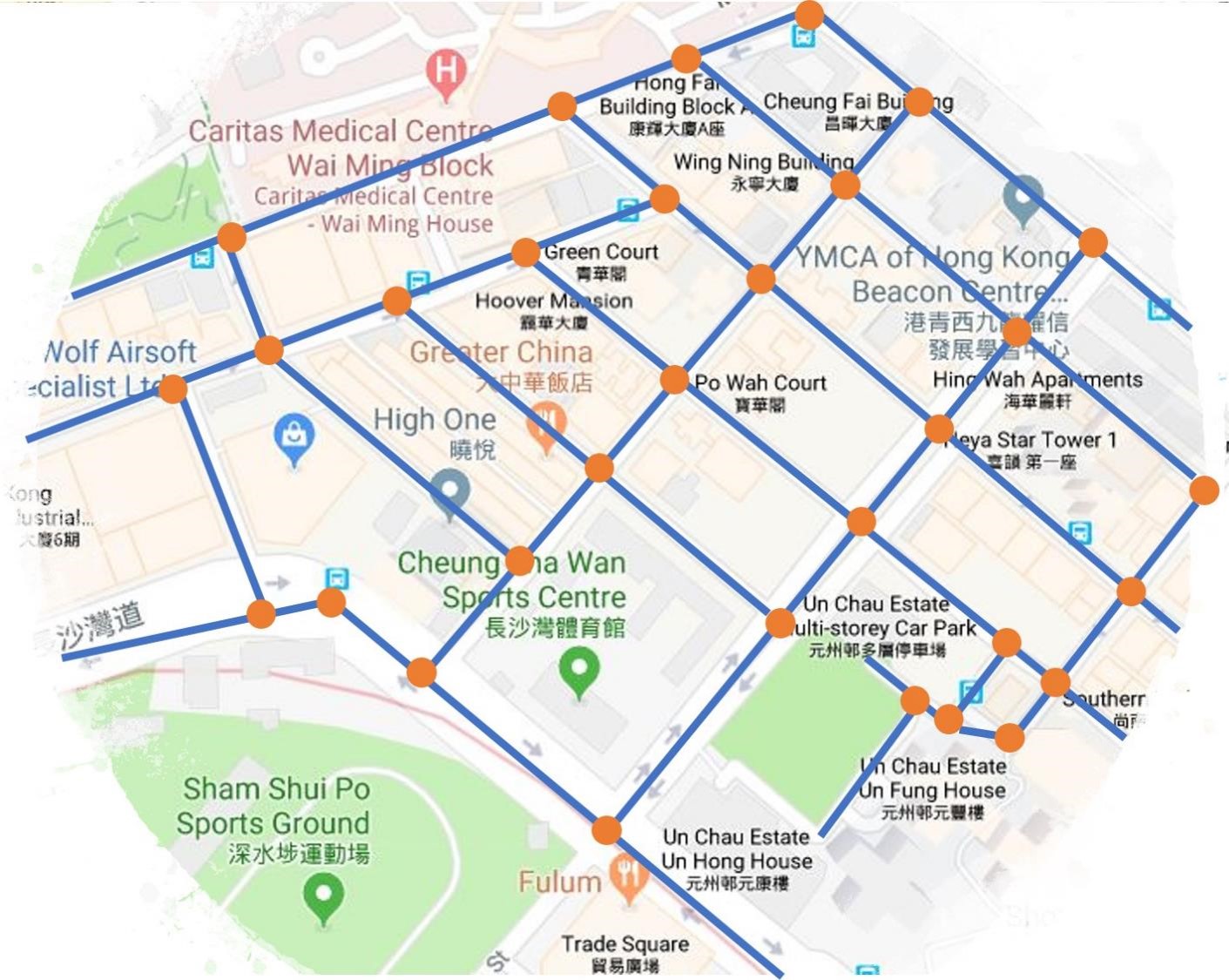}}
	\subfigure[The graph extracted from the map]{
		\includegraphics[width=0.45\textwidth]{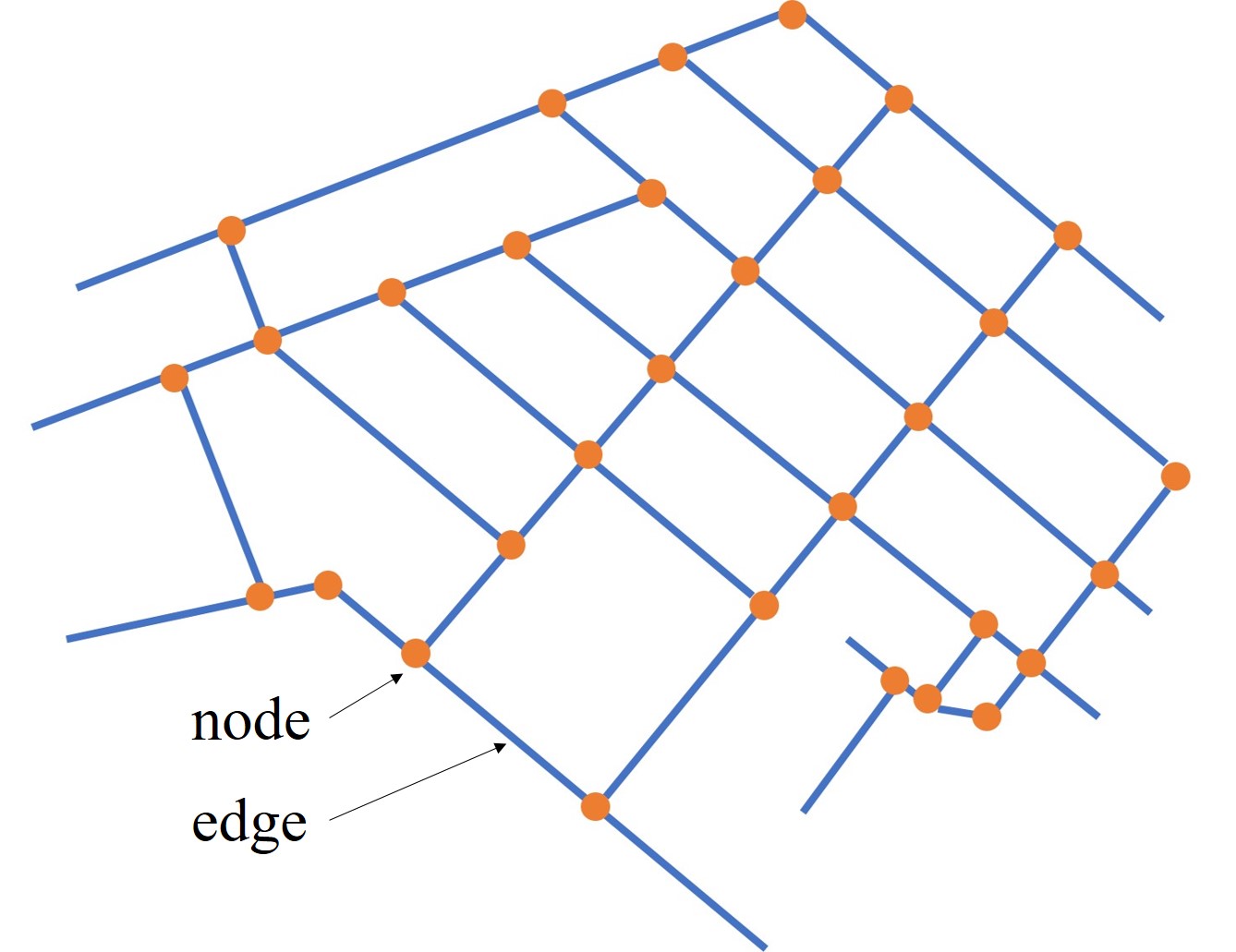}}
	\caption{Convert a practical map into a graph}
	\label{fig:mapinHK}
\end{figure}

\begin{figure}
    \centering
    \includegraphics[width=0.6 \textwidth]{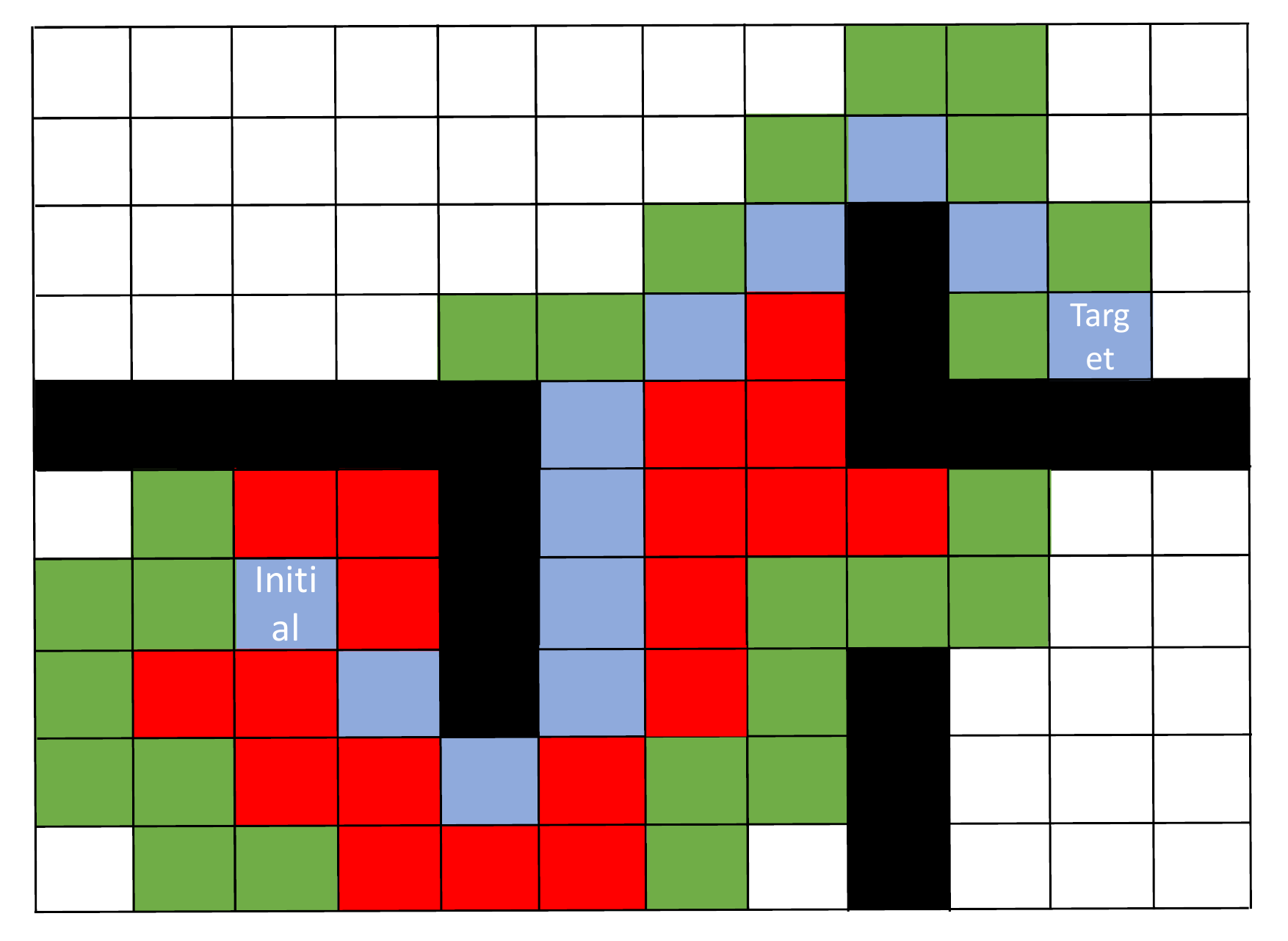}
    \caption{The path generated by A-star algorithm is in blue; the red parts are in $S_{close}$ and green parts are in $S_{open}$. The obstacles are in black.}
    \label{fig:a-star}
\end{figure}

A graph that consists of {\it nodes } and {\it edges} can be extracted from the practical scenarios in the real world, such as 
Fig.~\ref{fig:mapinHK}. In order to find an optimal path from the {\it Initial Node} to the {\it Target Node}, the A-star (or A*) algorithm \cite{hart1968astar} can be implemented in such graph searching problems, as demonstrated in Fig.~\ref{fig:a-star}.

In this method, one node is usually connected to 8 neighbor nodes, and the cost needed to move from node $a$ to its neighbor node $b$ is defined by $ c(n_a, n_b) $.
In addition, the cost function for the node $n$
%
\begin{align}
    f(n) = g^*(n) + h(n)
\end{align}
is used, where $g^*(n)$ refers to the optimal {\it cost-to-come} from the initial node to the node $n$, and $h(n)$ is heuristic function indicating the estimated {\it cost-to-go} from the node $n$ to the target node. The pseudo code is given in Algorithm \ref{algo:astar}.

\begin{algorithm}\label{algo:astar}
    \SetKwInOut{Input}{Input}
    \SetKwInOut{Output}{Output}
    \SetKwInOut{open}{$S_{open}$}
    \SetKwInOut{close}{$S_{close}$}
    \SetKwFunction{insert}{INSERT}
    \SetKwFunction{remove}{REMOVE}
    \caption{A-star algorithm}
    
    \Input{{\it Initial Node} $n_I$, {\it Target Node} $n_T$}
    \Output{The optimal path connecting {\it Initial Node} and {\it Target Node}}
    \open{a priority queue containing $n_I$}
    \close{nodes occupied by the obstacles}
    $n^* \gets$ the node with the least value of $f$ in $S_{open}$\;
    $S_{open}.$\remove{$n^*$} \tcp*{remove $n^*$ from $S_{open}$}
    $S_{close}.$\insert{$n^*$} \tcp*{insert $n^*$ into $S_{close}$}
    \While{$n^* \neq n_T$}
    {
        \For{all neighbor nodes $n$ around $n^*$}
        {
            $g(n) = g^*(n^*) + c(n^*,n)$\;
            \If{$n \in S_{open}$ and $g(n) < g^*(n)$ \label{line:inopen}}
            {
                $g^*(n) \gets g(n)$\;
                $n.f \gets g^*(n) + h(n)$\;
                $n.parent \gets n^*$\;
            }
            \If{$n \notin S_{open}$ and $n \notin S_{close}$ \label{line:notin}}
            {
                $S_{open}.$\insert{n}\;
                $g^*(n) \gets g(n)$\;
                $n.f \gets g^*(n) + h(n)$\;
                $n.parent \gets n^*$\;
            }
        }
        $n^* \gets$ the node with the least value of $f$ in $S_{open}$\; \label{line:argmin_astar}
        $S_{open}.$\remove{$n^*$}\;
        $S_{close}.$\insert{$n^*$}\;
  }
\end{algorithm}

It should be noted that the optimal path can be guaranteed as long as $h(n)$ is an underestimate of the true {\it cost-to-go} from $n$ to the $n_T$. And also, if $h(n)$ gets closer to the real value, the fewer nodes need to be investigated. Therefore, the heuristic function should be well defined and one option is the straight Euclidean distance between $n$ and $n_T$.

Moreover, the OPEN set $S_{open}$ is a priority queue and it needs to return the node efficiently with lowest $f$ value, as shown in Line~\ref{line:argmin_astar}. To do this, for example, the binary heap could be implemented. Additionally, the algorithm should check whether node $n^*$ belongs to the OPEN set $S_{open}$ and CLOSE set $S_{close}$ in Line~\ref{line:inopen} and \ref{line:notin}. So the time-saving search method should be taken into consideration, for instance, the pointer hash map may be utilized. 

\subsection{Path Verification}
\label{sec:pathverfication}
\textit{Path verification} problem is to check whether the path connecting two given poses always satisfies particular workspace conditions or other constraints. This is important as it can be used to ensure that any generated path can actually be executed in practice. In \cite{merlet1994trajectory,merlet2001parser,merlet2001generic,merlet2004guaranteed}, verification for traditional parallel robots was performed using interval analysis over various types of trajectories, including linear \cite{merlet1994trajectory}, gear-shaped \cite{merlet2001parser}, clothoid \cite{merlet2001generic} and conic \cite{merlet2004guaranteed} trajectories.
Even though the feasible sections of the workspace can be obtained in the form of intervals, the bisection process of the interval analysis is very computationally intensive and requires a small tolerance in order to achieve higher computational accuracy.

In \cite{tempel2015design}, the path verification of suspended cable robots for the wrench feasible and interference free conditions and other constraints were demonstrated using a discretization approach. The  was discretely sampled at segmentary steps where the workspace conditions are checked at each discrete pose. The drawback with the discretization approach is that there is no guarantee on the feasibility between consecutive sampled poses. Although this issue can be partially solved by increasing the sampling step, the computational cost would also increase significantly.

Additionally, a point-to-point  verification was proposed in \cite{gosselin2DoF2014} for a 2 degree-of-freedom (DoF) point-mass cable suspended robot.
The verification on positive cable tensions is performed for proposed trajectories. One of proposed trajectories, formulated by trigonometric functions, is validated by solving algebraic equations for the entire  rather than requiring to perform discretization. However, this technique cannot be employed to verify trajectories with varying orientation, due to the highly non-linear conditions caused by the trigonometric terms.

\chapterend
\pdfoutput=1
\chapter{Ray-based Interference Free Workspace (IFW)}

\section{Interference Conditions}\label{ssec:cable_interfer_cond}
In this subsection, three cases according to different interference scenarios are introduced. The interference conditions between two cable segments is first introduced in Sec. \ref{ssec:cond_seg_seg}, followed by the case of the cable segment interfering with a point in Sec. \ref{cond_seg_pt}. Finally, the interference occurring between cable segments and triangular faces is covered in Sec. \ref{cond_seg_triangle}.

\subsection{Interference between Two Segments}\label{ssec:cond_seg_seg}
\begin{figure}
    \centering
    \includegraphics[width=0.7\textwidth]{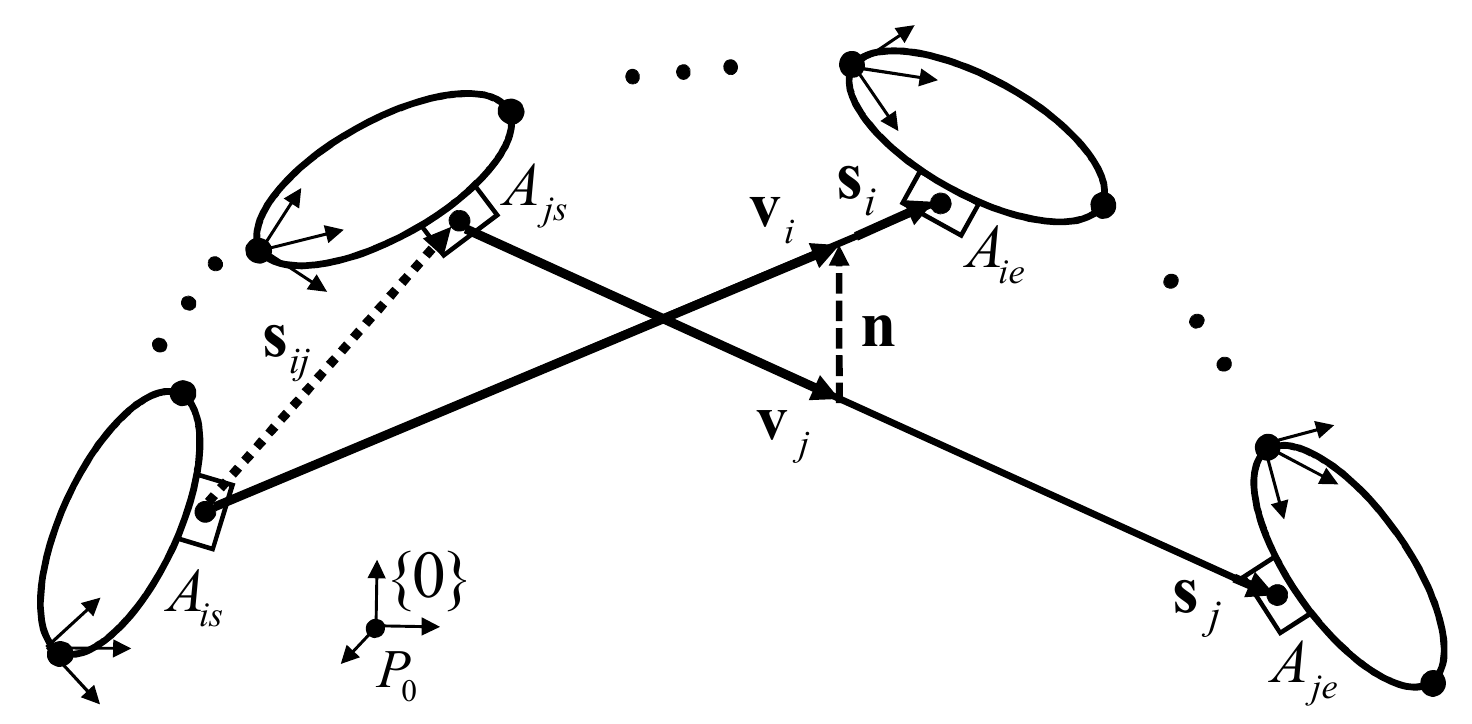}
    \caption{Interference between segment $ \mathbf{s}_i $ and segment $ \mathbf{s}_j $}
    \label{mindis_segs}
\end{figure}

The scenario of two \emph{non-parallel segments}, denoted by $\textbf{s}_i$, $\textbf{s}_j$ and $\textbf{s}_i \times \textbf{s}_j \neq \textbf{0}$ ($i \neq j$), is shown in Fig. \ref{mindis_segs}. The vector $\textbf{n}$ is the common perpendicular of these two segments and $\mathbf{s}_{ij}$ is the vector between the starting attachment points of the two segments, that is 
\begin{align}\label{eq:s_ij}
    \mathbf{s}_{ij} = {}^0\mathbf{r}_{A_{is}A_{js}}
\end{align}
By this notation, the following relationships can be defined 
\begin{align} \label{cp_vector}
    \textbf{n} = t~\textbf{s}_i \times \textbf{s}_j,\ \textbf{v}_i = t_i\ \textbf{s}_i,\ \textbf{v}_j = t_j\ \textbf{s}_j
\end{align}
where $t$, $t_i$, $t_j$ are scalar values, and the shortest distance between $\mathbf{s}_i$ and $\mathbf{s}_j$ is $\epsilon = \|\textbf{n}\| = |t|~\|\textbf{s}_i\times\textbf{s}_j\|$ and $\|\cdot\|$ refers to the Euclidean norm. \\
From Fig. \ref{mindis_segs}, the following equation can be obtained
\begin{align}
    \textbf{v}_i - \textbf{v}_j - \textbf{n} = \textbf{s}_{ij}
\label{cc-main}
\end{align}
Substituting \eqref{cp_vector} into \eqref{cc-main} results in
\begin{align}\label{lieareq}
    \textbf{M}\textbf{t} = \textbf{s}_{ij}
\end{align}
where $\textbf{M} = [\textbf{s}_i,\ -\textbf{s}_j,\ -\textbf{s}_i \times \textbf{s}_j]$, $\textbf{t} = [t_i,\ t_j,\ t]^T$. Since $ \mathbf{n} \perp \textbf{s}_i, \mathbf{n} \perp \textbf{s}_j$ and $\mathbf{n} \neq \mathbf{0}$ by definition, then \eqref{lieareq} is solvable. As such, the interference occurs between segments $\textbf{s}_i$ and $\textbf{s}_j$ if and only if 
\begin{align} \label{cond_seg_seg}
    0\leqslant t_i\leqslant 1,\ 0\leqslant t_j\leqslant 1,\ \epsilon \leqslant \epsilon_r
\end{align}
The required minimum distance $\epsilon_r$ can be defined as
\begin{align} \label{sd}
    \epsilon_r = d_{diameter} + d_{other}
\end{align}
where $d_{diameter}$ is the diameter of cables and $d_{other}$ refers to other required distances, such as a collision-safe distance depending on the specific applications and/or cable sagging values resulting from nonideal property of cables.

In the case of \emph{parallel segments}, i.e., $\textbf{s}_i \times \textbf{s}_j = \textbf{0}$, then the collision conditions can be more simply expressed as
\begin{align} \label{parallel_cond}
    \epsilon = \frac{\|\textbf{s}_i \times \textbf{s}_{ij}\|}{\|\textbf{s}_i\|} \leqslant \epsilon_r
\end{align}

\subsection{Interference between Segments and Points}\label{cond_seg_pt}
\begin{figure}
    \centering
    \includegraphics[width=0.6\textwidth]{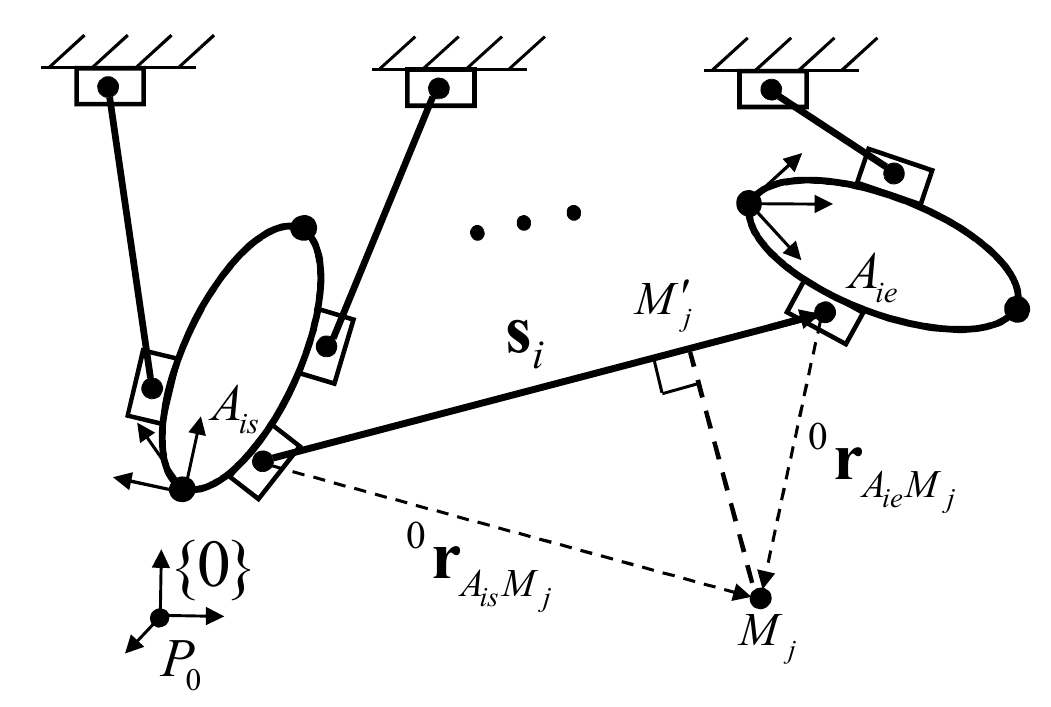}
    \caption{Interference between segment $ \mathbf{s}_i $ and point $ M_j $}
    \label{mindistancecob-pt}
\end{figure}

As in Fig. \ref{mindistancecob-pt}, the $M'_j$ indicates the projection of the point $M_j$ onto the line contacting the segment $\mathbf{s}_i$. Based on the relative position of $M'_j$ with respect to the segment $\mathbf{s}_i$, the minimum distance between $\mathbf{s}_i$ and $M_j$ can be expressed as the following piece-wise function
\begin{align}\label{cond-pt-seg}
    \epsilon = \left \{ 
    \begin{aligned}
        \|{}^{0}\textbf{r}_{A_{is}M_j}\|~~~~~~~~ &\text{if}~~ {}^{0}\textbf{r}_{A_{is}M_j} \cdot \textbf{s}_i \leqslant 0  \\
        \frac{\|\textbf{s}_i \times{}^{0}\textbf{r}_{A_{is}M_j}\|}{\|\textbf{s}_i\|}~~~~~  &\text{if}~~ 0 < {}^{0}\textbf{r}_{A_{is}M_j} \cdot \textbf{s}_i < \textbf{s}_i \cdot \textbf{s}_i \\
        \|{}^{0}\textbf{r}_{A_{ie}M_j}\|~~~~~~~~  &\text{if}~~ \textbf{s}_i \cdot \textbf{s}_i \leqslant {}^{0}\textbf{r}_{A_{is}M_j} \cdot\textbf{s}_i
    \end{aligned}\right.
\end{align}
where $A_{is}$ and $A_{ie}$ are end points of the segment $\mathbf{s}_i$, respectively. The interference condition is $\epsilon \leqslant \epsilon_r$.

\subsection{Interference between Segments and Triangles}\label{cond_seg_triangle}
\begin{figure}
    \centering
    \includegraphics[width=0.6\textwidth]{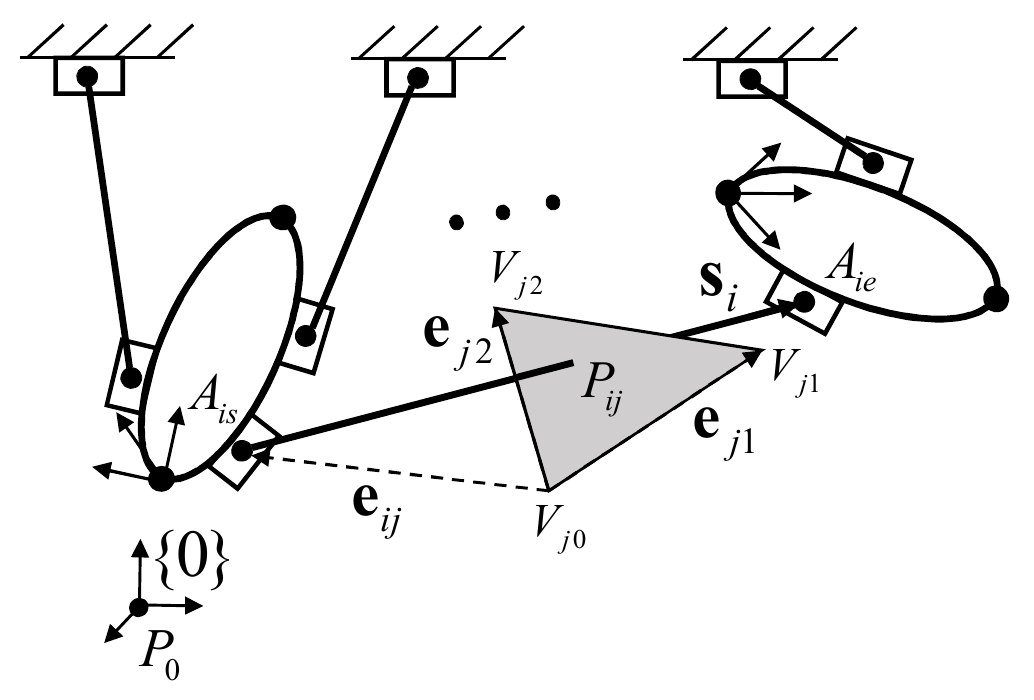}
    \caption{Interference between segment $ \mathbf{s}_i $ and $j$-th triangle}
    \label{cob-poly}
\end{figure}

From \cite{rayTri}, the $j$-th triangle is defined by its vertices, $V_{j0}$, $V_{j1}$ and $V_{j2}$ in Fig. \ref{cob-poly}. The intersection point $P_{ij}$ between $i$-th segment and $j$-th triangle is given by
\begin{align}\label{seg_tri}
    {}^{0}\mathbf{r}_{P_{0}A_{is}} + k\mathbf{s}_i = (1-k_1-k_2){}^{0}\mathbf{r}_{P_{0}V_{j0}}+k_1{}^{0}\mathbf{r}_{P_{0}V_{j1}}+k_2{}^{0}\mathbf{r}_{P_{0}V_{j2}}
\end{align}
where $k$ refers to the normalized distance from $A_{is}$ to $P_{ij}$, and $k_1, k_2$ are barycentric coordinates of $P_{ij}$ on the triangle. Let $\mathbf{e}_{j1} = {}^{0}\mathbf{r}_{P_{0}V_{j1}}-{}^{0}\mathbf{r}_{P_{0}V_{j0}}$, $\mathbf{e}_{j2} = {}^{0}\mathbf{r}_{P_{0}V_{j2}}-{}^{0}\mathbf{r}_{P_{0}V_{j0}}$ and $\mathbf{e}_{ij} =  {}^{0}\mathbf{r}_{P_{0}A_{is}}-{}^{0}\mathbf{r}_{P_{0}V_{j0}}$, then \eqref{seg_tri} can be rewritten as 
\begin{align}\label{eq_segTri}
    \mathbf{Q}\mathbf{k} = \mathbf{e}_{ij}
\end{align}
where $\mathbf{Q} = [-\mathbf{s}_i,\ \mathbf{e}_{j1},\ \mathbf{e}_{j2}]$ and $\mathbf{k} = [k,~k_1,~k_2]^T$. The linear equation \eqref{eq_segTri} is solvable if and only if $\mathbf{Q}$ is invertible, that is $\det{\mathbf{Q}} \neq 0$. In this case, the interference conditions can be stated as follows
\begin{align}\label{cond_segTri}
    \ 0\leqslant k \leqslant1,\ 0\leqslant k_1,\ 0\leqslant k_2,\ k_1 + k_2 \leqslant 1
\end{align}
In the case of $\det{\mathbf{Q}} = 0$, then $\mathbf{s}_i$ is parallel to the $j$-th triangle in virtue of $\det{\mathbf{Q}} = -\mathbf{s}_i \cdot (\mathbf{e}_{j1} \times \mathbf{e}_{j2})$, and the collision condition becomes
\begin{align}\label{cond_segTri_parallel}
    \frac{\|\mathbf{s}_i \times \mathbf{e}_{ij}\|}{\|\mathbf{s}_i\|} \leqslant \epsilon_r
\end{align}

\section{Ray-based IFW for Cables-Cables}
\label{sec:cable_cable_IFW}
Utilizing the interference conditions reviewed in Sec. \ref{ssec:cable_interfer_cond}, the IFW for cable segments is determined by the ray-based method. Furthermore, it will be shown that these processes could be generalized to any CDRs conveniently due to a proposed numerical method later on.

\subsection{Ray-based Cable Segment Equations}
\label{ssec:RB_cable_seg_eq}
The generalized coordinates of the $ n $-DoF $p$-link CDR can be described as $ \textbf{q} = [\textbf{q}^T_1, \textbf{q}^T_2, \textbf{q}^T_3, \cdots, \textbf{q}^T_p ]^T \in \mathbb{R}^n$, where $ \textbf{q}_w $ refers to the generalized coordinates of the $ w $-th link ($1 \leqslant w \leqslant p$). 
In the ray-based method, one of generalized coordinates $\tilde{q} \in \textbf{q}$ is selected as the \emph{unknown variable} while the remaining variables $ \bm{\kappa} =\textbf{q}\setminus\{\tilde{q}\} \in \mathbb{R}^{n-1}$ are considered to be constant variables. In this case, the following property regarding the form of the cable segment vector holds.
\begin{property}\label{Fact}
    The $i$-th cable segment vector $\mathbf{s}_i$ can be transformed into the following form for any $(\tilde{q},\bm{\kappa},\mathcal{G})$ combination
    \begin{align}\label{sVectMat}
        \mathbf{s}_i = [s_x, s_y, s_z]^T = \frac{1}{\rho(\tilde{q})}\mathbf{C}(\bm{\kappa},\mathcal{G})\mathbf{u}(\tilde{q})
    \end{align}
    where $\rho(\tilde{q}) \in \mathbb{R}$ and $\mathbf{u}(\tilde{q}) \in \mathbb{R}^3$ are functions of $\tilde{q}$ while $\mathbf{C}(\bm{\kappa},\mathcal{G}) \in \mathbb{R}^{3 \times 3}$ is a constant matrix, called Coefficient Matrix, which only depends on $\bm{\kappa}$ and the geometrical parameters $\mathcal{G}$ (Property \ref{P1}) of the CDR.
\end{property}

Depending on the type of $\tilde{q}$, Theorems \ref{them1} and \ref{them2} describe the analytical forms of the cable segment vector \eqref{sVectMat}. 
%
\begin{theorem}\label{them1}
    If $\tilde{q}$ is an orientation variable, then $\mathbf{s}_i$ can be expressed as a fraction where all the numerator and denominator terms are univariate polynomial equations with 2-nd degree, that is
    $$ \rho = u^2 + 1, \mathbf{u} = [u^2, u, 1]^T, \mathbf{C} \in \mathbb{R}^{3\times3}\ \text{where}\ u = \tan(\tilde{q}/2)$$
\end{theorem}
%
\begin{proof}
    Without loss of generality, suppose that $\tilde{q}$ is selected from the $\mathbf{q}_w$. Therefore, it is known that $\tilde{q}$ is contained in the rotation matrix ${}^{w-1}R_w$ in the form of $\sin{\tilde{q}}$ and $\cos{\tilde{q}}$, whose degrees are at most 1.
    
    \noindent
    Using the Weierstrass substitution
	\begin{equation}\label{wei_substition}
	    u = \tan\frac{\tilde{q}}{2}\ \ \cos \tilde{q} = \frac{1-u^2}{1+u^2}\ \ \sin\tilde{q} = \frac{2u}{1+u^2}
	\end{equation}
	to $\sin{\tilde{q}}$ and $\cos{\tilde{q}}$, then ${}^{w-1}R_w$ can be rewritten as
	\begin{align}\label{w-1_R_w}
	    {}^{w-1}R_w \in \frac{1}{u^2+1}
	    \begin{bmatrix}
	    {O}(u^2) & {O}(u^2)&{O}(u^2)\\
	    {O}(u^2) & {O}(u^2)&{O}(u^2)\\
	    {O}(u^2) & {O}(u^2)&{O}(u^2)
	    \end{bmatrix}
	\end{align}
	where $O(u^2)$ refers to the quadratic univariate polynomial equations of $u$, and can be explicitly expressed in the form
	\begin{align}\label{vectorOUSquare}
	    c_{2}u^2+c_{1}u+c_{0} = [c_{2}, c_{1}, c_{0}][u^2, u, 1]^T
	\end{align}
    Based on the relative position of link $w$ with link $s$ and link $e$, three different cases should be considered in construction $\mathbf{s}_i$.
    \renewcommand{\theenumi}{\roman{enumi}}
	\begin{enumerate}
		\item \label{P2-1} When $w \leqslant s < e$, then ${}^{w-1}R_w$ exists in \eqref{mcdr_s_e} and \eqref{mcdr_s_ee}, which can be expressed explicitly as
		%
		\begin{align}\label{ris}
        {}^0\textbf{r}_{P_{0}A_{is}} = 
        \underbrace{\sum_{k = 0}^{w-1} [{}^0R_k {}^k\textbf{r}_{P_{k}P_{k+1}}]}_{\mathbf{a}} 
        + \underbrace{\sum_{k = w}^{s-1} [{}^0R_k {}^k\textbf{r}_{P_{k}P_{k+1}}] + {}^0R_s {}^s\textbf{r}_{P_{s}A_{is}}}_{\mathbf{b}} \nonumber \\
        {}^0\textbf{r}_{P_{0}A_{ie}} = 
        \underbrace{\sum_{k = 0}^{w-1} [{}^0R_k {}^k\textbf{r}_{P_{k}P_{k+1}}]}_{\mathbf{a}} 
        + \underbrace{\sum_{k = w}^{e-1} [{}^0R_k {}^k\textbf{r}_{P_{k}P_{k+1}}] + {}^0R_e {}^e\textbf{r}_{P_{e}A_{ie}}}_{\mathbf{c}}
		\end{align}
		%
		In $\mathbf{b}$ and $\mathbf{c}$, ${}^0R_k = {}^0R_{w-1}{}^{w-1}R_w{}^{w}R_k$ while ${}^0R_{w-1}$ and ${}^{w}R_k$ are constant matrices since $\tilde{q} \in \mathbf{q}_w$ is only involved in the ${}^{w-1}R_w$. Therefore, based on \eqref{w-1_R_w}, ${}^0R_k$ in $\mathbf{b}$ and $\mathbf{c}$ takes the form of
		\begin{align}\label{o_R_k}
		    {}^0R_k \in  \frac{1}{u^2+1}[O(u^2), O(u^2), O(u^2)]^T\ \ (k = w,\cdots,e)
		\end{align}
		Furthermore, since $\tilde{q}$ is an orientation variable, then all elements in $\mathcal{G}$ are constant vectors. So $\mathbf{s}_i$ can be transformed from $\eqref{kin_mcdr}$ as
		\begin{align}\label{s_i_vector}
		    \mathbf{s}_i = \mathbf{c} - \mathbf{b} \in \frac{1}{u^2+1}[O(u^2), O(u^2), O(u^2)]^T
	    \end{align}
		Substituting \eqref{vectorOUSquare} into \eqref{s_i_vector}, yields
		\begin{align}\label{s_i_explicit}
		    \mathbf{s}_i = \frac{1}{u^2+1}
		    \begin{bmatrix}
		    c_{x2}&c_{x1}&c_{x0}\\
		    c_{y2}&c_{y1}&c_{y0}\\
		    c_{z2}&c_{z1}&c_{z0}
		    \end{bmatrix}
		    \begin{bmatrix}
		    u^2\\ u\\ 1
		    \end{bmatrix}
		\end{align}
		\item 
		When $s < w \leqslant e$, then ${}^{w-1}R_w$ only exists in \eqref{mcdr_s_ee}. It implies that ${}^0\textbf{r}_{P_{0}A_{is}}$ from \eqref{mcdr_s_e} is a constant vector. Same as \eqref{ris}, ${}^0\textbf{r}_{P_{0}A_{ie}} = \mathbf{a} + \mathbf{c}$, where $\mathbf{a}$ is also a constant vector as ${}^0R_k$ $(k=1,\cdots,w-1)$ does not involve ${}^{w-1}R_w$. As such, from \eqref{o_R_k} and \eqref{kin_mcdr}, it is shown that $\mathbf{s}_i$ can be transformed in the form of \eqref{s_i_explicit}.
		
		\item \label{P2-2} 
		When $s < e < w$, ${}^{w-1}R_w$ vanishes in \eqref{mcdr_s_e} and \eqref{mcdr_s_ee}. As such, ${}^0\textbf{r}_{P_{0}A_{is}}$ and $ {}^0\textbf{r}_{P_{0}A_{ie}}$ are both constant vectors. Thus $\textbf{s}_i$ is also a constant vector from \eqref{kin_mcdr}. If denoting $\textbf{s}_i = [c_x, c_y, c_z]^T$, then it can be rewritten as $\mathbf{s}_i = [c_x(u^2+1), c_y(u^2+1), c_z(u^2+1)]^T/(u^2+1)$, thereby it is equivalent to state $\mathbf{s}_i$ in the form of \eqref{s_i_explicit}, in which $c_{x2}=c_{x0}=c_x$, $c_{y2}=c_{y0}=c_y$, $c_{z2}=c_{z0}=c_z$ and $c_{x1}=c_{y1}=c_{z1}=0$.
	\end{enumerate}
	From all three cases, it can conclude that if $\tilde{q}$ is an orientation variable, then $\mathbf{s}_i$ can be expressed in the form of \eqref{sVectMat} with $\rho = u^2 + 1, \mathbf{u} = [u^2, u, 1]^T, \mathbf{C} \in \mathbb{R}^{3\times3}$ for $u = \tan \tilde{q}$.
\end{proof}

\begin{theorem}\label{them2}
    If $\tilde{q}$ is a translation variable, then $\mathbf{s}_i$ can be expressed as a set of linear equations, that is
    $$ \rho = 1, \mathbf{u} = [u, 1]^T, \mathbf{C} \in \mathbb{R}^{3\times2}\ \text{where}\ u = \tilde{q} $$
\end{theorem}
\begin{proof}
    Suppose the $\tilde{q}$ is selected from $\mathbf{q}_w$. Since it is a translation variable, then $\tilde{q}$ appears within the vector ${}^w\mathbf{r}_{P_wP_{w+1}}$ with at most degree 1. Moreover, all rotation matrices ${}^0R_{j}, {}^0R_{s}, {}^0R_{e}$ in \eqref{mcdr_s_e} and \eqref{mcdr_s_ee} are constant matrices.
    Similar to the proof in Theorem \ref{them1}, there are 3 cases to be considered.
    \renewcommand{\theenumi}{\roman{enumi}}
    \begin{enumerate}
        \item When $w \leqslant s < e$, the ${}^w\mathbf{r}_{P_wP_{w+1}}$ appears in \eqref{mcdr_s_e} and \eqref{mcdr_s_ee}. From \eqref{ris}, it can be shown that
        \begin{align}\label{linearO}
            {}^0\textbf{r}_{P_{0}A_{is}}, {}^0\textbf{r}_{P_{0}A_{ie}} \in [O(u), O(u), O(u)]^T
        \end{align}
        Thus, $\mathbf{s}_i$ in \eqref{kin_mcdr} can be rewritten as
        \begin{align}\label{s_i_transl}
            \mathbf{s}_i = 
            \begin{bmatrix}
                c_{x1}&c_{x0}\\
                c_{y1}&c_{y0}\\
                c_{z1}&c_{z0}
            \end{bmatrix}
            \begin{bmatrix}
                u\\ 1
            \end{bmatrix}
        \end{align}
        \item
        When $s< w \leqslant e$, then ${}^w\mathbf{r}_{P_wP_{w+1}}$ only appears in \eqref{mcdr_s_ee}, thereby ${}^0\textbf{r}_{P_{0}A_{ie}}$ is in the form of \eqref{linearO} and ${}^0\textbf{r}_{P_{0}A_{is}}$ is a constant vector. As such, $\mathbf{s}_i$ can be described in the form of \eqref{s_i_transl} as well.
        \item
        When $s < e < w$, then ${}^0\textbf{r}_{P_{0}A_{is}}, {}^0\textbf{r}_{P_{0}A_{ie}}$ are both constant vectors. So $\mathbf{s}_i$ takes form of \eqref{s_i_transl} with $c_{x1} = c_{y1} = c_{z1} = 0$.
    \end{enumerate}
    Reference to \eqref{sVectMat}, it can be concluded that $\rho = 1$ and $\mathbf{C} \in \mathbb{R}^{3\times2}$ are the coefficients with respect to $\mathbf{u} = [u, 1]^T$.
\end{proof}

\subsection{Generalized Method to Determine the Coefficient Matrix}
The \textit{Coefficient Matrix} $\mathbf{C}$ in \eqref{sVectMat} is derived from the constant-valued $\bm{\kappa}$ and geometrical parameters $\mathcal{G}$ according to the proof in Sec. \ref{ssec:RB_cable_seg_eq}. In order to determine $\mathbf{C}$ for arbitrary CDRs models, a numerical method for any $ (\tilde{q}, \bm{\kappa}, \mathcal{G})$ is proposed. The main idea of this method is similar to that proposed in \cite{ghasem}. 

Based on the explicit formulations of \eqref{s_i_explicit} and \eqref{s_i_transl}, the relationship in \eqref{sVectMat} can be denoted as 
\begin{align}\label{rho_si}
    \rho(\tilde{q}) \mathbf{s}_i =
    [
        \mathbf{c}_x, \mathbf{c}_y, \mathbf{c}_z
    ]^T
    \mathbf{u}(\tilde{q})
\end{align}
where $\mathbf{c}_x^T, \mathbf{c}_y^T$ and $\mathbf{c}_z^T$ indicate row vectors of $\mathbf{C}$. If $\tilde{q}$ is an orientation variable, then there are 9 unknown coefficients to be determined due to \eqref{s_i_explicit}. It is worth noting that for one known pose $\mathbf{q}$ of the CDR, three entries $s_x, s_y, s_z$ in $\mathbf{s}_i$ could be obtained by \eqref{kin_mcdr}. Thus in order to determine 9 unknown coefficients in \eqref{rho_si}, there needs 3 known poses $ (\tilde{q}_{k}, \bm{\kappa})$ $(k = 1, 2, 3)$. On the other hand, if $\tilde{q}$ is a translation variable, it is shown that 2 known poses $(\tilde{q}_{k}, \bm{\kappa})$ $(k = 1, 2)$ are needed to generate 6 unknown elements in $\mathbf{C}$ considering \eqref{s_i_transl}. Notice that $\tilde{q}$ refers to the type of unknown variable while $\tilde{q}_k$ indicate some scalar value.

Here $\tilde{q}_k$ could be sampled from its prescribed range $[\tilde{q}_{min}, \tilde{q}_{max}]$. Then $\mathbf{u}_k$, $\rho_k$ are calculated and cable segments $\mathbf{s}_{ik}$ are obtained from \eqref{kin_mcdr} according to the known poses $(\tilde{q}_{k}, \bm{\kappa})$ and geometrical parameters $\mathcal{G}$ of the CDR. As a result, $\mathbf{C}$ corresponding to $(\tilde{q}, \bm{\kappa}, G)$ can be determined utilizing the following relationship
\begin{align}\label{determineCoefficients}
    \begin{bmatrix}
    \mathcal{U}_1 \\ \vdots \\ \mathcal{U}_m 
    \end{bmatrix}
    \begin{bmatrix}
        \mathbf{c}_x  \\ \mathbf{c}_y  \\ \mathbf{c}_z
    \end{bmatrix}
    =
    \begin{bmatrix}
        \rho_1 \mathbf{s}_{i1} \\ \vdots \\ \rho_m \mathbf{s}_{im} 
    \end{bmatrix}
\end{align}
where
\begin{align}
    \mathcal{U}_k = 
    \begin{bmatrix}
        \mathbf{u}_k^T & \mathbf{0}     & \mathbf{0}     \\
        \mathbf{0}     & \mathbf{u}_k^T & \mathbf{0}     \\
        \mathbf{0}     & \mathbf{0}     & \mathbf{u}_k^T \\
    \end{bmatrix}
\end{align}
for $k = 1, \dots, m$ in which $ m = 3 $ or $ m = 2 $ depending on whether $\tilde{q}$ is an orientation or a translation variable, respectively. It should be noted that \eqref{determineCoefficients} is a linear equation and could be solved easily.

\subsection{Ray-based Interference Free Conditions}
Substituting the ray-based cable segment equations determined in Sec. \ref{sec:cable_cable_IFW} into the interference conditions in Sec. \ref{ssec:cable_interfer_cond}, the ray-based interference free conditions can be obtained.

Since $\textbf{n} \perp \textbf{s}_i$ and $\textbf{n} \perp \textbf{s}_j$ by definition, then $\mathbf{M}$ in \eqref{lieareq} has full rank and $\mathbf{t}$ can be solved uniquely by
\begin{align}\label{4items}
    \textbf{t} &= \mathbf{M}^{-1} \mathbf{s}_{ij} \nonumber\\
	           &= \frac{1}{\det{\mathbf{M}}}
				\begin{bmatrix}
    				\det{[\mathbf{s}_{ij}, -\textbf{s}_j, -\textbf{s}_i \times \textbf{s}_j]}\\
    				\det{[\textbf{s}_i, \mathbf{s}_{ij}, -\textbf{s}_i \times \textbf{s}_j]}\\
    				\det{[\textbf{s}_i, -\textbf{s}_j, \mathbf{s}_{ij}]}\\
				\end{bmatrix}
				:= \frac{1}{d}
				\begin{bmatrix}
					n_{t_i}\\ n_{t_j} \\ n_{t}
				\end{bmatrix}
\end{align}
Note that $\det(\cdot$) refers to the determinant.
Based on the Theorems \ref{them1} and \ref{them2}, the following corollary can be obtained.
\begin{coro}\label{corollary_1}
    The terms $d, n_{t_i}, n_{t_j}$ and $n_{t}$ can be expressed as
    \begin{align}\label{dntintjntForm}
        d = \tilde{d}(u)/\rho^4, n_{t_i} = \tilde{n}_{t_i}(u)/\rho^4, n_{t_j} = \tilde{n}_{t_j}(u)/\rho^4, n_{t} = \tilde{n}_{t}(u)/\rho^3
    \end{align}
    More specifically, if $\tilde{q}$ is an orientation variable, then 
    \begin{align}\label{dntintjnt_ori}
       \tilde{d}(u), \tilde{n}_{t_i}(u), \tilde{n}_{t_j}(u) \in O(u^8),~\tilde{n}_{t}(u) \in O(u^6)
    \end{align}
    and if $\tilde{q}$ is a translation variable, then 
    \begin{align}\label{dntintjnt_tra}
       \tilde{d}(u), \tilde{n}_{t_i}(u), \tilde{n}_{t_j}(u) \in O(u^4),~\tilde{n}_{t}(u) \in O(u^3)
    \end{align}
\end{coro}
\begin{proof}
    Using the definition of segments $i$ and $j$ from \eqref{rho_si}, then $\textbf{s}_i \times \textbf{s}_j$ can be described as
    \begin{align}\label{cross_sisj}
        \textbf{s}_i\times\textbf{s}_j = \frac{1}{\rho^2}
        \begin{bmatrix}
        \mathbf{c}_{yi}^T\mathbf{U}\mathbf{c}_{zj} - \mathbf{c}_{yj}^T\mathbf{U}\mathbf{c}_{zi}\\
        \mathbf{c}_{zi}^T\mathbf{U}\mathbf{c}_{xj} - \mathbf{c}_{xi}^T\mathbf{U}\mathbf{c}_{zj}\\
        \mathbf{c}_{xi}^T\mathbf{U}\mathbf{c}_{yj} - \mathbf{c}_{yi}^T\mathbf{U}\mathbf{c}_{xj}
        \end{bmatrix}
    \end{align}
    where $\mathbf{U} = \mathbf{u}\mathbf{u}^T$. As such, Theorem \ref{them1} implies that $$\textbf{s}_i\times\textbf{s}_j \in [O(u^4), O(u^4), O(u^4)]^T/\rho^2$$ for the orientation variable $\tilde{q}$. 
    Furthermore, it could be found that $ d = \det\mathbf{M} = \|\textbf{s}_i \times \textbf{s}_j\|^2 $ in \eqref{4items}. As such, substituting \eqref{cross_sisj} into $d$, yields $ d = \tilde{d}(u)/\rho^4 $ with $\tilde{d}(u) \in O(u^{8}) $. 
    However, if $\tilde{q}$ is a translation variable, then $\textbf{s}_i \times \textbf{s}_j \in [O(u^2), O(u^2), O(u^2)]^T/\rho^2$ since $\mathbf{u} = [u, 1]^T$ from the Theorem \ref{them2}, and $\tilde{d}(u) \in O(u^4)$ accordingly.
    
    Note that the vector $\mathbf{s}_{ij} = {}^0\mathbf{r}_{A_{is}A_{js}} = {}^0\mathbf{r}_{P_{0}A_{js}} - {}^0\mathbf{r}_{P_{0}A_{is}}$. Similar to $\mathbf{s}_i$, it can be shown that $\mathbf{s}_{ij}$ can be described in the form of \eqref{sVectMat} and satisfies the Theorem \ref{them1} and Theorem \ref{them2} as well. Therefore, \eqref{4items} can be rewritten as 
    \begin{align}
        \begin{bmatrix}
            n_{t_i}\\ n_{t_j} \\ n_{t}
        \end{bmatrix}
        \in \text{adj}\left(
        \begin{bmatrix}
            \frac{O(u^2)}{\rho} & \frac{O(u^2)}{\rho} & \frac{O(u^4)}{\rho^2}\\
            \frac{O(u^2)}{\rho} & \frac{O(u^2)}{\rho} & \frac{O(u^4)}{\rho^2}\\
            \frac{O(u^2)}{\rho} & \frac{O(u^2)}{\rho} & \frac{O(u^4)}{\rho^2}
        \end{bmatrix}\right)
        \begin{bmatrix}
           \frac{O(u^2)}{\rho} \\\frac{O(u^2)}{\rho}\\ \frac{O(u^2)}{\rho}
        \end{bmatrix}
    \end{align}
    Therefore $n_{t_i}, n_{t_j}$ and $n_{t}$ can be presented in the from of \eqref{dntintjntForm}. Moreover, $\tilde{n}_{t_i}(u), \tilde{n}_{t_j}(u)$ and $\tilde{n}_{t}(u)$ can be derived into univariate polynomial equations as described in \eqref{dntintjnt_ori} and \eqref{dntintjnt_tra}, respectively.
\end{proof}

In the following, the interference conditions of non-parallel cable segments ($\mathbf{s}_i \times \mathbf{s}_j \neq \mathbf{0}$) are first considered. Substituting \eqref{4items} into \eqref{cond_seg_seg} yields
\begin{align}\label{inequlaities}
    n_{t_i} \geqslant 0,\ n_{t_j} \geqslant 0,\ d - n_{t_i} \geqslant 0,\ d - n_{t_j} \geqslant 0,\ \epsilon_r^2 d - n_{t}^2 \geqslant 0
\end{align}
Note that above conditions hold for $d = \|\mathbf{s}_i \times \mathbf{s}_j\|^2 > 0$. Subsequently, substituting \eqref{dntintjntForm} into \eqref{inequlaities} allows a set of univariate polynomial equations of $u$ to be obtained. Mathematically, if define
\begin{align}
    \mathcal{L}_{np} = \{\tilde{n}_{t_i}(u), \tilde{n}_{t_j}(u), \tilde{d}(u) - \tilde{n}_{t_i}(u), \tilde{d}(u) - \tilde{n}_{t_j}(u),  \epsilon_r^2\rho^2\tilde{d}(u)-\tilde{n}^2_{t}(u)\}\nonumber
\end{align}
then $\forall \varrho_i \in \mathcal{L}_{np},~\varrho_i \in O(u^{\eta_i}) $ where the $\eta_i$ can be determined by substituting \eqref{dntintjnt_ori} or \eqref{dntintjnt_tra} into $\varrho_i(u)$.
Therefore, for the case of $\mathbf{s}_i \times \mathbf{s}_j \neq \mathbf{0}$, the interference parts along the proposed path $\mathbf{P}(T)$ between cable segments $i$ and $j$ from \eqref{cond_seg_seg} can be transformed as follows
\begin{align}\label{w_ij_non}
    W^{ij}_{np} = \{u: \tilde{d}(u)>0, \varrho_i(u) \geqslant 0, \forall \varrho_i(u) \in \mathcal{L}_{np}\}
\end{align}

For the case of parallel segments, i.e., $\mathbf{s}_i \times \mathbf{s}_j = \mathbf{0}$, then substituting $\mathbf{s}_i$ and $\mathbf{s}_{ij}$ in the form of \eqref{sVectMat} into \eqref{parallel_cond}, yields a univariate polynomial equation of $u$, that is $\varrho(u) = \epsilon_r^2\|\textbf{s}_i\|^2- \|\textbf{s}_i \times \textbf{s}_{ij}\|^2$.
Then the corresponding interference regions can be stated as
\begin{align}\label{w_ij_para}
    W^{ij}_{p} = \{u: \tilde{d}(u)=0, \varrho(u) \geqslant 0\}
\end{align}

As a result, the ray-based interference workspace among all $m$ cables is given by
\begin{align}
    W = \bigcup_{i=1}^{m}\bigcup_{j=1}^{i-1}(W^{ij}_{np}\cup W^{ij}_{p})
\end{align}
Since Theorems \ref{them1} and \ref{them2}, then the ray-based interference workspace with respect to $\tilde{q}$ becomes
\begin{align}
    \begin{split}\label{transfer2q}
        W_{inter} &= \{\tilde{q}: \tilde{q} = 2\tan^{-1}u, \forall u \in W\},~\text{for $\tilde{q}$ is an orientation DoF}\\
        W_{inter} &= W,~\text{for $\tilde{q}$ is a translation DoF}
    \end{split}
\end{align}
Therefore, ray-based IFW of $\tilde{q}$ between all cable segments is the complementary sets of $W_{inter}$ in the prescribe range of $\tilde{q}$, i.e.,
\begin{align}\label{rayBasedIFW}
    W_{free} = \{\tilde{q}: \tilde{q} \in [\tilde{q}_{min}, \tilde{q}_{max}], \tilde{q} \notin W_{inter}\}
\end{align}

\section{Ray-based IFW for Cables-Obstacles}
\label{sec:cable_obstacle_IFW}
The interference between cables and obstacles can affect the workspace as well. As such, it will be shown that the ray-based method (RBM) is capable of being applied to determine the IFW between cables and several different types of obstacles.

\subsection{Polyhedral Obstacles}\label{polyhedra}
Given an any-shaped obstacle, it can be approximated as the polyhedron with a finite collection of flat faces, straight edges and vertices. Although the intersection between cable segments and polyhedra had been solved in  \cite{rayPoly} and \cite{opt}, the interference free conditions could not be transformed to be a set of univariate polynomial inequalities. This makes them not suited for use in the RBM.

\begin{figure}
    \centering
    \includegraphics[width=0.6\textwidth]{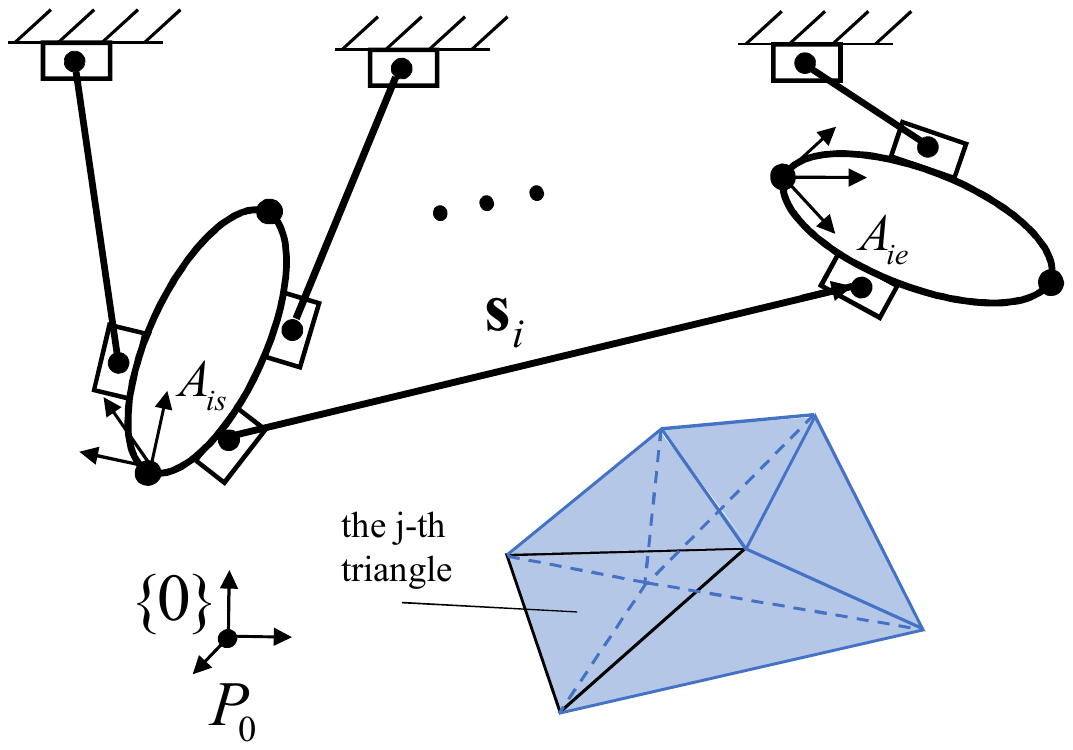}
    \caption{The polyhedral obstacle}
    \label{lab_polyhedron}
\end{figure}

In order to obtain the ray-based IFW, polyhedral obstacles should be represented by triangular faces and defined by their vertices, as demonstrated in Fig. \ref{lab_polyhedron}. Based on the ideas in \cite{rayTri}, it will be shown that the intersection conditions of cables and triangles can be transformed to be a set of univariate polynomial inequalities. Hence the ray-based IFW of cable segments and polyhedral obstacles can be determined.

When $\det{\mathbf{Q}} \neq 0$, the \eqref{eq_segTri} can be solved by Cramer's rule
\begin{align}
    \mathbf{k} &= \mathbf{Q}^{-1}\mathbf{e}_{ij} \nonumber\\
	&=\frac{1}{\det{\mathbf{Q}}}
    \begin{bmatrix}
    	\det{[\mathbf{e}_{ij}, \mathbf{e}_{j1}, \mathbf{e}_{j2}]}\\
    	\det{[-\mathbf{s}_i, \mathbf{e}_{ij}, \mathbf{e}_{j2}]}\\
        \det{[-\mathbf{s}_i, \mathbf{e}_{j1}, \mathbf{e}_{ij}]}\\
    \end{bmatrix}
    = \frac{1}{d}
    \begin{bmatrix}
        n_k\\ n_{k_1} \\ n_{k_2}
    \end{bmatrix}
\end{align}
Similar to the Corollary \ref{corollary_1}, the following corollary can be obtained.
\begin{coro}\label{corollary_2}
    The terms $d, n_k, n_{k_1}$ and $n_{k_2}$ can be expressed as
    \begin{align}\label{dnknk1nk2Form}
        d = \tilde{d}(u)/\rho,~n_k = \tilde{n}_{k}(u)/\rho,~n_{k_1} = \tilde{n}_{k_1}(u)/\rho^2,~n_{k_2} = \tilde{n}_{k_2}(u)/\rho^2
    \end{align}
    where if $\tilde{q}$ is an orientation variable, then 
    \begin{align}\label{dnknk1nk2_ori}
       \tilde{d}(u), \tilde{n}_{k}(u) \in O(u^2),~\tilde{n}_{k_1}(u), \tilde{n}_{k_2}(u) \in O(u^4)
    \end{align}
    if $\tilde{q}$ is a translation variable, then 
    \begin{align}\label{dnknk1nk2_tra}
       \tilde{d}(u), \tilde{n}_{k}(u) \in O(u),~\tilde{n}_{k_1}(u), \tilde{n}_{k_2}(u) \in O(u^2)
    \end{align}
\end{coro}
\begin{proof}
    Substituting the $\mathbf{s}_i$ in the form of \eqref{sVectMat} and constant $\mathbf{e}_{j1}$ and $\mathbf{e}_{j2}$ into $d = \det{\mathbf{Q}}$, yields $d = \tilde{d}(u)/\rho$ with $\tilde{d}(u) \in O(u^2)$ from the Theorem \ref{them1} and $\tilde{d}(u) \in O(u)$ from the Theorem \ref{them2}. 
    
    Furthermore, since $\mathbf{e}_{ij} = {}^{0}\mathbf{r}_{P_{0}A_{is}}-{}^{0}\mathbf{r}_{P_{0}V_{j0}}$ and ${}^{0}\mathbf{r}_{P_{0}A_{is}}$ in the form of \eqref{sVectMat} as well as the constant ${}^{0}\mathbf{r}_{P_{0}V_{j0}}$, then $\mathbf{e}_{ij}$ can be expressed as \eqref{sVectMat}. Therefore taking $\mathbf{e}_{ij}$ into $n_{k} = \det{[\mathbf{e}_{ij}, \mathbf{e}_{j1}, \mathbf{e}_{j2}]}$, obtains $n_{k} = \tilde{n}_k(u)/\rho$ with $\tilde{n}_k(u) \in O(u^2)$ and $\tilde{n}_k(u) \in O(u)$, respectively.
    
    The formulations of $\tilde{n}_{k_1}(u)$ and $\tilde{n}_{k_2}(u)$ in \eqref{dnknk1nk2_ori} and \eqref{dnknk1nk2_tra} can be derived in a similar manner.
\end{proof}

When $\mathbf{s}_i$ is not parallel to the $j$-th triangle, i.e., $d = \det{\mathbf{Q}} \neq 0$, if define 
\begin{align}
    \mathcal{L}_{np} = \{& \tilde{n}_k(u), \tilde{d}(u)-\tilde{n}_k(u), \tilde{n}_{k_1}(u), \tilde{n}_{k_2}(u), \nonumber\\
                          & \rho\tilde{d}(u)-[\tilde{n}_{k_1}(u)+\tilde{n}_{k_2}(u)]\}
\end{align}
then $\forall \varrho_i \in \mathcal{L}_{np}$, $\varrho_i \in O(u^{\eta_i})$. Here $\eta_i$ can be calculated by substituting the \eqref{dnknk1nk2_ori} or \eqref{dnknk1nk2_tra} into $\varrho_i(u)$ accordingly.
Therefore, for the case of $d \neq 0$, the interference regions from the collision conditions \eqref{cond_segTri} can be described equivalently as
\begin{align}
    W_{np}^{ij} = &\{ u: \tilde{d}(u) > 0, \varrho_i(u) \geqslant 0, \forall \varrho_i(u) \in \mathcal{L}_{np}\} \nonumber\\ 
                   &\cup \{u: \tilde{d}(u) < 0, \varrho_i(u) \leqslant 0, \forall \varrho_i(u) \in \mathcal{L}_{np}\}
\end{align}

For the parallel case, i.e., $d = \det{\mathbf{Q}} = 0$, from \eqref{cond_segTri_parallel} it could be found that $ \varrho =  \epsilon_r^2\|\mathbf{s}_i\|^2 - \|\mathbf{s}_i \times \mathbf{e}_{ij}\|^2 $ is an univariate polynomial equation of $u$.
Therefore, the interference regions for this parallel case is given by
\begin{align}
    W_{p}^{ij} = \{ u: \tilde{d}(u) = 0, \varrho(u) \geqslant 0\}
\end{align}
As a result, the ray-based interference workspace between $m$ cables and polyhedral obstacles composed of $r$ triangles is
\begin{align}\label{W_seg_tri}
    W = \bigcup_{i=1}^{m}\bigcup_{j=1}^{r}(W^{ij}_{np}\cup W^{ij}_{p})
\end{align}
Hence, the ray-based IFW $W_{free}$ about $\tilde{q}$ for cables and obstacles can be determined by substituting the \eqref{W_seg_tri} into \eqref{transfer2q} and \eqref{rayBasedIFW}. 

\subsection{Cylindrical Obstacles}\label{cylinder}
\begin{figure}
    \centering
    \includegraphics[width=0.6\textwidth]{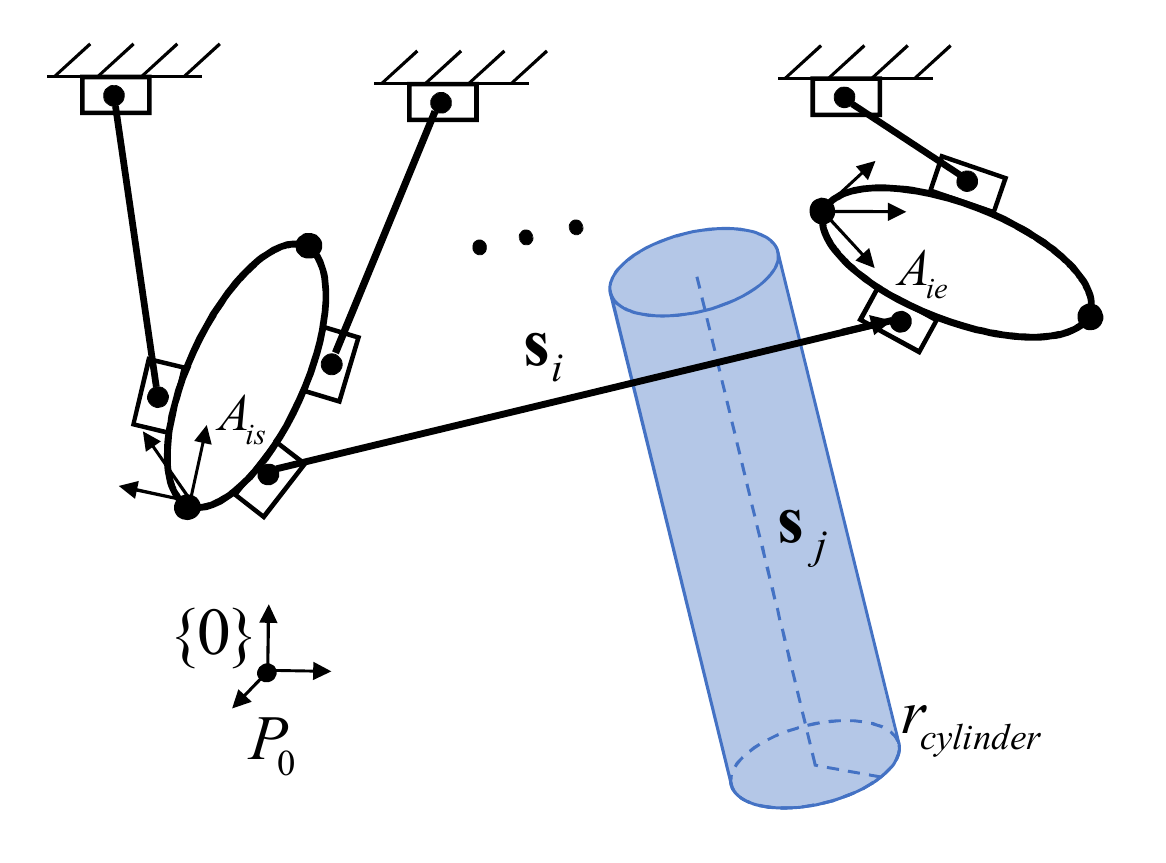}
    \caption{The cylindrical obstacle}
    \label{lab_cylinder}
\end{figure}

The cylindrical obstacle can be seen as the segment $\mathbf{s}_j$ with a constant radius $r_{cylinder}$, as depicted in Fig. \ref{lab_cylinder}. Therefore, the interference of segments to cylinders can be detected in the similar methods proposed in Section \ref{sec:cable_cable_IFW}. The main differences are as follows. 

First the $d, n_{t_i}, n_{t_j}$ and $n_t$ in \eqref{4items} can be expressed differently as 
\begin{align}
     d = \tilde{d}(u)/\rho^2,~n_{t_i} = \tilde{n}_{t_i}(u)/\rho^2,~n_{t_j} = \tilde{n}_{t_j}(u)/\rho^3,~n_{t} = \tilde{n}_{t}(u)/\rho^2
\end{align}
The reason is that the $\mathbf{s}_j$ presenting the $j$-th cylindrical obstacle is a constant vector, thereby the $\mathbf{s}_i \times \mathbf{s}_j $ takes the same form as $\mathbf{s}_i$ in \eqref{sVectMat}. Therefore, from \eqref{4items} it is shown that $ \tilde{d}(u), \tilde{n}_{t_i}(u), \tilde{n}_{t}(u) \in O(u^4)$, $\tilde{n}_{t_j}(u) \in O(u^6)$ for the orientation variable $\tilde{q}$, and  $\tilde{d}(u), \tilde{n}_{t_i}(u), \tilde{n}_{t}(u) \in O(u^2)$, $\tilde{n}_{t_j}(u) \in O(u^3)$ for the translation variable $\tilde{q}$.

The other difference is the required minimum distance between segments and cylinders. It becomes $\epsilon_r = r_{diameter} + r_{cylinder} + d_{other}$.

\subsection{Spherical Obstacles}\label{sphere}
\begin{figure}
    \centering
    \includegraphics[width=0.6\textwidth]{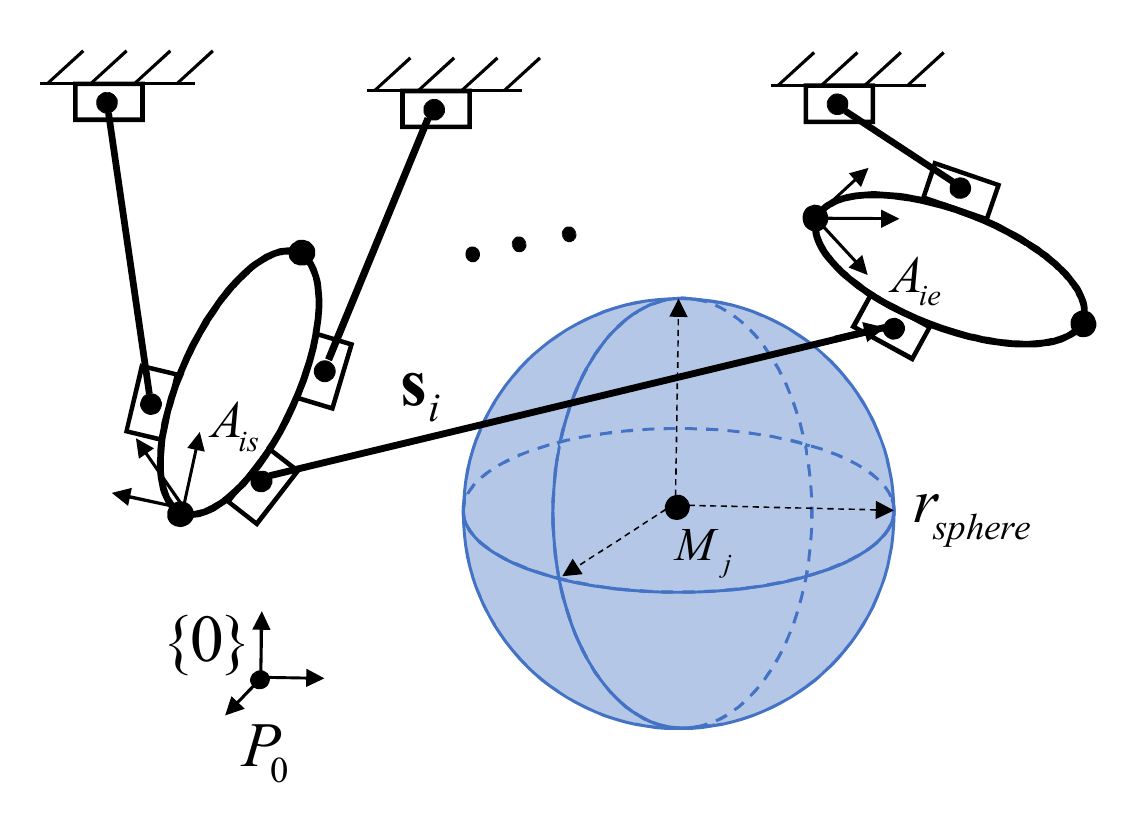}
    \caption{The spherical obstacle}
    \label{lab_sphere}
\end{figure}

As in Fig. \ref{lab_sphere}, the spherical object can be seen as the fixed center point $M_j$ with a constant radius $r_{sphere}$. Thus $\epsilon_r = r_{diameter} + r_{sphere} + d_{other}$. 

From the interference conditions \eqref{cond-pt-seg}, if define
\begin{align}\label{cond_pt_seg_rayBased}
    \begin{split}
        \mathcal{L}_1 &= \{ -{}^0\mathbf{r}_{A_{is}M_j} \cdot \mathbf{s}_i,~\epsilon_r^2 - \|{}^0\mathbf{r}_{A_{is}M_j}\|^2\} \\
        \mathcal{L}_2 &= \{ {}^0\mathbf{r}_{A_{is}M_j} \cdot \mathbf{s}_i,~\|\mathbf{s}_i\|^2-{}^0\mathbf{r}_{A_{is}M_j} \cdot \mathbf{s}_i,~\epsilon_r^2 \|\mathbf{s}_i\|^2-\|{}^0\mathbf{r}_{A_{is}M_j} \times \mathbf{s}_i\|^2\} \\
        \mathcal{L}_3 &= \{ {}^0\mathbf{r}_{A_{is}M_j} \cdot\mathbf{s}_i-\|\mathbf{s}_i\|^2,~\epsilon_r^2 - \|{}^0\mathbf{r}_{A_{ie}M_j}\|^2\} \\
    \end{split}
\end{align}
then $\forall \varrho_i \in \mathcal{L}_k, k\in \{1,2,3\}$, yields $\varrho_i \in O(u^{\eta_i})$, and $\eta_i$ is calculated by substituting the ${}^0\mathbf{r}_{A_{is}M_j}$, ${}^0\mathbf{r}_{A_{ie}M_j}$ and $\mathbf{s}_i$ into the $\varrho_i(u)$. Note that ${}^0\mathbf{r}_{A_{is}M_j} = {}^0\mathbf{r}_{P_{0}M_j} - {}^0\mathbf{r}_{P_{0}A_{is}}$ and ${}^0\mathbf{r}_{A_{ie}M_j} = {}^0\mathbf{r}_{P_{0}M_j} - {}^0\mathbf{r}_{P_{0}A_{ie}}$ also can be transformed as \eqref{sVectMat}.

Then the interference regions between the $i$-th segment and $j$-th sphere defined by its center point can be expressed as follows
\begin{align}
    W^{ij} = \bigcup_{k=1}^{3} \{u: \varrho_i(u) \geqslant 0, \forall \varrho_i(u) \in \mathcal{L}_k \}
\end{align}
As such, the interference regions between all $m$ cables and $h$ spherical obstacles are 
\begin{align}\label{W_seg_pt}
    W = \bigcup_{i=1}^{m}\bigcup_{j=1}^{h}W^{ij}
\end{align}
Finally, the ray-based IFW $W_{free}$ with respect to $\tilde{q}$ for cables and spherical obstacles can be determined by taking \eqref{W_seg_pt} into \eqref{transfer2q} and \eqref{rayBasedIFW} as well.

\subsection{Ellipsoidal Obstacles}
The ellipsoid can be formulated as a set of points $\mathbf{x}$ such that
\begin{align}\label{eq_ellipsoid}
    (\mathbf{x} - \mathbf{c})^T\mathbf{A}(\mathbf{x} - \mathbf{c}) = 1
\end{align}
where $\mathbf{c}$ is the center of the ellipsoid, and $\mathbf{A}$ is a positive definite matrix, whose eigenvectors refer the principal axes of the ellipsoid and eigenvalues are the reciprocals of the squares of the length of semi-axes. Since $\mathbf{A}$ is positive definite, then it can be decomposed as $\mathbf{A} = \mathbf{Q} \mathbf{\Lambda} \mathbf{Q}^T$. The $\mathbf{Q}$ is an orthogonal matrix whose columns are the eigenvectors of $\mathbf{A}$  and $\mathbf{\Lambda}$ is a diagonal matrix whose elements are the eigenvalues of $\mathbf{A}$. Substituting $\mathbf{A} = \mathbf{Q} \mathbf{\Lambda} \mathbf{Q}^T$  into \eqref{eq_ellipsoid}, yields 
\begin{align}\label{ellipsoid_mid}
     \begin{split}
         (\mathbf{x} - \mathbf{c})^T\mathbf{Q} \mathbf{\Lambda} \mathbf{Q}^T(\mathbf{x} - \mathbf{c}) &= 1 \\
        (\mathbf{x} - \mathbf{c})^T\mathbf{Q} (\mathbf{\Lambda}^{1/2})^T \mathbf{\Lambda}^{1/2} \mathbf{Q}^T(\mathbf{x} - \mathbf{c}) &= 1 \\
     \end{split}
\end{align}
where $\mathbf{\Lambda}^{1/2}$ is the matrix whose diagonal elements are the square roots of that in $\mathbf{\Lambda}$.
As such, the affine transformation 
\begin{align}\label{ellipsoid_affine}
    \tilde{\mathbf{x}} = \mathbf{\Lambda}^{1/2} \mathbf{Q}^T(\mathbf{x} - \mathbf{c})
\end{align}
transforms the ellipsoid to the unit sphere centered at the origin since \eqref{ellipsoid_mid} becomes $ \tilde{\mathbf{x}}^T\tilde{\mathbf{x}}= 1$. 

It is worth noting that $\mathbf{\Lambda}, \mathbf{Q}$ and $ \mathbf{c}$ are constant from the ellipsoid. Therefore, the $\tilde{\mathbf{s}}_i$ transformed form $\mathbf{s}_i$ by \eqref{ellipsoid_affine} can be rewritten in the form of \eqref{sVectMat} and satisfy Theorems \ref{them1} and \ref{them2}. Similar to ${}^0\tilde{\mathbf{r}}_{P_{0}A_{is}}$ and ${}^0\tilde{\mathbf{r}}_{P_{0}A_{ie}}$.

\begin{figure}
    \centering
    \includegraphics[width=1\textwidth]{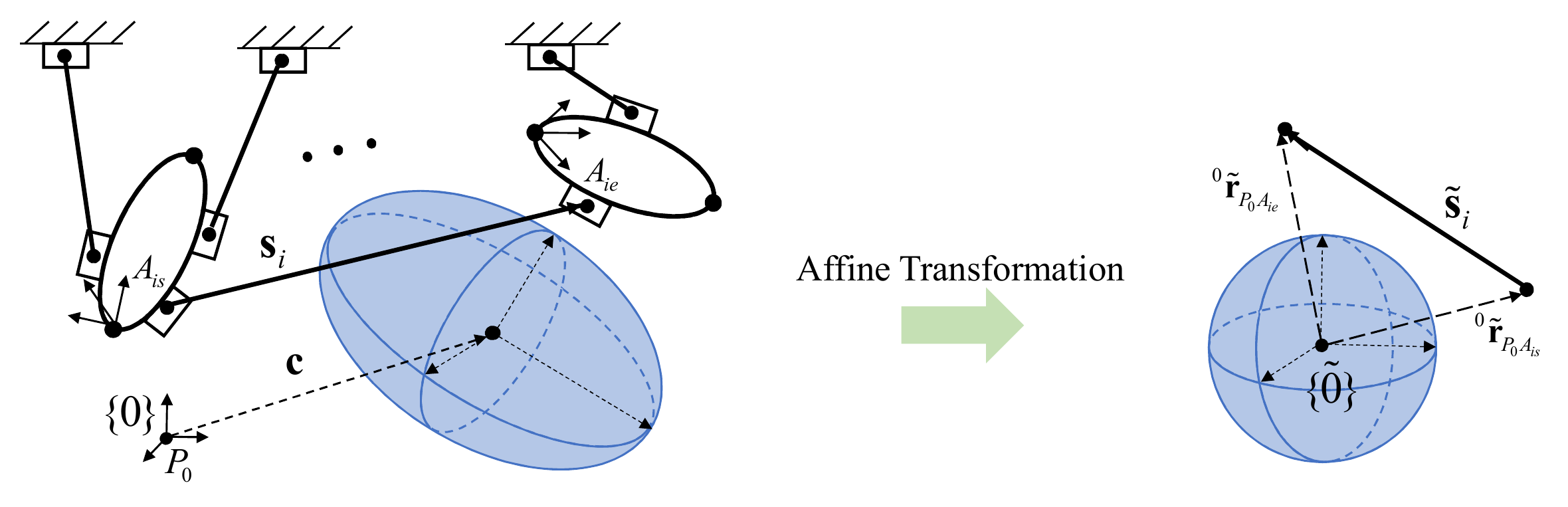}
    \caption{The ellipsoidal obstacle}
    \label{lab_ellipsoid}
\end{figure}

It is shown that the intersection of the segment $\mathbf{s}_i$ and ellipsoid is transformed by \eqref{ellipsoid_affine} as the interference between $\tilde{\mathbf{s}}_i$ and the unit sphere centered at origin, as shown in Fig. \ref{lab_ellipsoid}. Thus, replacing $\mathbf{s}_i, {}^0\mathbf{r}_{A_{is}M_j}$, ${}^0\mathbf{r}_{A_{ie}M_j}$ in \eqref{cond_pt_seg_rayBased} by $\tilde{\mathbf{s}}_i$, ${}^0\tilde{\mathbf{r}}_{A_{is}P_{0}} = - {}^0\tilde{\mathbf{r}}_{P_{0}A_{is}}, {}^0\tilde{\mathbf{r}}_{A_{ie}P_{0}} = - {}^0\tilde{\mathbf{r}}_{P_{0}A_{ie}}$ and $\epsilon_r = 1$, the ray-based IFW between cable segments and ellipsoids can be obtained in the similar methods described in Sec. \ref{sphere}.

\subsection{Conic Obstacles}
For the intersection of segments and cones, the conservative ray-based IFW is determined.

\begin{figure}
    \centering
    \includegraphics[width=0.6\textwidth]{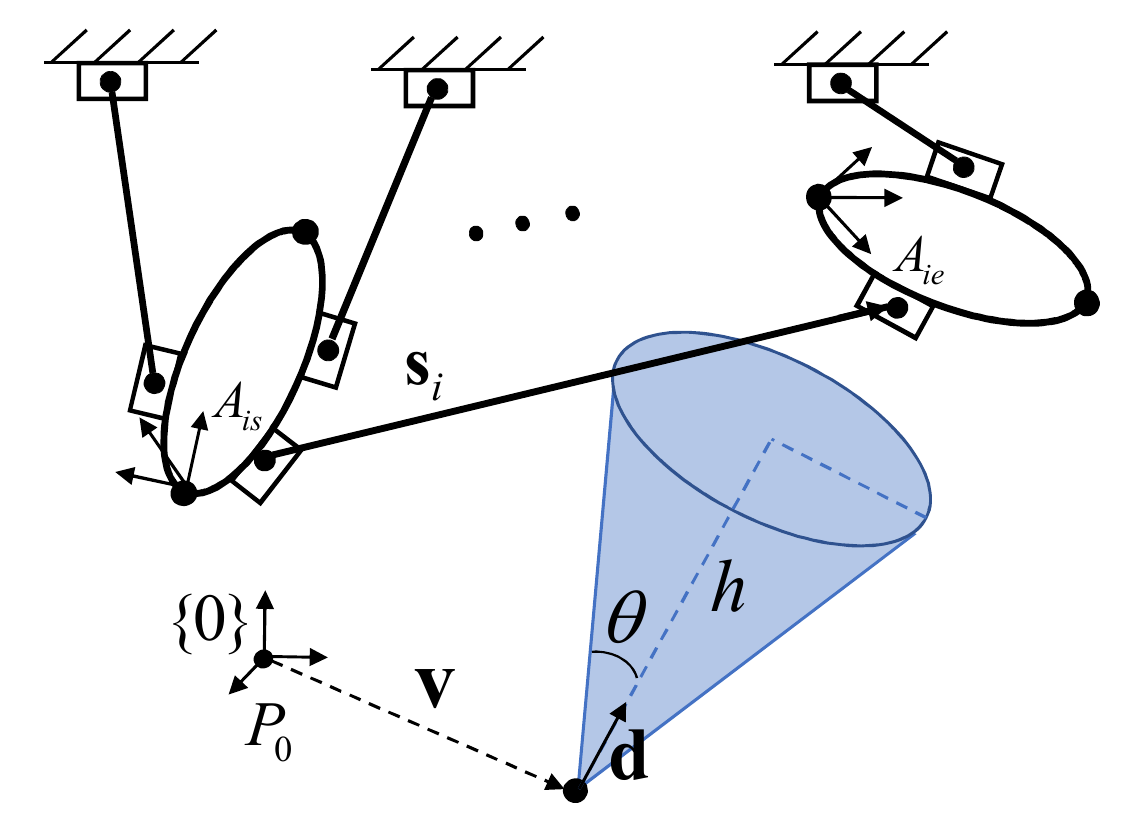}
    \caption{The conic obstacle}
    \label{lab_cone}
\end{figure}

In Fig. \ref{lab_cone}, the cone defined by the vertex $\mathbf{v}$, axis direction $\mathbf{d}$, height $h$ and the angle $\theta$ between axis and boundary, can be expressed as a set of points $\mathbf{x}$ such that
\begin{align}
    (\mathbf{x}-\mathbf{v})^T\mathbf{M}(\mathbf{x}-\mathbf{v}) = 0 \label{cond_cone3}
\end{align}
where $\mathbf{M} = \mathbf{d}\mathbf{d}^T - (\cos\theta)^2 I$. Then substituting $\mathbf{x} = {}^0\mathbf{r}_{P_{0}A_{is}} + t \mathbf{s}_i$, which indicates a point along the line containing $\mathbf{s}_i$, into the \eqref{cond_cone3}, yields
\begin{align}\label{eq_cond_cone}
    c_2 t^2 + 2 c_1 t + c_0 = 0
\end{align}
where $\mathbf{\Delta} = {}^0\mathbf{r}_{P_{0}A_{is}} - \mathbf{v}$, $c_0 = \mathbf{\Delta}^T \mathbf{M} \mathbf{\Delta}$, $c_1 = \mathbf{s}_i^T \mathbf{M} \mathbf{\Delta}$ and $c_2 = \mathbf{s}_i^T \mathbf{M} \mathbf{s}_i$. 

If $c_2 \neq 0$, then the  discriminant is $\delta = c_1^2 - c_2c_0$. When $\delta < 0$, then \eqref{eq_cond_cone} does not have the real-valued roots, which means the line containing $\mathbf{s}_i$ does not intersect the cone. So neither segment $\mathbf{s}_i$ can contact with the cone. However if $\delta > 0$, the solutions of \eqref{eq_cond_cone} is $t = (-c_1 \pm \sqrt{\delta})/c_2$
containing the square root, whereas $t$ can not be transformed as the polynomial equation of $u$. Therefore, ray-based IFW of segments and cones can be determined conservatively by the intersection of lines and cones utilizing the ray-based method.  

It should note that the $\mathbf{M}$ and $\mathbf{v}$ are constant and ${}^0\mathbf{r}_{P_{0}A_{is}}$ can take the form of \eqref{sVectMat}, thereby the $\mathbf{\Delta}$ can be transformed in \eqref{sVectMat} as well. Therefore, the coefficients $c_2$, $c_1$, $c_0$ in \eqref{eq_cond_cone} could be rewritten as a set of polynomial equations of $u$. As such, after substituting $c_2(u), c_1(u), c_0(u)$ into $\delta < 0$, the conservative ray-based IFW between segments and the cones is given by
\begin{align}
    W_{free} = \{u: c_2(u) > 0, \delta(u) < 0\} \cup \{u: c_2(u) < 0, \delta(u) < 0\}
\end{align}

\section{Ray-based IFW for Cables-Links}\label{subsec:cable-links}
In this subsection, the interference between cable segments and links (or the end-effector) is considered.
If there are cylindrical links, the $W_{free}$ can be generated in the similar method presented in Sec. \ref{cylinder}. Generally, the links can be approximated as polyhedra, thereby the method proposed in Sec. \ref{polyhedra} can be employed to determine the ray-based IFW accordingly.

\section{Simulation and Conclusion}

\subsection{The 6-DoF CDPR}\label{ssec:caseCDPR}
\begin{figure*}[htbp]
	\centering
	\subfigure[Case 1: CDPR]{
		\label{cs1_sub1_Model}
		\includegraphics[width=0.3\textwidth]{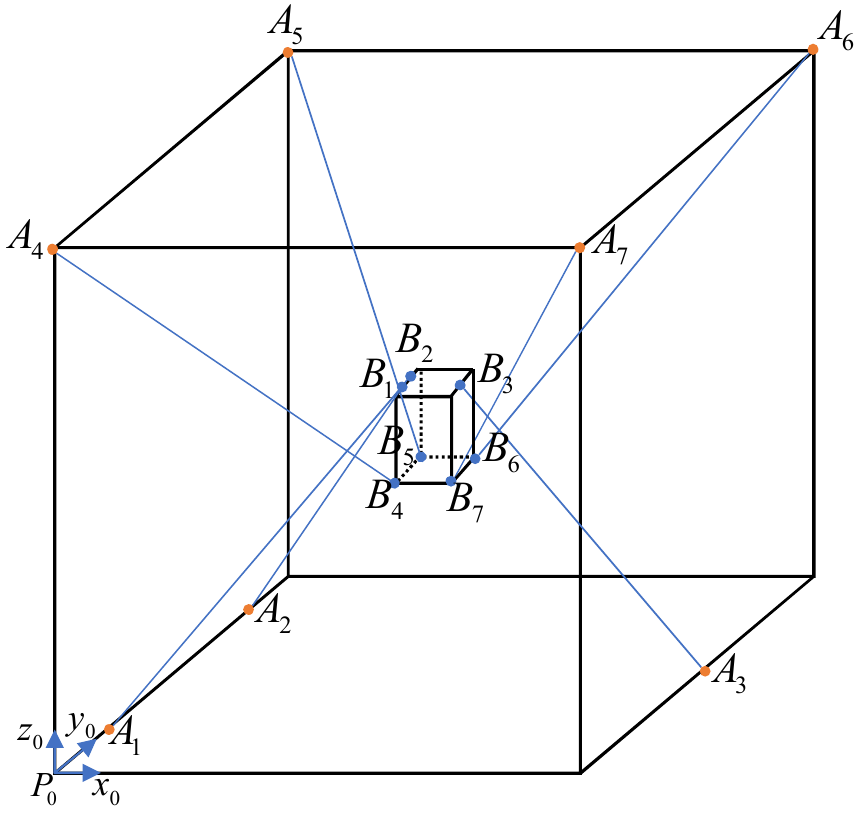}}~\hfil
	\subfigure[Case 2: CDPR+box-shaped obstacle]{
		\label{cs1_sub2_Model}
		\includegraphics[width=0.3\textwidth]{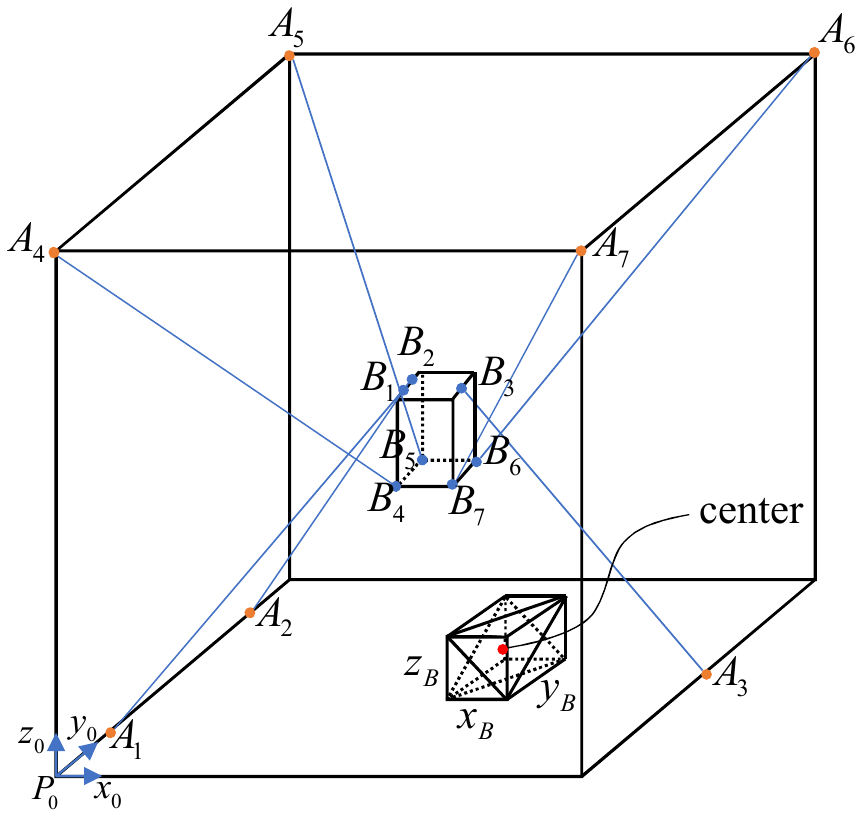}}~\hfil
	\subfigure[Case 3: CDPR+tree]{
		\label{cs1_sub3_Model}
		\includegraphics[width=0.3\textwidth]{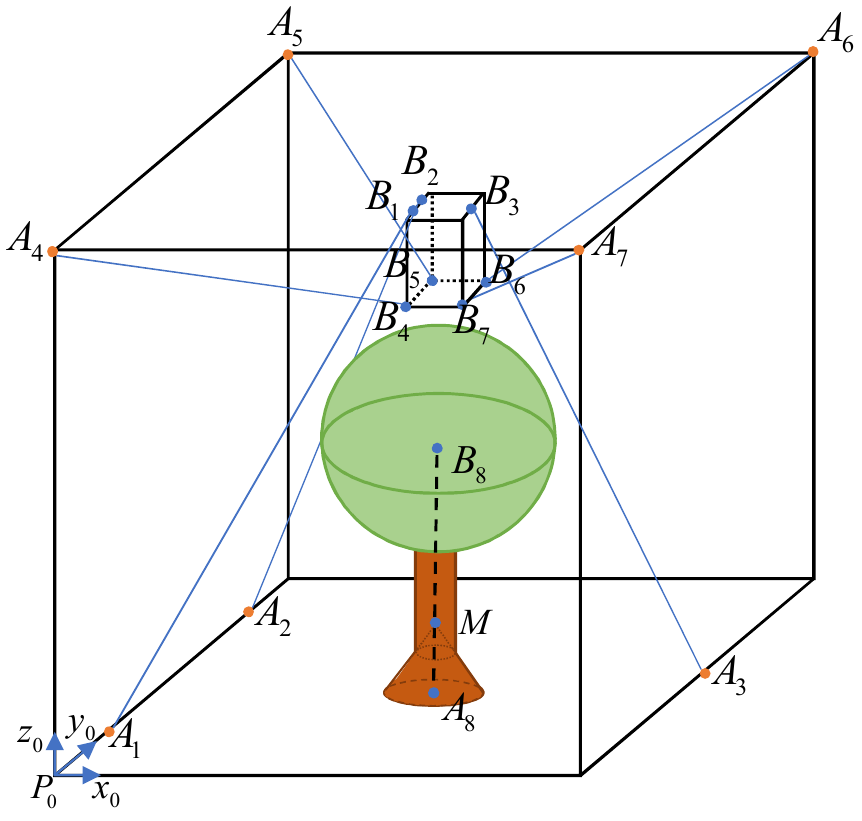}}~\vfil
	\subfigure[ConstOri-IFW for CDPR]{
		\label{cs1_sub1}
		\includegraphics[width=0.3\textwidth]{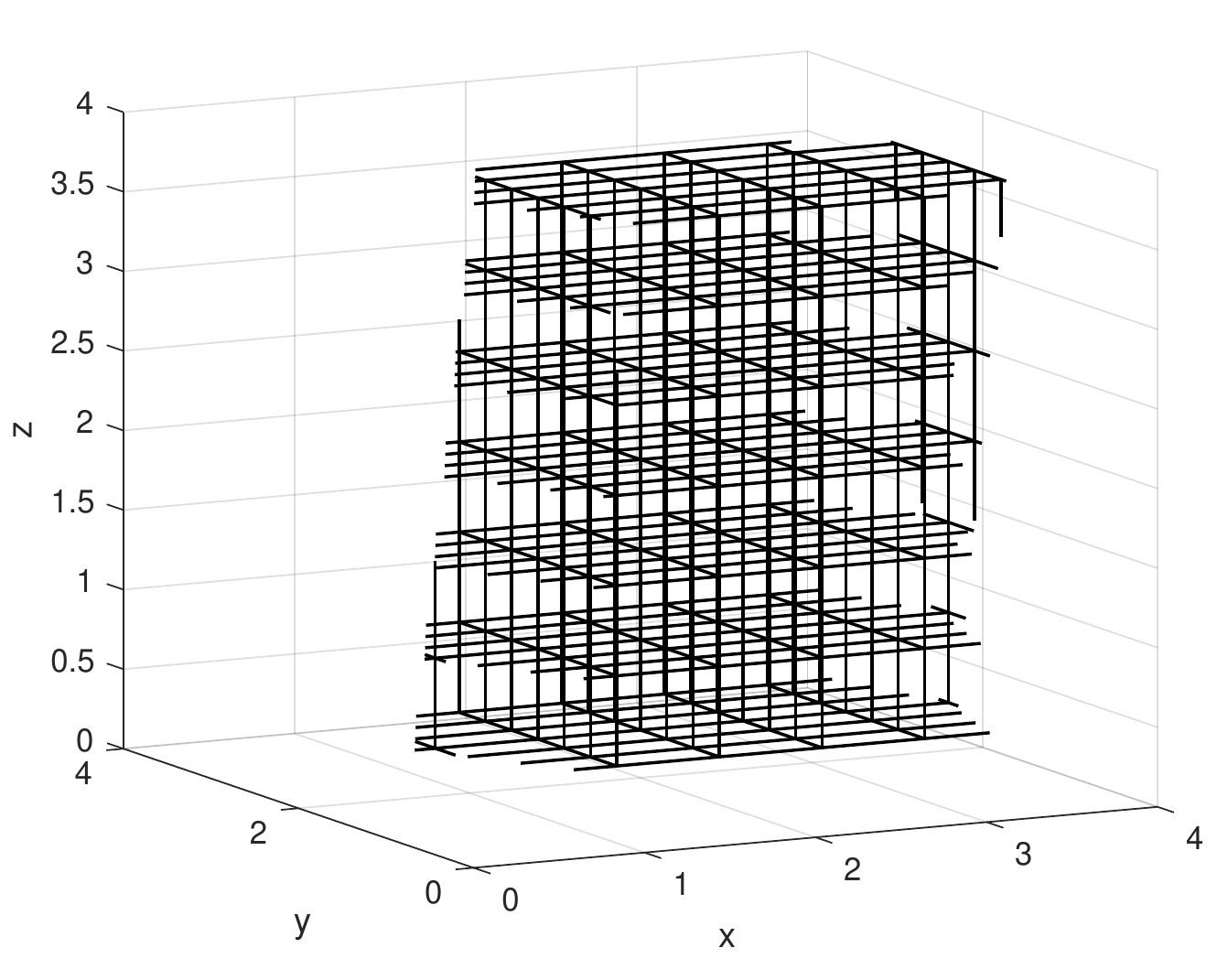}}~\hfil
	\subfigure[ConstOri-IFW for box-shaped obstacle]{
		\label{cs1_sub2}
		\includegraphics[width=0.3\textwidth]{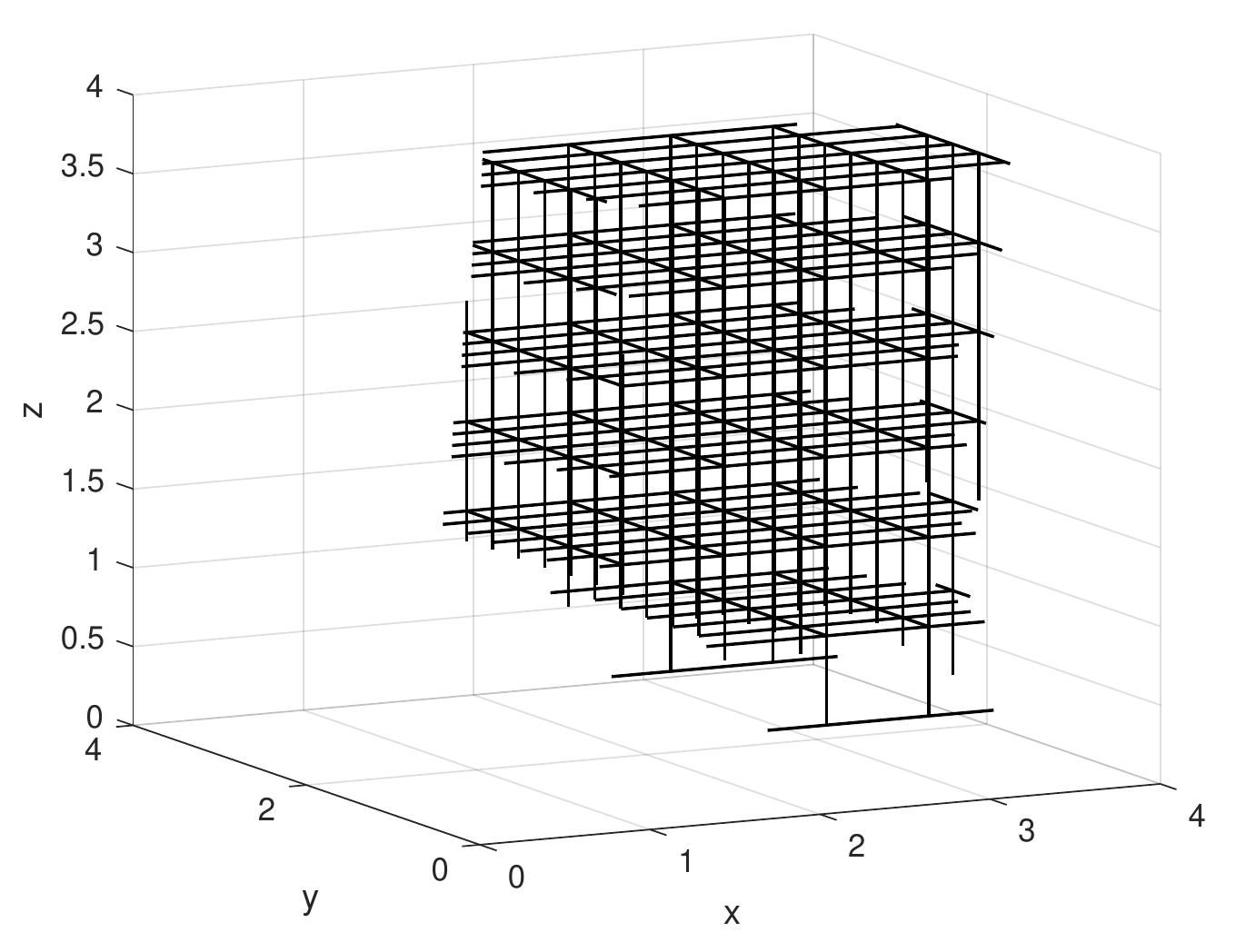}}~\hfil
	\subfigure[ConstOri-IFW for tree]{
		\label{cs1_sub3}
		\includegraphics[width=0.3\textwidth]{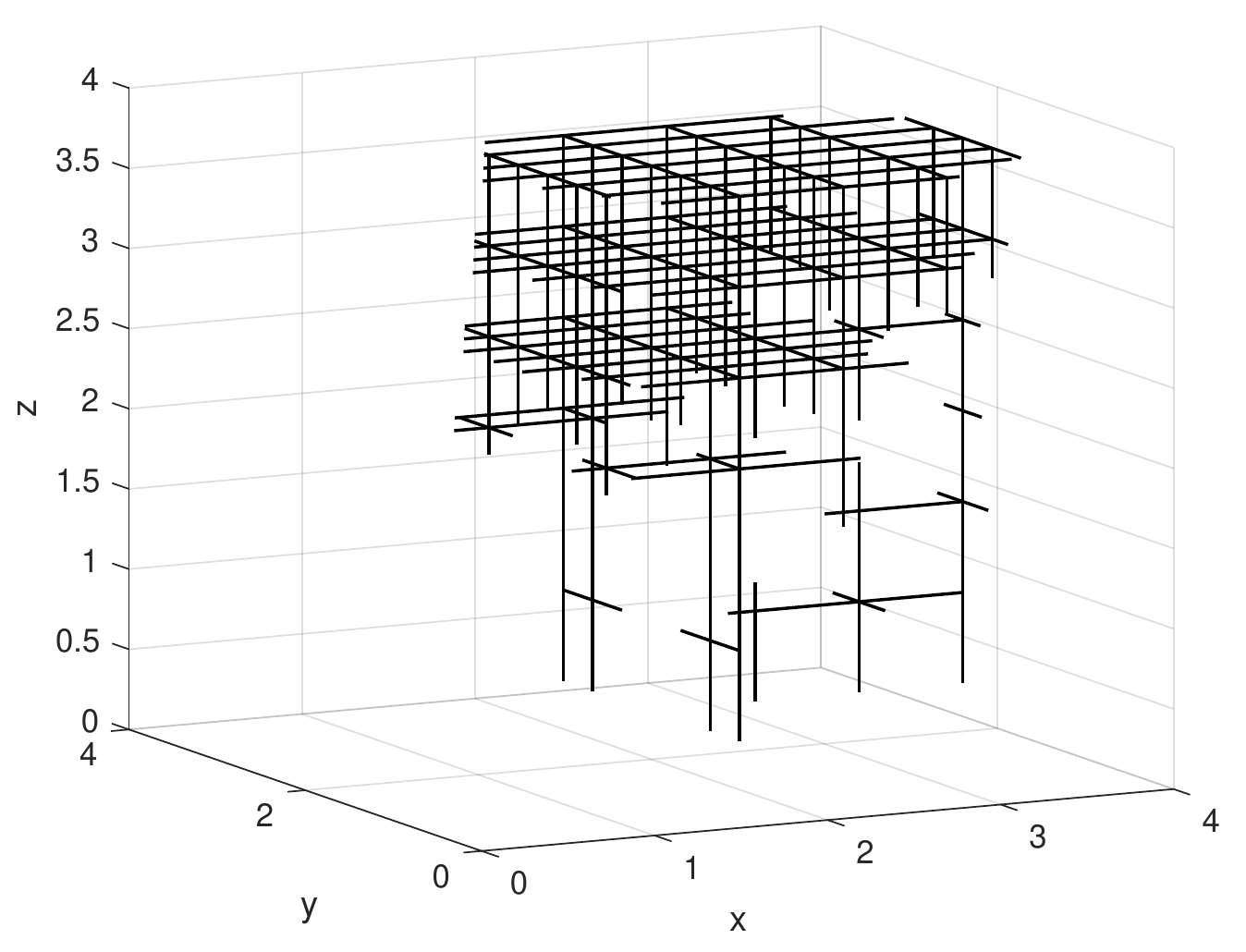}}
	\caption{CDPR with different objects and constant orientation IFW}
	\label{model_constIFW}
\end{figure*}

The ray-based IFW of a spatial CDPR with 6 DoFs, i.e., $\mathbf{q} = [x, y, z, \alpha, \beta, \gamma]$ is first studied. As shown in Fig. \ref{model_constIFW}, it is composed of a floating end-effector, called frame $\{1\}$, with 7 cables attached to the based, say frame $\{0\}$. The cable distal positions are given in Table \ref{case1_CDPR} and cable diameter is $r_{diameter} = 0.02$.
For the purpose of ease 3D visualization, the end-effector is assumed to be constant orientation, where $\alpha = \beta = \gamma = 0$. The ranges of translation DoFs are $x \in [0.2, 3.8], y \in [1.1, 2.9], z \in [0.3, 3.7]$ and they are selected to be discretized by 7 steps each. 

The Fig.~\ref{cs1_sub1} demonstrates the constant-orientation IFW between cables themselves and it is observed that the workspace with low $x$ values is affected seriously since the collision between cables $1$, $2$, $4$ and $5$ will reduce feasible workspace more than interference between cable $3$ and cables $6$, $7$.

With the same CDPR and discretization conditions, one rectangular solid is placed, as shown in Fig.~\ref{cs1_sub2_Model}. Here it is approximated by 12 triangles and its position and parameters are displayed in Table~\ref{case1_CDPR}. The resulting ray-based IFW for cable-cable and cable-obstacle is shown in Fig.~\ref{cs1_sub2}. Compared with IFW for cable interfering in Fig.~\ref{cs1_sub1}, it is obvious that a majority part of the proposed workspace with low $z$ becomes infeasible by virtue of collision between cables and the obstacle at bottom.

\begin{figure}[htbp]
	\centering
	\subfigure[Ray-based IFW]{
		\label{cs1_sub2_EE}
		\includegraphics[width=0.4\textwidth]{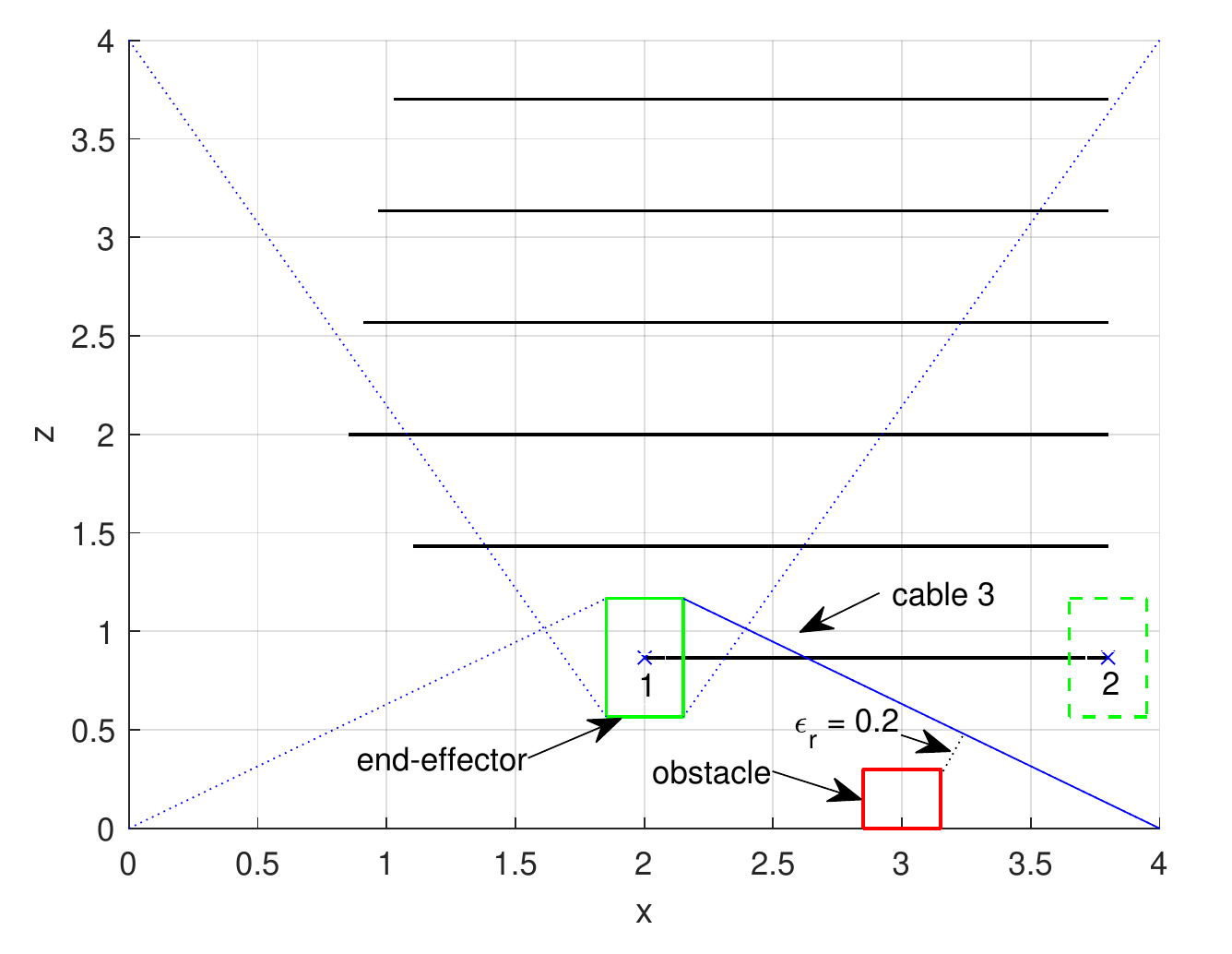}}~\hfil
	\subfigure[Point-wise-based IFW]{
		\label{pt:cs1_sub2_EE}
		\includegraphics[width=0.38\textwidth]{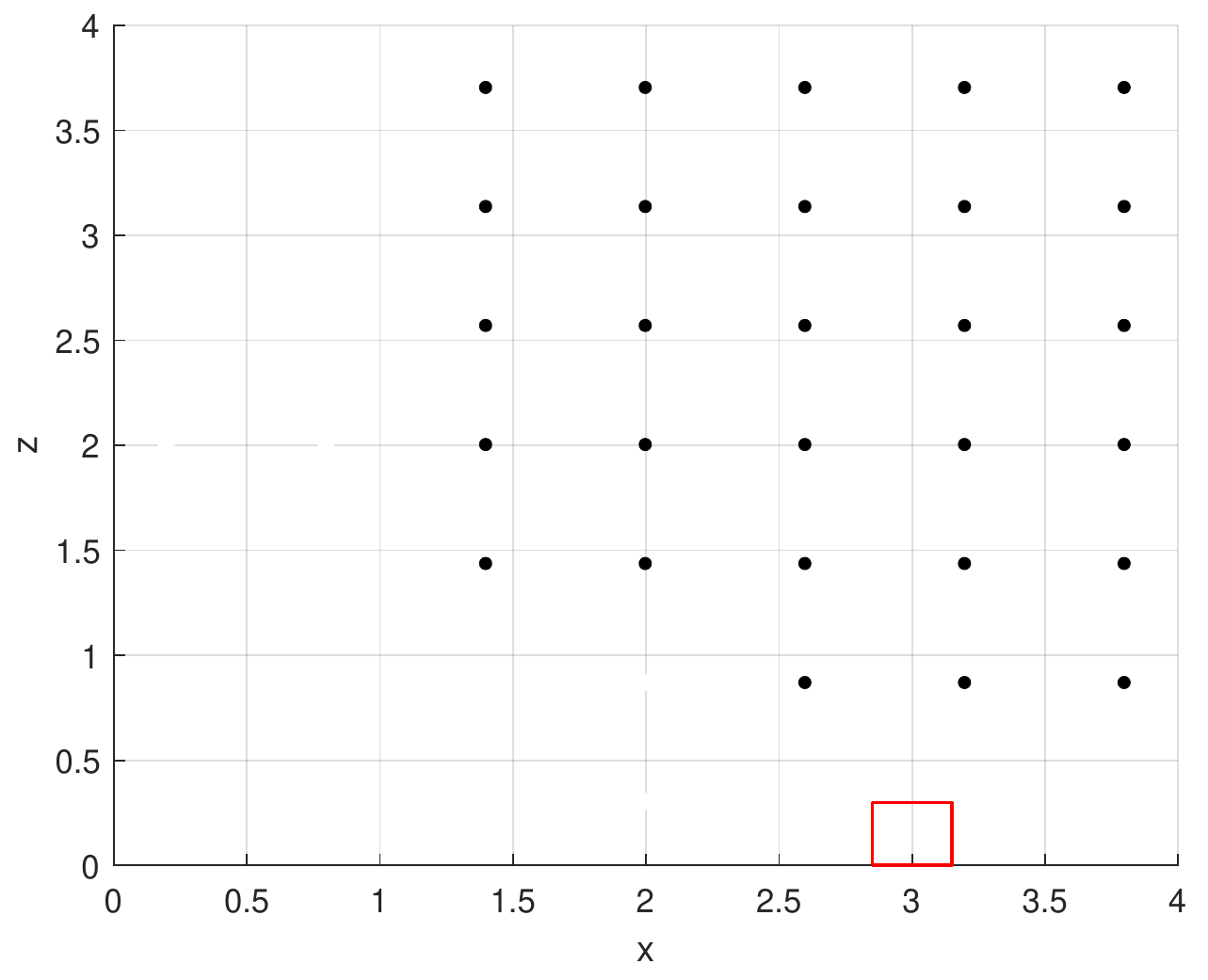}}
    \caption{$x -z $ cross sections of IFW for cable-cable and cable-obstacle with $y = 2, \alpha = \beta = \gamma = 0$ and $\epsilon_r = 0.2$. The coordinate of the point 1 can be explicitly determined as $[2.002, 2, 0.8667, 0, 0, 0]$ by the RBM.}
	\label{}
\end{figure}

In particular, $x-z$ cross-sections of above IFW with $ y =2 $ determined by the ray-based and point-wise approaches are detailed in Figs. \ref{cs1_sub2_EE} and \ref{pt:cs1_sub2_EE}, respectively. It is observed that the resulting workspaces are represented by rays and points. 

From the cross-sections of ray-based IFW, the continuity information between rays is known comparing with point-wise approach. It is shown in Fig.~\ref{cs1_sub2_EE} that the end-effector can move without any interference along the ray from point 1 to point 2. However, since at point 1 the minimum distance between cable $3$ and the obstacle reach to required minimum distance $\epsilon_r = 0.2$, the end-effector can not go further to the direction of $x$ decrease. But in Fig.~\ref{pt:cs1_sub2_EE}, only isolated points are demonstrated without any connectivity information from ray-based IFW. 

In Fig.~\ref{cs1_sub3_Model}, a tree-like object, that is more complicated than regular box-shaped object before, is located in the workspace. This obstacle consists of a sphere (a special case for the ellipsoid) centered at point $B_8$, a cylinder from $A_8$ to $B_8$ and a inverted cone started form point $M$, and specific parameters are shown in Table \ref{case1_CDPR}. The resulting IFW for cable-cable and cable-tree is shown in Fig. \ref{cs1_sub3}. 

Additionally, the computational cost for each case is provided in Table \ref{time_consuming}, where all results are analyzed in Windows 10 computer with Intel(R) Core(TM) i7-6500U CPU. The computational efficiency of the ray-based method (RBM) is directly observed between the first column and second column comparing with the point-wise method. 

Furthermore, with the growth of discretization steps $\tau$ for each DoF, the time consumed from the point-wise method increases faster than that from the ray-based method. The reason is that the point-wise method has an algorithmic complexity of $\mathcal{O}(\tau^3)$, while for the RBM it is $\mathcal{O}(\tau^2)$. Therefore, it could say that the RBM has advantages of computing workspace especially for the finer discretization compared with the point-wise method. 

From columns 3 and 4 in the Table \ref{time_consuming}, it is also shown that even for a simple polyhedral object, it still costs more time than that of the complex object, which could be combined by cylinders, ellipsoids (spheres) and cones. That is because the algorithm needs traversing $7 \times 12$ pairs of cable segments and triangles for the case 2, but only $7 \times 3$ iterations between 7 cable segments and one point $B_8$, one segment $A_8B_8$ as well as one cone $M$ are needed for the case 3. As such, it can conclude that the RBM could handle interference between cable segments and polyhedral objects, but it is more efficient for ray-based IFW if objects can be approximated by geometric primitives: cylinders, ellipsoids and cones.
%

\begin{table}[ht]
	\caption{Structure Parameters ($ m $) of the CDPR and objects} 
	\centering 
	\begin{tabular}{c c c} 
		\hline\hline 
		Cable $ i $ & $ {}^0\mathbf{r}_{P_0A_i} $ & $ {}^1\mathbf{r}_{P_1B_i} $\\ 
		\hline 
		1 & $ (0,1,0)  $ & $ (-0.15,-0.1,0.3) $  \\ 
		2 & $ (0,3,0) $ & $ (-0.15,0.1,0.3) $  \\
		3 & $ (4,2,0) $& $ (0.15,0,0.3) $  \\
		4 & $ (0,0,4) $ & $ (-0.15,-0.2,-0.3) $  \\
		5 & $ (0,4,4) $ & $ (-0.15,0.2,-0.3) $  \\
		6 & $ (4,4,4) $ & $ (0.15,0.2,-0.3) $  \\
		7 & $ (4,0,4) $ & $ (0.15,-0.2,-0.3) $  \\
		\hline\hline
		Box-shaped Obstacle & Center: $(3, 2, 0.15) $ \\
		\hline
		$x_B = 0.3$ & $ y_B = 0.5$ & $ z_B = 0.3$\\
		\hline\hline
		Tree-shaped Obstacle \\
		\hline
		$A_8 = (2, 2, 0)$ & $B_8 = (2, 2, 1.5)$  & $M = (2, 2, 0.3)$ \\
		$r_{sphere} = 0.4$ & $ r_{cylinder} = 0.12 $ & $\theta = \pi /6$\\
		\hline
	\end{tabular}
	\label{case1_CDPR} 
\end{table}

%
\begin{table}[ht]
	\caption{Computation Time (s) in generating the workspace} 
	\centering 
    \begin{tabular}{c c c c c}
    \hline\hline
    \multirow{2}{*}{Steps} & \multicolumn{1}{c}{Point-wise method} & \multicolumn{3}{c}{Ray-based method} \\   \cline{3-5}            & Case 1 & Case 1 & Case 2 & Case 3 \\    \hline
     $20$   &   89.1323 &  87.0698 &  255.5473 & 111.3884 \\       
     $30$   &  288.4506	&  187.0063 & 575.4497 & 248.1464 \\  
     $40$   &  677.2744	& 330.3755 & 1011.9294 & 465.7290 \\
    \hline
	\end{tabular}
	\label{time_consuming}  
\end{table}
	
%


\subsection{The 4-DoF MCDR}
\begin{figure}[htbp]
	\centering
	\includegraphics[width=0.55\textwidth]{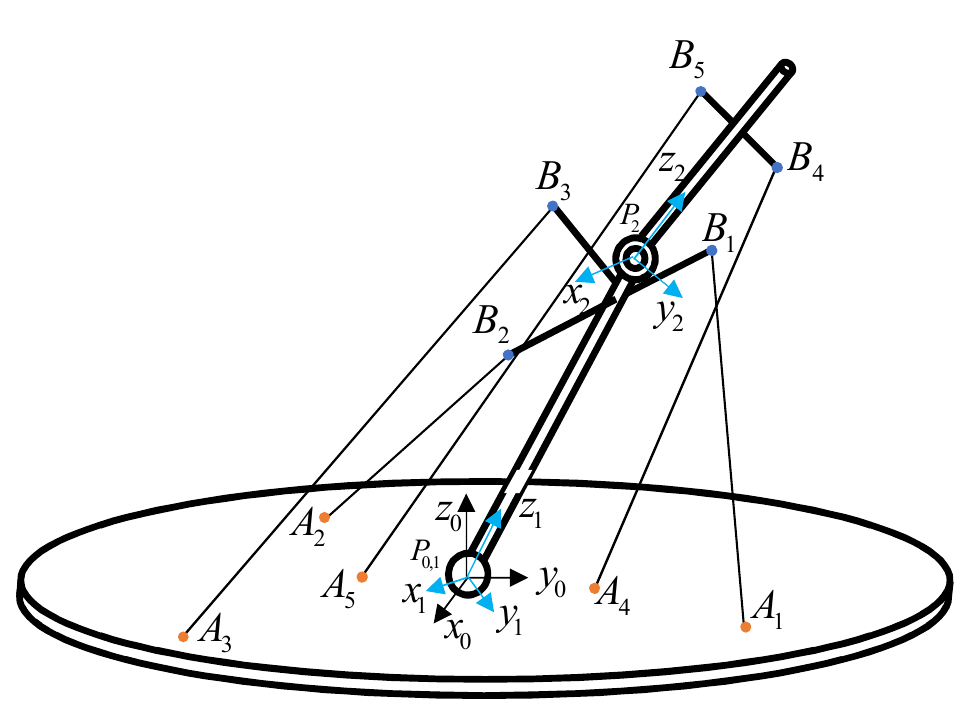}
	\caption{Schematic of the 4-DoF MCDR with one ball joint and one revolute joint. The joint space is defined by $ q\ = [\alpha, \beta, \gamma, \theta]^T $.}
	\label{case2_fig}
\end{figure}

A two-link MCDR with 5 cables is shown in Fig.~\ref{case2_fig}, where Link 1 (frame $\{1\}$) is connected to the base (frame $\{0\}$) by a spherical joint, followed by the link 2 (frame $\{2\}$) through a revolute joint. The generalized coordinates are $\mathbf{q} = [\alpha, \beta, \gamma, \theta]$, in which $\alpha, \beta, \gamma$ indicate the Euler angles of the spherical joint and $\theta$ represents the relative angle of the revolute joint. The structure parameters are given in Table~\ref{table:case2_para}, where the coordinates of points $A_1$ to $A_5$ are shown in the frame $\{0\}$, $B_1$ to $B_3$ are expressed in the frame $\{1\}$ and $B_4$, $B_5$ are formulated in frame $\{2\}$, respectively.

\begin{figure*}
    \centering
    \subfigure[Ray-based IFW for cable-cable and cable-links ($\epsilon_r = 0.02$)]{
    \label{ffig:ifw}
    \includegraphics[width=0.4\textwidth]{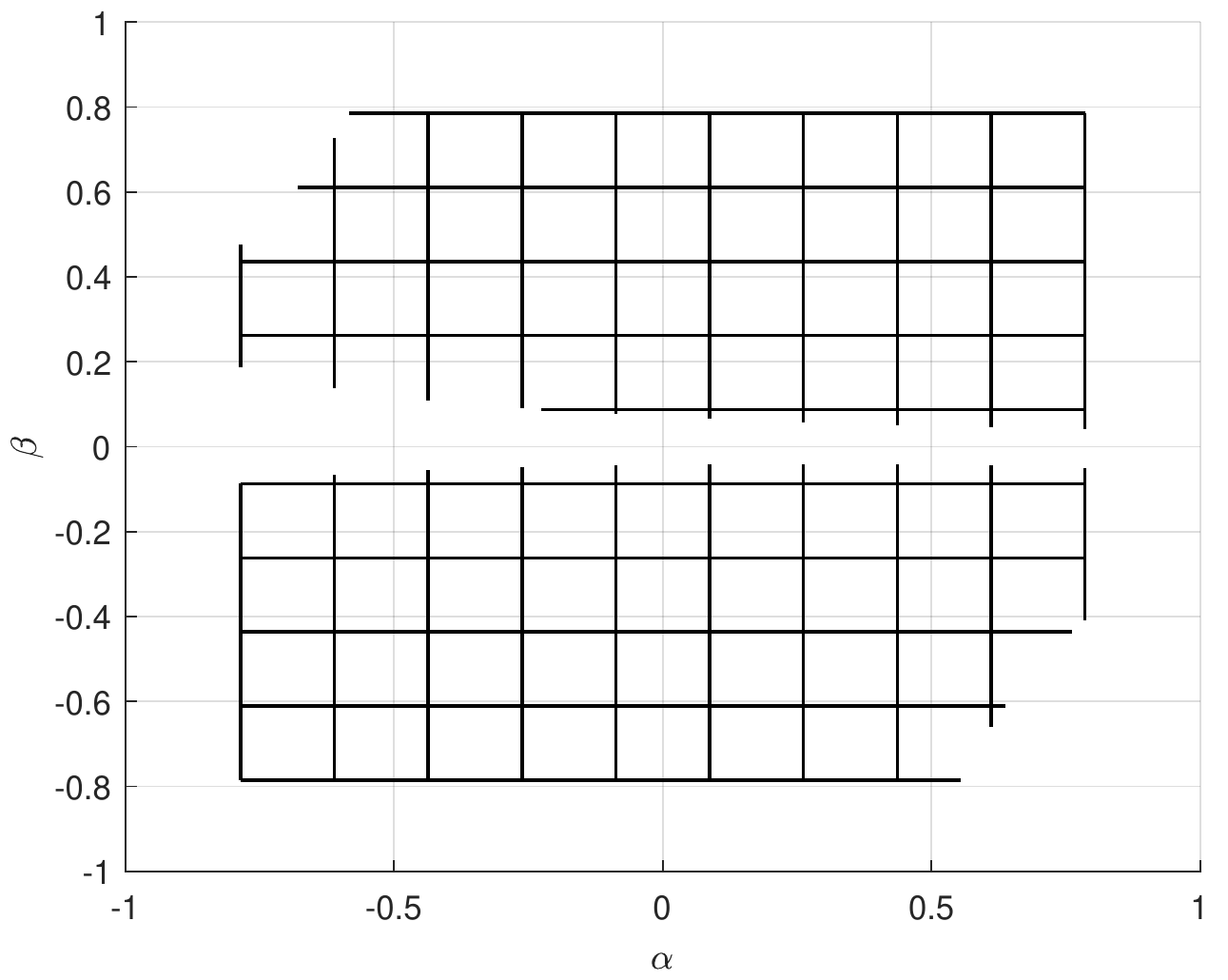}}~\hfil
	\subfigure[Ray-based WCW]{
    \label{ffig:wcw}
    \includegraphics[width=0.4\textwidth]{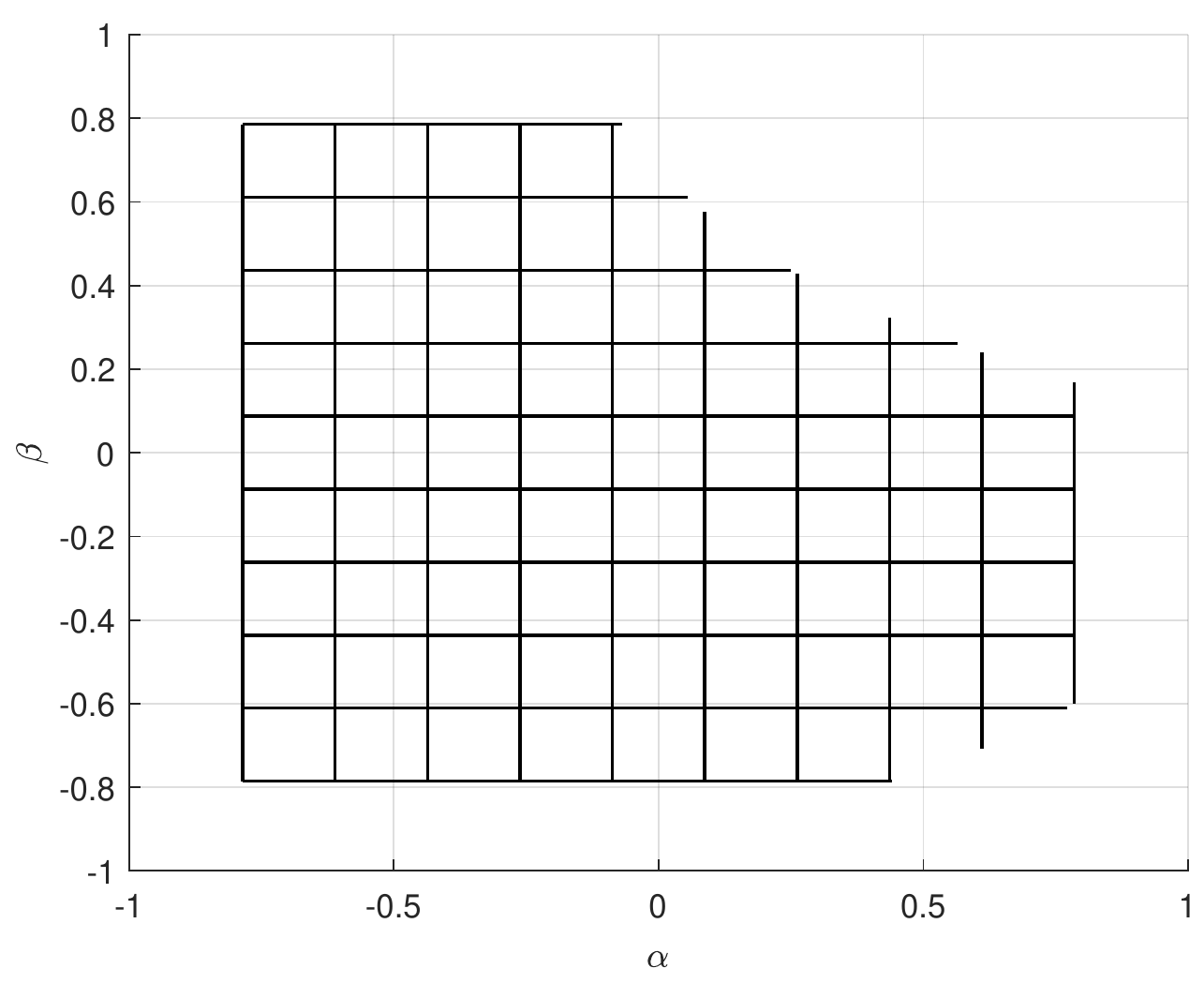}}~\vfil
	\subfigure[Ray-based workspace combining IFW and WCW]{
    \label{ffig:ifwcw}
    \includegraphics[width=0.4\textwidth]{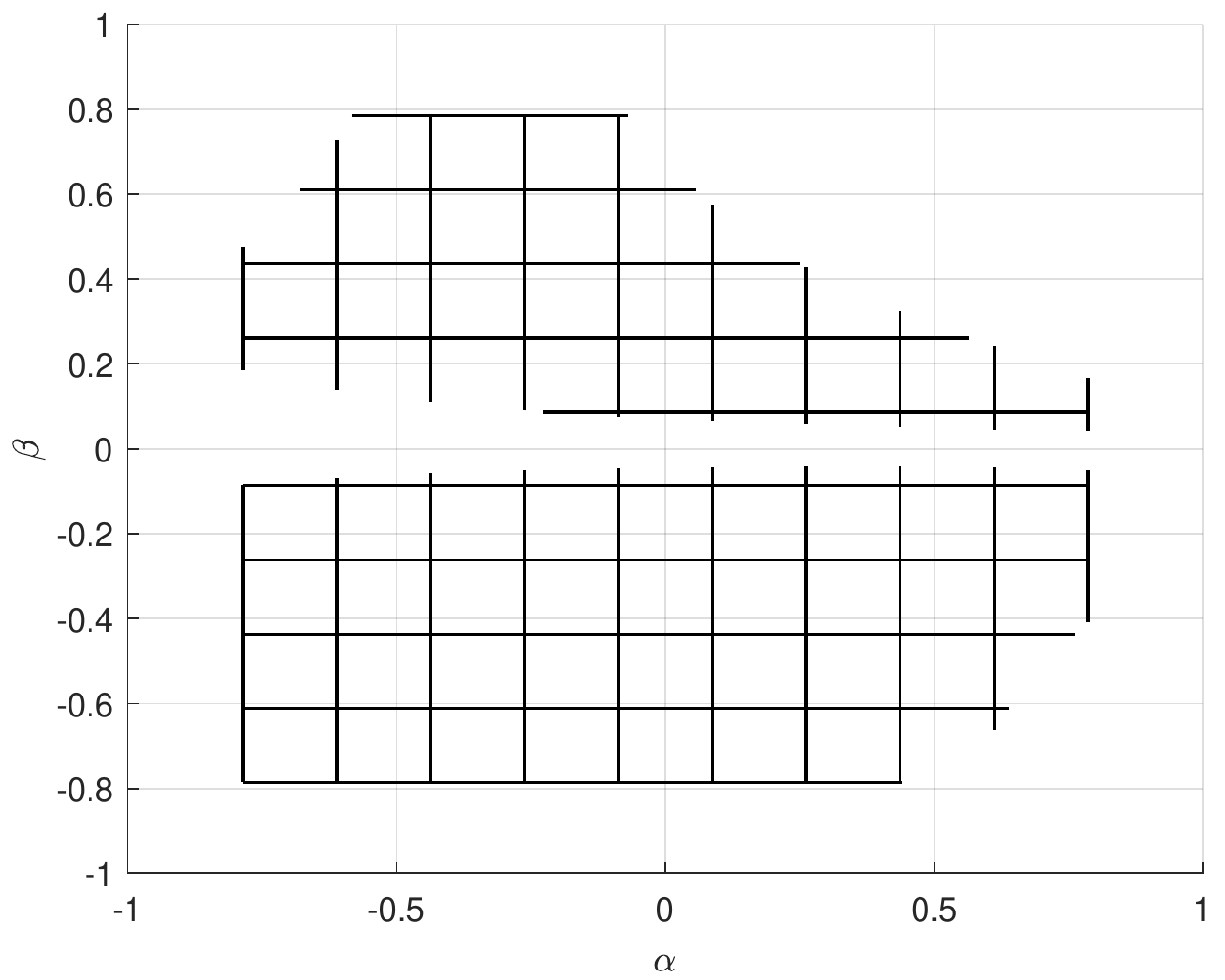}}~\hfil
    \subfigure[Point-wise-based workspace combining IFW and WCW]{
    \label{ffig:ptifwcw}
    \includegraphics[width=0.4\textwidth]{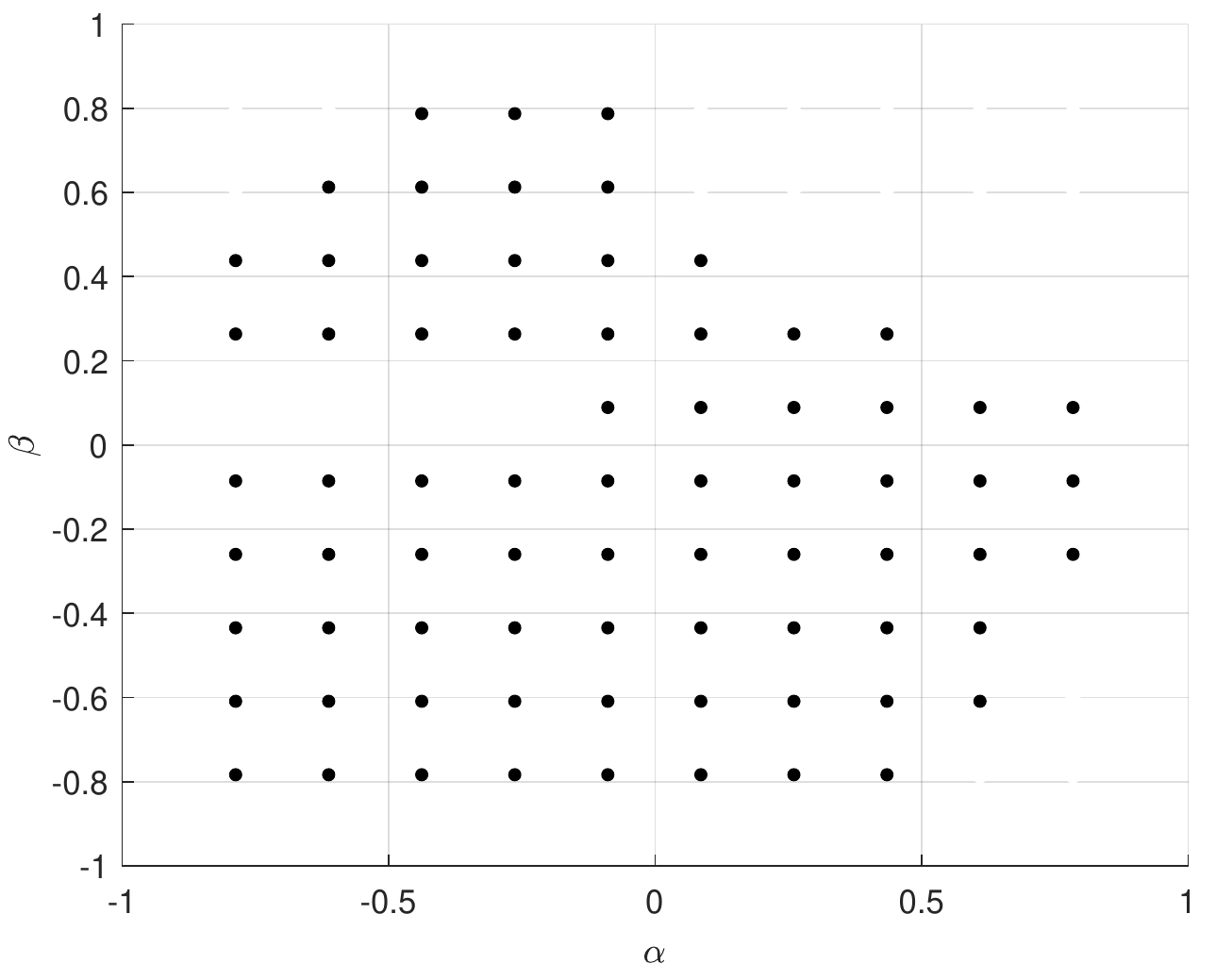}}~\hfil
    \caption{$ \alpha - \beta $ cross sections of the workspace for the MCDR with $\gamma = -1/12 \pi$, $\theta = 1/12 \pi$ and $\alpha, \beta \in [-1/4 \pi, 1/4 \pi] $. There are 10 steps discretized for $\alpha$ and $\beta$ each. }
    \label{fig:ifw_wcw}
\end{figure*}

The $\alpha - \beta$ cross sections of the ray-based IFW determined by the proposed method and the ray-based wrench closure workspace (WCW) generated from \cite{DL_wcw} are detailed in Figs.~\ref{ffig:ifw} and \ref{ffig:wcw}, respectively. Furthermore, it is convenient to merge two different types of workspace into one through intersection sets of corresponding rays, as demonstrated in Fig.~\ref{ffig:ifwcw}. It is easily and efficiently to use RBM for the workspace analysis under different workspace conditions by one generalized algorithm at the same time. The hybrid workspace not only provides the feasible working areas under numerous workspace constraints, but the connectivity information also holds inherently. For this reason, it is observed that there are two separate regions divided by a slight gap in Fig.~\ref{ffig:ifwcw}, while it is impossible to find that by the point-wise method under the same conditions in Fig.~\ref{ffig:ptifwcw}. 

\begin{figure*}[htbp]
	\centering
	\subfigure[Graph of ray-based IFW]{
		\label{ffig:cs2_Gifw}
		\includegraphics[width=0.4\textwidth]{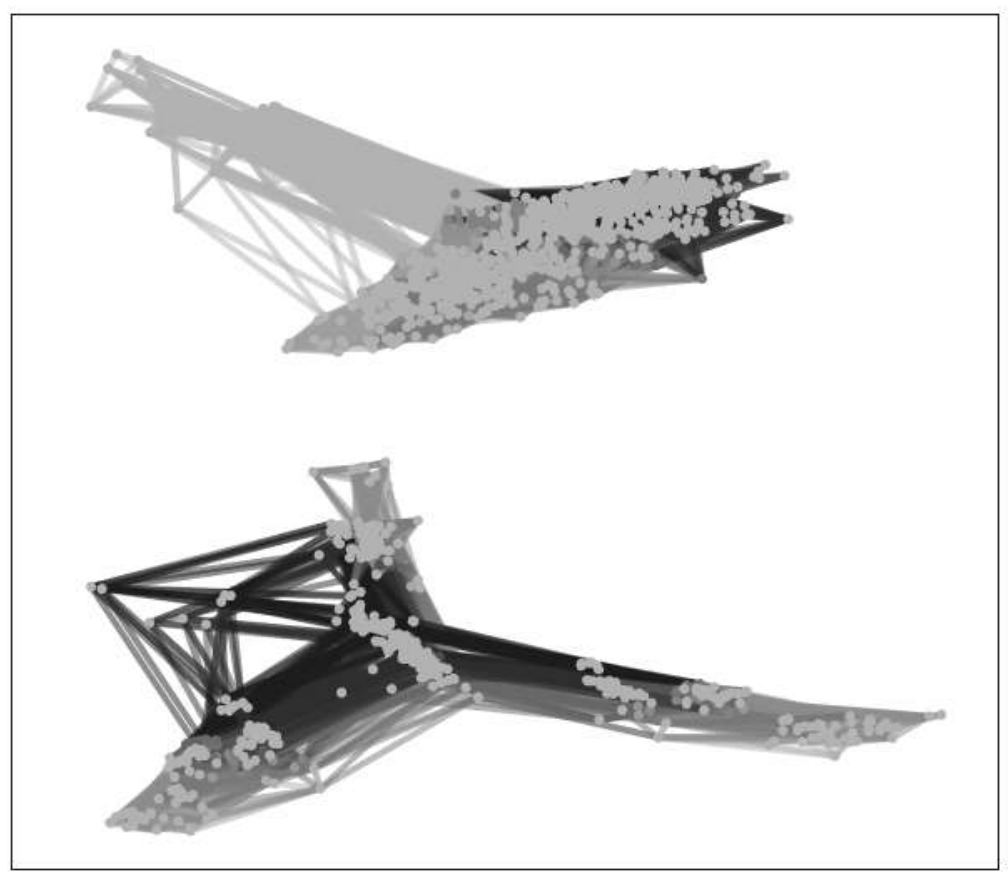}}~\hfil
	\subfigure[Graph of ray-based WCW]{
		\label{ffig:cs2_Gwcw}
		\includegraphics[width=0.4\textwidth]{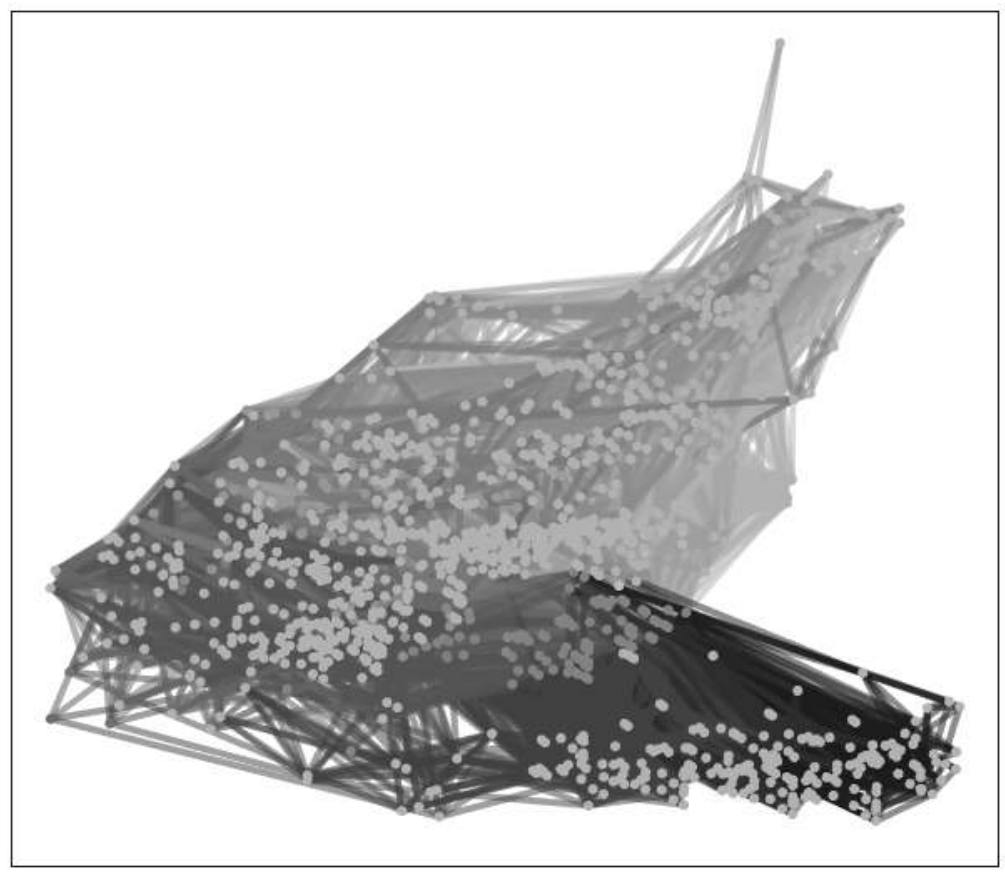}}~\vfil
	\subfigure[Graph of ray-based hybrid workspace combining IFW and WCW]{
		\label{ffig:cs2_GifWcw}
		\includegraphics[width=0.4\textwidth]{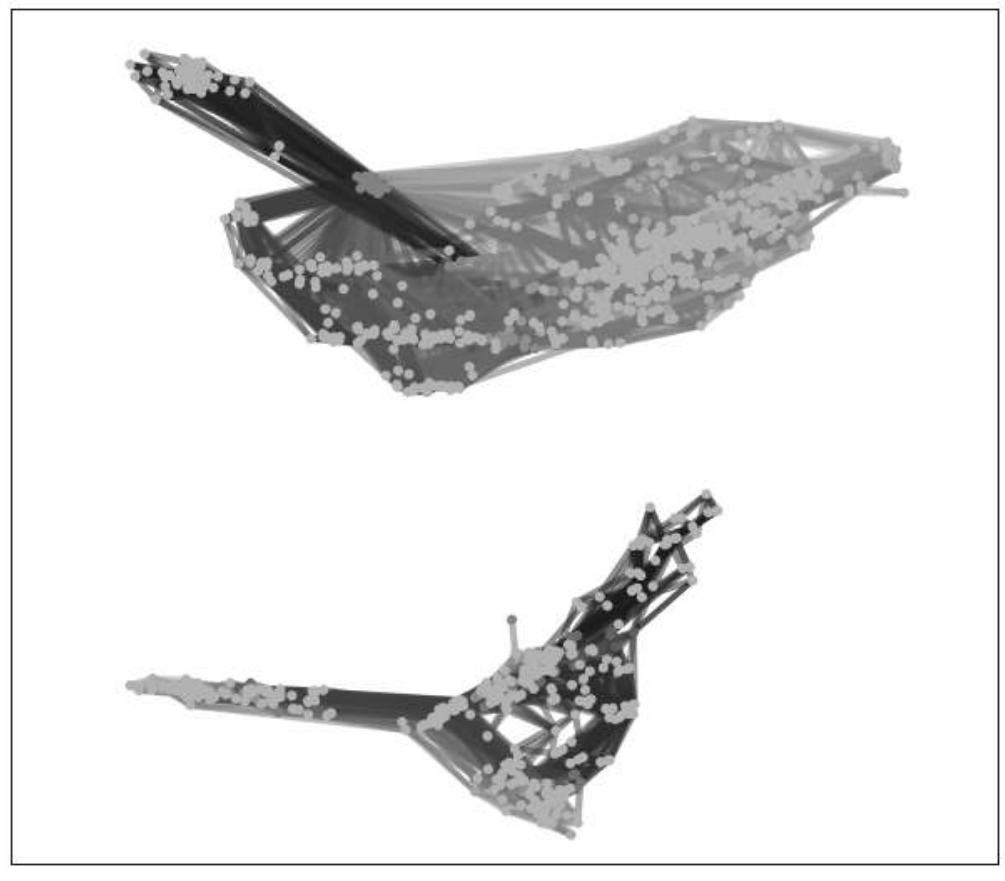}}
	\caption{Graph representation for the ray-based workspace based on \cite{ghasem}. }
	\label{cs2_graph_ifw_wcw}
\end{figure*}

Additionally, in order to present the whole workspace with 4 DoFs as well as keeping the connectivity information from the RMB, the graph representation based on \cite{ghasem} is employed and the results are given in Fig.~\ref{cs2_graph_ifw_wcw}. Here some insights can be observed globally. 
\begin{enumerate}
    \item In Fig.~\ref{ffig:cs2_Gifw}, there are two separate regions with similar size for IFW, which is consistent with the observation from the cross-section in Fig.~\ref{ffig:ifw}.   
    \item The ray-based WCW is displayed as one part in Fig.~\ref{ffig:cs2_Gwcw}, which means the proposed cable attachments have great chance to manipulate the MCDR within one working area if neglecting other workspace constraints.
    \item In fact, the hybrid workspace Fig.~\ref{ffig:cs2_GifWcw} becomes two disconnected parts for this MCDR, which can be verified in Fig.~\ref{ffig:ifwcw}
\end{enumerate}
%

\begin{table}[ht]
	\caption{Structure Parameters of the 4-DoF MCDR $( m ) $} 
	\centering 
	\begin{tabular}{c c c} 
		\hline\hline 
		Cable $ i $ & $ A_i $ & $ B_i $\\ 
		\hline 
		1 & $ (1,1,0)  $ & $ (-0.2121, 0.2121, 0.6) $  \\ 
		2 & $ (-1,-1,0) $& $ (0.2121, -0.2121, 0.6) $  \\
		3 & $ (1,-1,0) $ & $ (-0.3536, -0.3536, 0.6) $  \\
		4 & $ (0,0.3,0) $ & $ (0, 0.1, 0.4) $  \\
		5 & $ (0,-0.3,0) $ & $ (0, -0.1, 0.4) $  \\
		\hline 
	\end{tabular}
	\label{table:case2_para} 
\end{table}

\chapterend

\pdfoutput=1
\chapter{Ray-based Path Planning and Verification}



\section{Path Planning on the Ray-based Workspace}

\subsection{Initial Path}
\label{sec:segPath}
\begin{figure}
    \centering
    \includegraphics[width=0.7 \textwidth]{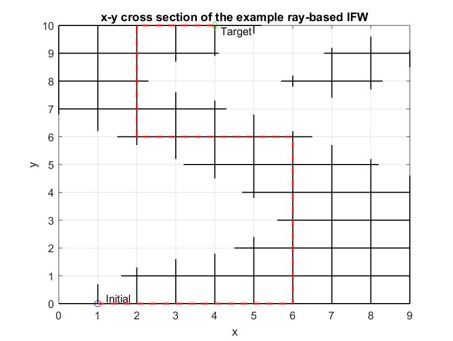}
    \caption{The 2D ray-based workspace}
    \label{fig:initialpath}
\end{figure}

In this section, the 2D ray-based IFW is taken as an example. Since the ray-based workspace contains the continuity information as introduced in above section, one path through the initial and target points along the rays is found and displayed as the red dash line in Fig.~\ref{fig:initialpath}. This path is called the \textit{initial path}. It is shown that the initial path is composed of multiple line segments and it is a feasible path satisfying interference free conditions since it strictly follows rays on the ray-based IFW generated by RBM.

However, it is obvious that the initial path is unnatural for robot motion due to turning points with $90^\circ$ between conjunctive segments. Furthermore, because of the acceleration and deceleration along each segment, it would be time-consuming and low efficiency if the end-effector follows the initial path. Therefore, the feasible, optimal and smooth path should be well planned.

\subsection{Proposed Planning Method}
\label{sec:pathplanning}
A new method for path planning based on the ray-based workspace is proposed in this section, where the A-star algorithm and B\'ezier curves are used.

\subsubsection{Step 1: using A-star algorithm}
\label{ssec:Astar}
\begin{figure}[htbp]
    \centering
    \includegraphics[width=0.7 \textwidth]{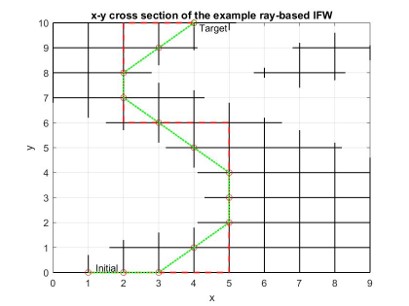}
    \caption{Path found by the A-star algorithm}
    \label{fig:Astar}
\end{figure}

First of all, intersecting points between rays in the ray-based workspace are selected as grid points, then using A-star algorithm as introduced in Sec.~\ref{sssec:astar}, the shortest path from the initial point to the target point is found as shown as the green path in Fig.~\ref{fig:Astar}. It is shown that the green path is still composed of multiple segments, even if it is more 'smooth' than the initial path. Therefore, the B\'ezier curve is employed to generate smooth path and is introduced in the following section.

\subsubsection{Step 2: using B\'ezier curve}
\label{ssec:bezier}
\begin{figure}[htbp]
    \centering
    \includegraphics[width=0.7 \textwidth]{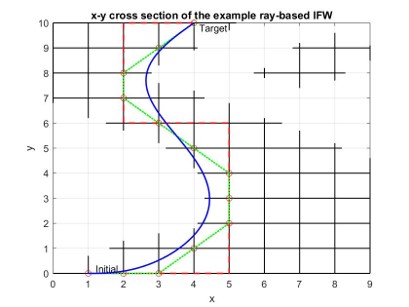}
    \caption{Smooth path fitted by B\'ezier curve}
    \label{fig:bezierCurve}
\end{figure}

In this section, the B\'ezier curve is used to yield a smooth path on the basis of the path obtained by the A-star algorithm. Here points on the green path found by the A-star algorithm, shown as red circles in Fig.~\ref{fig:Astar},  are selected as control points for the  B\'ezier curve. So the smooth path fitted by the B\'ezier curve is demonstrated as blue curve in Fig.~\ref{fig:bezierCurve}.

However, the following question is whether the blue path satisfies the interference free conditions. Since it can only be guaranteed that interference free conditions hold on rays, the feasibility of the proposed blue curve should be verified. Thus, the path verification is presented in the following section.

\section{Proposed Path}

\subsection{Path in Translation}
\label{sec:ppTrans}

\begin{figure*}
    \centering
    \subfigure[Linear path]{
    \label{ffig:line}
    \includegraphics[width=0.4\textwidth]{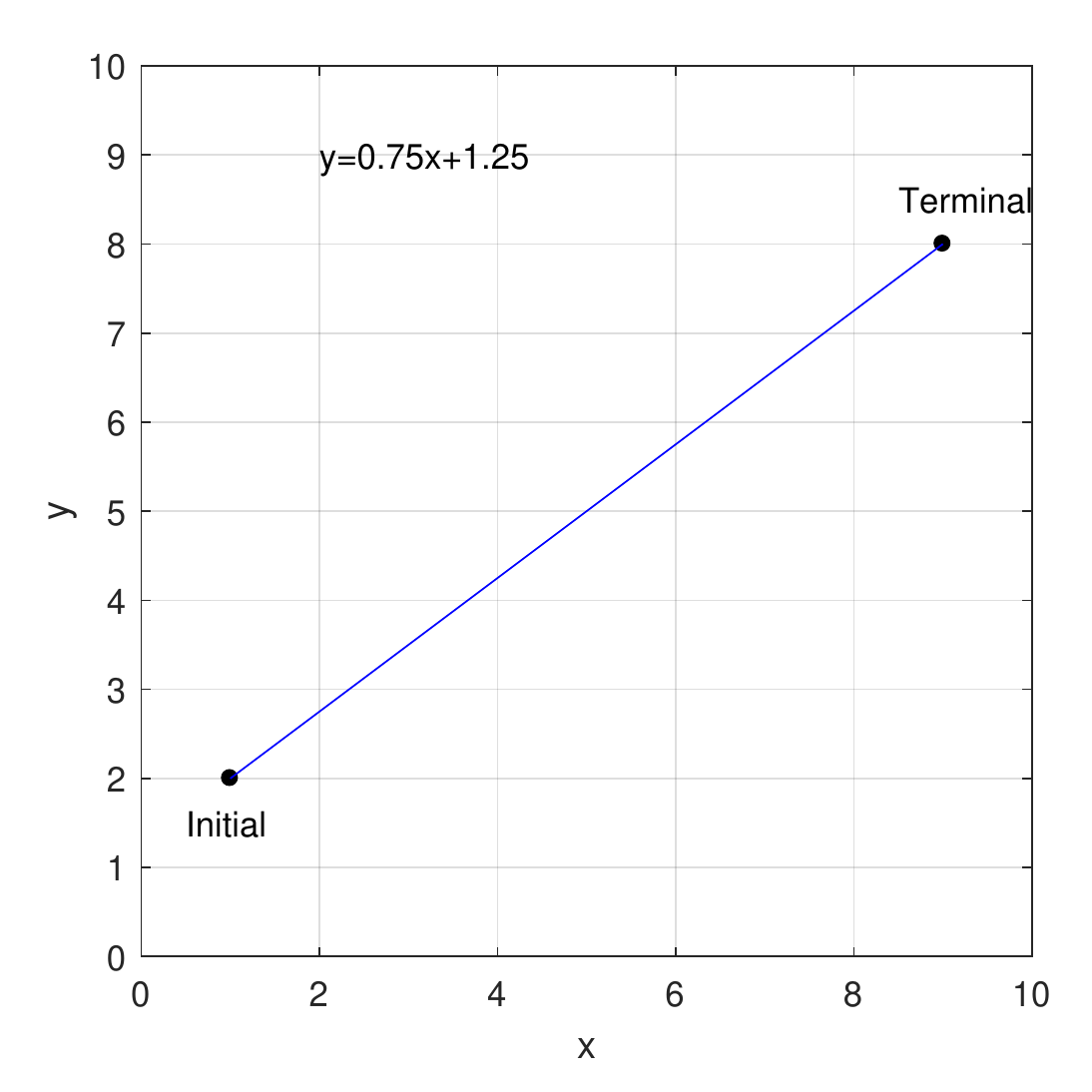}}~\hfil
	\subfigure[Quadratic polynomial path]{
    \label{ffig:quad}
    \includegraphics[width=0.4\textwidth]{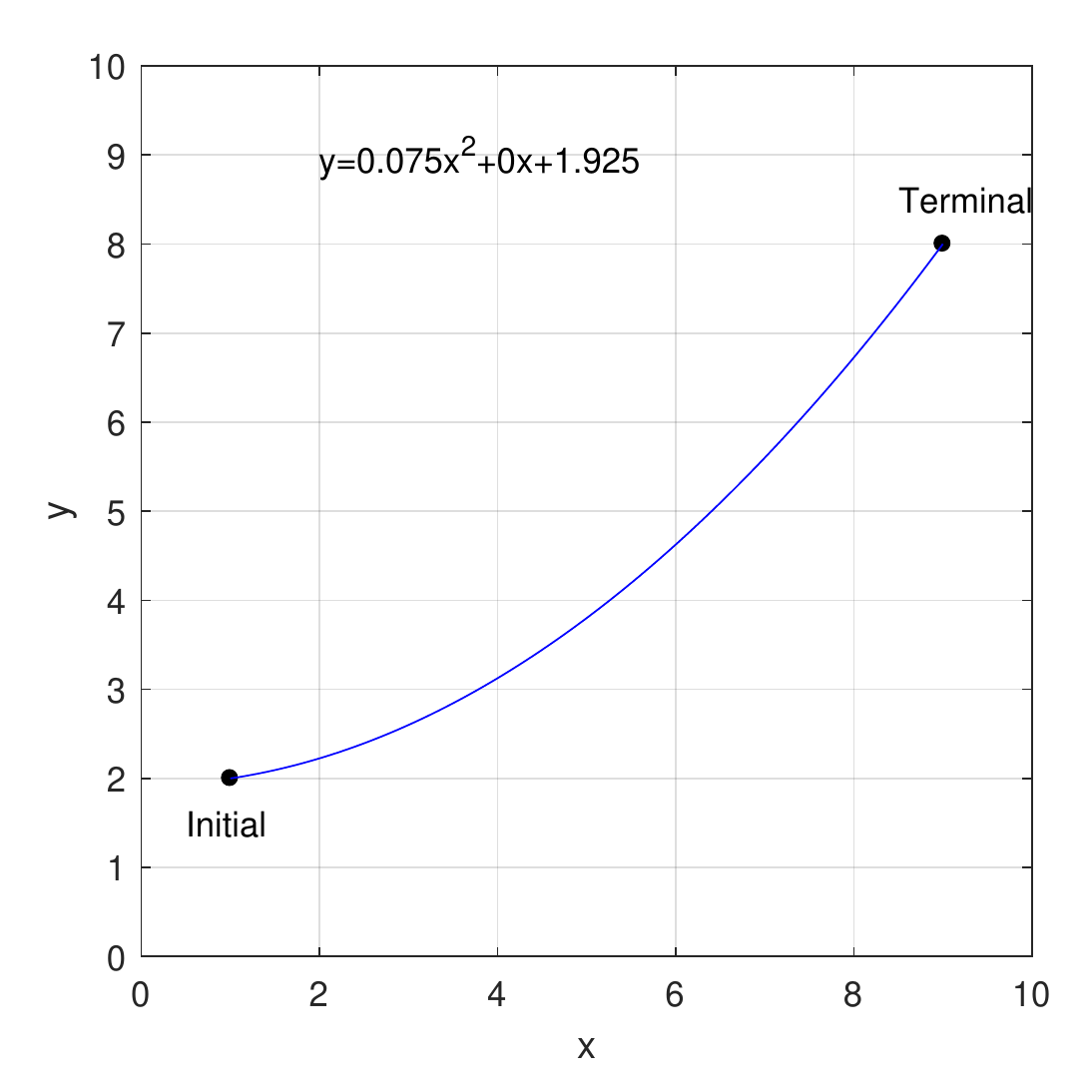}}~\vfil
	\subfigure[Cubic polynomial path]{
    \label{ffig:cubi}
    \includegraphics[width=0.4\textwidth]{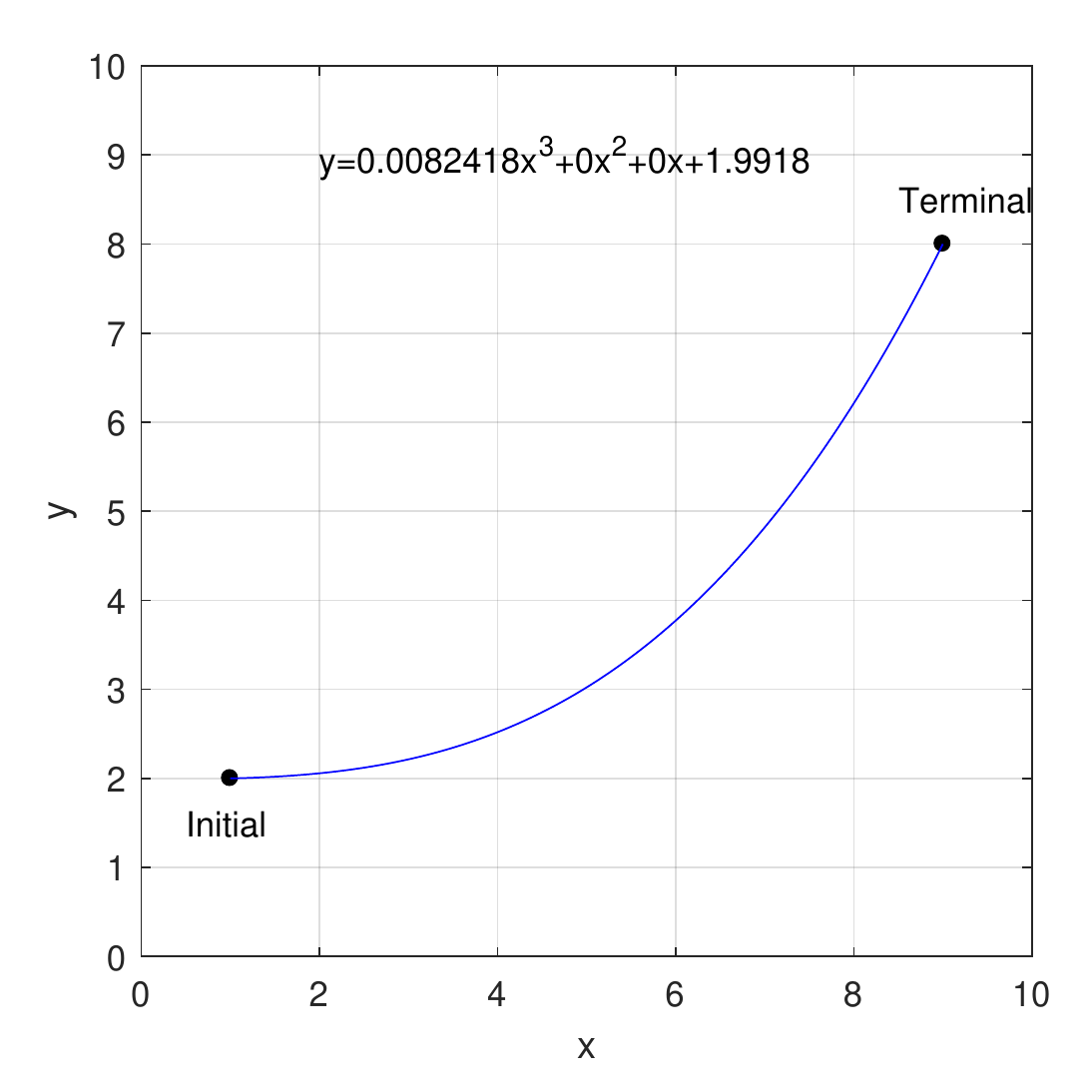}}~\hfil
    \subfigure[B\'ezier path with 3 control points (red circles)]{
    \label{ffig:Bspline}
    \includegraphics[width=0.4\textwidth]{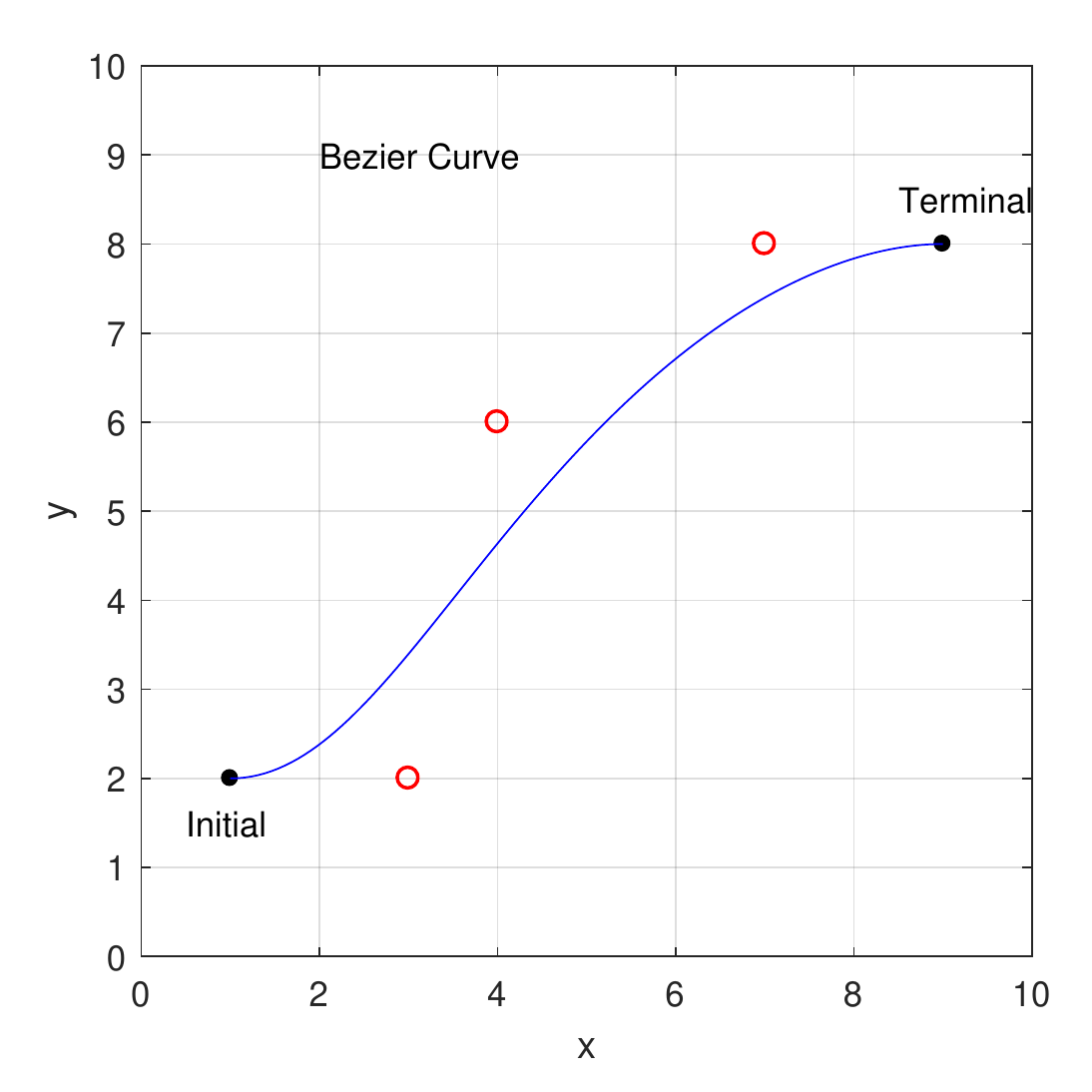}}~\hfil
    \caption{The path between two given points}
    \label{fig:pathTrans}
\end{figure*}

In Fig.~\ref{fig:pathTrans}, different paths in the $x-y$ plane are generated by different approaches. Fig.~\ref{ffig:line} shows a straight line between two points, then paths in Figs.~\ref{ffig:quad} and \ref{ffig:cubi} are obtained by the quadratic and cubic curves, respectively. Mathematically, these curves can be written as polynomial equations of first, second and third degree. In general, the path in translation $\mathbf{T}$ can be expressed as univariate polynomial equations of normalized time $\tau$, that is
\begin{align}\label{path_transPoly}
    \mathbf{T}(\tau) =
    \begin{bmatrix}
        x(\tau)\\y(\tau)\\z(\tau)
    \end{bmatrix}=
    \begin{bmatrix}
        \sum_{i=0}^{k_x}c_{i}\tau^{i}\\
        \sum_{i=0}^{k_y}c_{i}\tau^{i}\\
        \sum_{i=0}^{k_z}c_{i}\tau^{i}
    \end{bmatrix}
    ,~ \tau\in[0,1]
\end{align}
where $k_x,k_y$, $k_z$ are degrees of polynomial equations $x(\tau)$, $y(\tau)$, $z(\tau)$.

In addition, the B\'ezier curve is used to interpolate between initial and terminal poses as shown in Fig.~\ref{ffig:Bspline}, where control points are placed and displayed as red circles. Applying the binomial theorem, the B\'ezier curve of degree $n$ in the polynomial form is given as follows.
\begin{align}\label{path_transBezier}
    &\mathbf{T}(\tau) = 
    \sum^{n}_{k=0}\mathbf{c}_k \tau^k \\
    \mathbf{c}_k= \frac{n!}{(n-k)!}&\sum_{i=0}^{k}\frac{(-1)^{i+k}\mathbf{r}_{P_i}}{i!(k-i)!},~~ \tau\in[0,1]  \nonumber
\end{align}
where $\mathbf{r}_{P_i} \in \mathbb{R}^{3}$ is the location of the $i$-th control point.

\subsection{Path in Orientation}
\label{ssec:ppOrien}
\subsubsection{Rotation Representations}
In order to represent the rotation, several scenarios can be used.
\begin{itemize}
    \item Euler angles. They are used to present a series of rotation along mutually orthogonal axes. Typically, the rotations along $x$, $y$ and $z$ axes in Cartesian space are often called as $x-roll$, $y-roll$ and $z-roll$ and can be expressed as rotation matrices. 
    
    \item Axis-angle. Based on the Euler's rotation theorem, which indicates any rotation or sequence of rotations of a rigid body in the 3D space can be equivalently formulated as a pure rotation about a single fixed axis, the axis-angle representation includes a unit vector axis and a scalar as the rotation angle. 
    
    \item Quaternion. Utilizing three complex numbers $\mathbf{i}, \mathbf{j}, \mathbf{k}$ and a real number $s$, a quaternion can be expressed explicitly as
    \begin{align}
        \mathbf{Q} = s + v_i \mathbf{i} +  v_j \mathbf{j} +  v_k \mathbf{k}
    \end{align}
    where $v_i, v_j, v_k \in \mathbb{R}$ and $\mathbf{i}^2=\mathbf{j}^2=\mathbf{k}^2=\mathbf{i}\mathbf{j}\mathbf{k}=-1$. In addition, the quaternion could be rewritten in the 4-dimension vector form, i.e., 
    \begin{align}\label{quatVector}
        \mathbf{Q} = [s, \mathbf{v}^T]^T = [s, v_i, v_j, v_k]^T 
    \end{align}
    in which $s \in \mathbb{R}$ is the {\it real part} and $\mathbf{v} \in \mathbb{R}^3$ is the {\it vector part}.
    
    \begin{defn} 
        The quaternion $\mathbf{Q}$ is called a unit quaternion if
        \begin{align}
            \|\mathbf{Q}\| = s^2 + \mathbf{v}\cdot \mathbf{v} = 1
        \end{align}
    \end{defn}
\end{itemize}

One problem from the representation of Euler angles is singularity, which is known as the {\it Gimbal Lock}. As for the axis-angle representation, it is not straightforward to combine two sequential rotations as a equivalent axis and angle. However, the quaternion representation overcomes these problems and has desired properties -- no Gimbal lock, ease composition, simple interpolation, etc. Therefore, in order to efficiently generate a smooth path in the orientation, the quaternion representation is used.

Before introducing the interpolation method in quaternions, the relationship between the quaternion and rotation matrix is first recalled.
\begin{theorem}\label{prop_Quaternion2Rotation}
    Given the unit quaternion $\mathbf{Q} = [s, v_i, v_j, v_k]$, then the rotation matrix $R \in \mathbb{R}^{3\times3}$  can be equivalently expressed as 
    \begin{align}\label{eq_Quaternion2Rotation}
        R =
        \begin{bmatrix}
		    1-2(v_j^2+v_k^2)&2(v_iv_j - v_ks)&(v_iv_k + v_js)\\
		    2(v_iv_j + v_ks)&1-2(v_i^2+v_k^2)&2(v_jv_k-v_is)\\
		    2(v_iv_k-v_js)&2(v_jv_k+v_is)&1-2(v_i^2+v_j^2)
	    \end{bmatrix}
    \end{align}
\end{theorem}

\subsubsection{Spherical Linear Quaternion Interpolation ({\it Slerp})}
Spherical linear quaternion interpolation ({\it Slerp}) is one of interpolation functions to generate the shortest path between two given quaternions.

Given unit quaternions $\mathbf{Q}_s$, $\mathbf{Q}_e$ and the scalar $t \in [0,1]$, it has
\begin{align}
    &\cos{\theta} = \mathbf{Q}_s \sbullet \mathbf{Q}_e \label{angleQuat}\\
    & \mathbf{Q} = \frac{\mathbf{Q}_s \sin{[(1-t)\theta]}+\mathbf{Q}_e \sin{(t\theta)}}{\sin{\theta}} := \textit{Slerp}(\mathbf{Q}_s, \mathbf{Q}_e, t) \label{slerpExpress}
\end{align}
Based on the formulations \eqref{quatVector}, the dot product in \eqref{angleQuat} can be defined as
\begin{align}
    \mathbf{Q}_s \sbullet \mathbf{Q}_e = s_1 \cdot s_2 + v_{i1} \cdot v_{i2} + v_{j1} \cdot v_{j2} + v_{k1} \cdot v_{k2}
\end{align}
Here $\theta$ in \eqref{angleQuat} is called as the angle between quaternions $\mathbf{Q}_s$ and $\mathbf{Q}_e$ and $\theta \in [0, \pi]$. Note that $\sin{\theta}$ and $\cos{\theta}$ are constant values once $\mathbf{Q}_s$ and $\mathbf{Q}_e$ are given. 

Note that the following property is useful and the proof process can be found in \cite{mitquaternion}.
\begin{property}\label{slerpUnit}
    The quaternion $\mathbf{Q}$ generated from {\it Slerp} at any $t \in [0, 1]$  is an unit quaternion, that is
    \begin{align}
        \|\mathbf{Q}\| = \| \textit{Slerp}(\mathbf{Q}_s, \mathbf{Q}_e, t)\| = 1,~ t \in [0, 1]
    \end{align}
\end{property}

\subsubsection{Ray-based \textit{Slerp} Expression}
The relationship in \eqref{slerpExpress} can be rewritten as follows,
\begin{align}\label{slerpSinCos}
    \mathbf{Q} &= \frac{1}{\sin{\theta}}\left[\mathbf{Q}_s \sin{(\theta - t \theta}) + \mathbf{Q}_e \sin{(t\theta)}\right]\nonumber\\
      &= \frac{1}{\sin{\theta}}\{\mathbf{Q}_s [\sin{\theta}\cos{(t\theta)} - \cos{\theta}\sin{(t\theta)}] + \mathbf{Q}_e \sin{(t\theta)}\}
\end{align}
Substituting the Weierstrass substitution 
\begin{align}\label{wei_sub}
    T = \tan(\frac{t\theta}{2})~~ \cos{(t\theta)} = \frac{1-T^2}{1+T^2}~~ \sin{(t\theta)}= \frac{2T}{1+T^2}
\end{align}
into  \eqref{slerpSinCos}, then the quaternion $\mathbf{Q}$ from {\it Slerp} \eqref{slerpExpress} is transformed as a fraction where the numerator and denominator terms are univariate polynomial equations with respect to $T$ of degree 2, i.e.,
\begin{align}\label{slerpPoly}
    \mathbf{Q}(T) &= \frac{1}{1+T^2}\left[-\mathbf{Q}_s T^2 + \frac{2}{\sin{\theta}}(\mathbf{Q}_e - \mathbf{Q}_s \cos{\theta})T + \mathbf{Q}_s\right]
\end{align}
where these coefficients are calculated from the given quaternions $\mathbf{Q}_s$ and $\mathbf{Q}_e$ and their angle $\theta$ \eqref{angleQuat}.

It is worth noting that $\theta$ is constant value in \eqref{slerpPoly} while $t$ is actually the {\it unknown variable} in the ray-based method (RBM). Moreover, since $t \in [0,1]$, then from \eqref{wei_sub} the range of $T$ is determined as
\begin{align}\label{rangeOfT}
    T \in [0, \tan(\theta/2)],~ \theta \in [0, \pi]
\end{align}

Considering the formulation \eqref{quatVector}, then \eqref{slerpPoly} can be expanded to 4 univariate polynomial fractions of $T$, that is
\begin{align}\label{slerpPoly4items}
    \begin{split}
        s(T) &= \frac{1}{1+T^2}\left[-s_1 T^2 + \frac{2}{\sin{\theta}}(s_2 - s_1 \cos{\theta})T + s_1\right]\\
        v_i(T) &= \frac{1}{1+T^2}\left[-v_{i1} T^2 + \frac{2}{\sin{\theta}}(v_{i2} - v_{i1} \cos{\theta})T + v_{i1}\right]\\
        v_j(T) &= \frac{1}{1+T^2}\left[-v_{j1} T^2 + \frac{2}{\sin{\theta}}(v_{j2} - v_{j1} \cos{\theta})T + v_{j1}\right]\\
        v_i(T) &= \frac{1}{1+T^2}\left[-v_{k1} T^2 + \frac{2}{\sin{\theta}}(v_{k2} - v_{k1} \cos{\theta})T + v_{k1}\right]
    \end{split}
\end{align}
Furthermore, from the Property~\ref{slerpUnit}, it implies that $\mathbf{Q}(T)$ is an unit quaternion for any $T \in [0, \tan(\theta/2)]$.
As such based on the Theorem~\ref{prop_Quaternion2Rotation}, substituting \eqref{slerpPoly4items} into \eqref{eq_Quaternion2Rotation} yields the equivalent rotation matrix from $\mathbf{Q}(T)$ in the polynomial form
\begin{align}\label{rotationMatrix_T}
    R(T) \in 
    \frac{1}{(1+T^2)^2}
    \begin{bmatrix}
        O(T^4)& O(T^4)& O(T^4)\\
        O(T^4)& O(T^4)& O(T^4)\\
        O(T^4)& O(T^4)& O(T^4)\\
    \end{bmatrix}
\end{align}

\subsection{Ray-based Path}
\label{sec:RBMpp}
The generalized coordinates of the CDPR can be defined as $\mathbf{q} =[\mathbf{T}^T, \mathbf{Q}^T]^T= [x,y,z,s,v_i,v_j,v_k]^T$, where $\mathbf{T} = [x,y,z]^T$ are translation DoFs while $\mathbf{Q} = [s,v_i,v_j,v_k]^T$ is the quaternion presenting the orientation of the CDPR. 

The initial and terminal poses, i.e., $\mathbf{q}_s=[\mathbf{T}_s^T,\mathbf{Q}_s^T]^T$ and $\mathbf{q}_e=[\mathbf{T}_e^T,\mathbf{Q}_e^T]^T$ are given in advance. Then the path in orientation space from $\mathbf{Q}_s$ to $\mathbf{Q}_e$, i.e., $\mathbf{Q}(T)=[s(T),v_i(T),v_j(T),v_k(T)]$ can be generated by the ray-based \textit{Slerp} \eqref{slerpPoly}. 
Regarding the translational coordinates, the relationship between $\tau$ and $T$ can be arbitrarily defined. For simplicity, the following linear relationship is proposed
\begin{align}\label{relation_tT}
    \tau = \frac{T}{\tan(\theta/2)}
\end{align}
Substituting \eqref{relation_tT} into \eqref{path_transPoly}, the trajectory from $\mathbf{T}_s$ to $\mathbf{T}_e$
could be determined as 
\begin{align}\label{rayBasedTrans}
    \mathbf{T}(T) \in
       [ O(T^{k_x}), O(T^{k_y}), O(T^{k_z})]^T
\end{align}

In summary, the 7-dimension path in translation and orientation spaces determined by the ray-based method could be formulated as
\begin{align}\label{rayBasedPath}
    \mathbf{P}(T) =
    \begin{bmatrix}
        \mathbf{T}(T)\\\mathbf{Q}(T)
    \end{bmatrix}
    \in
    \mathbb{R}^7
    ,~T \in [0, \tan(\theta/2)]
\end{align}

\section{Ray-based Path Verification}
\label{sec:RBMpathCheck}
In this subsection, the cable interference free conditions are used to verify the feasibility of the proposed path from \eqref{rayBasedPath}. 

First of all, based on the kinematics of the CDPR in Fig.~\ref{mod_cdpr}, the $i$-th cable segment can be described as follows
\begin{align}
    \mathbf{s}_i = {}^0\mathbf{r}_{P_0P_1}+{}^0R_1 {}^1\mathbf{r}_{P_1A_{ie}}-{}^0\mathbf{r}_{P_0A_{is}}
\end{align}
Substituting the proposed path \eqref{rayBasedPath} and \eqref{rotationMatrix_T} into above equation, then yields the ray-based cable segment equations with respect to $T$ as follows
\begin{align}\label{s_i_T}
    \mathbf{s}_i(T)
    =
    \mathbf{T}(T)+
    R(T)
    {}^1\mathbf{r}_{P_1A_{ie}}-{}^0\mathbf{r}_{P_0A_{is}}
\end{align}
where ${}^1\mathbf{r}_{P_1A_{ie}}\in \mathbb{R}^3$ and ${}^0\mathbf{r}_{P_0A_{is}} \in \mathbb{R}^3$ are geometrical parameters $\mathcal{G}$ (Property \ref{P1}), thereby they are both constant vectors.

From \eqref{rayBasedTrans} and \eqref{rotationMatrix_T}, \eqref{s_i_T} implies the following theorem similar to Theorem \ref{them1}.
\begin{theorem}
    $\mathbf{s}_i$ could be written as a fraction where the numerator and denominator terms are univariate polynomial equations of $T$, i.e.,
    \begin{align}\label{rayBasedSeg_PP}
        \mathbf{s}_i \in \frac{1}{(1+T^2)^2}
        \begin{bmatrix}
            O(T^{k_x+4})\\O(T^{k_y+4})\\O(T^{k_z+4})
        \end{bmatrix}
        ~\text{where}~T = \tan(t\theta)/{2}
    \end{align}
\end{theorem}

Substituting the ray-based cable segment equation \eqref{rayBasedSeg_PP} into ray-based interference conditions in Sec. \ref{sec:cable_cable_IFW}, the proposed path $\mathbf{P}(T)$ is verified under the interference free conditions for cable-cable.

For simplicity, let's assume that $k_x = k_y = k_z = k \neq 0 $, by which means that the polynomial equations of translational DoFs have the same degree. As such, given two cables $\mathbf{s}_i = [s_{ix}, s_{iy}, s_{iz}]^T$ and $\mathbf{s}_j=[s_{jx}, s_{jy}, s_{jz}]^T$, from \eqref{rayBasedSeg_PP} it has
\begin{align}\label{rayBaseds_is_j_T}
    \mathbf{s}_i, \mathbf{s}_j \in 
    \frac{1}{(1+T^2)^2}
    \begin{bmatrix}
        O(T^{k+4})\\O(T^{k+4})\\O(T^{k+4})
    \end{bmatrix}
\end{align}
then
\begin{align}\label{rayBasedCrossProd_sisj}
    \mathbf{s}_i \times \mathbf{s}_j = 
    \begin{bmatrix}
        s_{iy}s_{jz} - s_{iz}s_{jy}\\
        s_{iz}s_{jx} - s_{ix}s_{jz}\\
        s_{ix}s_{jy} - s_{iy}s_{jx}\\
    \end{bmatrix}
    \in
    \frac{1}{(1+T^2)^4}
    \begin{bmatrix}
        O(T^{2k+8})\\O(T^{2k+8})\\O(T^{2k+8})
    \end{bmatrix}
\end{align}
By the definition of $\mathbf{s}_{ij} = {}^0\mathbf{r}_{A_{is}A_{js}}$ in ~\eqref{eq:s_ij}, it could find that $\mathbf{s}_{ij}$ is a constant vector for the CDPR as shown in Fig.~\ref{mod_cdpr}, that is
\begin{align}
    \mathbf{s}_{ij}
    \in 
    [O(1), O(1), O(1)]^T
    \end{align}
From the \eqref{4items}, yields
\begin{align}\label{n_titjtd_Part1}
    n_{t_i} &=
    \det{[\mathbf{s}_{ij}, -\textbf{s}_j, -\textbf{s}_i \times \textbf{s}_j]}\nonumber \\
    &=
    \det\left(
    \begin{bmatrix}
         O(1)&  \frac{O(T^{k+4})}{(1+T^2)^2} &\frac{O(T^{2k+8})}{(1+T^2)^4}\\
         O(1)&  \frac{O(T^{k+4})}{(1+T^2)^2} &\frac{O(T^{2k+8})}{(1+T^2)^4}\\
         O(1)&  \frac{O(T^{k+4})}{(1+T^2)^2} &\frac{O(T^{2k+8})}{(1+T^2)^4}
    \end{bmatrix}
    \right)
    \in
    \frac{O(T^{3k+12})}{(1+T^2)^6}
\end{align}
If defining $n_{t_i} = \tilde{n}_{t_i}/(1+T^2)^6$, thus 
\begin{align}\label{tilde_n_titjtd_Part1}
    \tilde{n}_{t_i} \in O(T^{3k+12})
\end{align}
Similarly, it obtains
\begin{align}\label{n_titjtd_Part2}
    n_{t_j} \in \frac{O(T^{3k+12})}{(1+T^2)^6},~n_t\in \frac{O(T^{2k+8})}{(1+T^2)^4},~d\in\frac{O(T^{4k+16})}{(1+T^2)^8}
\end{align}
Furthermore,
\begin{align}\label{tilde_n_titjtd_Part2}
    \tilde{n}_{t_j} \in O(T^{3k+12}),~\tilde{n}_t\in O(T^{2k+8}),~\tilde{d} \in O(T^{4k+16})
\end{align}

The interference between non-parallel cables is first considered. So substituting \eqref{n_titjtd_Part1} and \eqref{n_titjtd_Part2} into the interference conditions \eqref{cond_seg_seg}, yields 
\begin{align}\label{interferenceInequ}
    (1+T^2)^2 &\tilde{n}_{t_i} \geqslant 0, ~\tilde{d}-(1+T^2)^2 \tilde{n}_{t_i} \geqslant 0, \nonumber\\
    (1+T^2)^2 &\tilde{n}_{t_j} \geqslant 0, ~\tilde{d}-(1+T^2)^2 \tilde{n}_{t_j} \geqslant 0,
    ~\epsilon_r^2\tilde{d}-\tilde{n}_t^2 \geqslant 0
\end{align}
Taking \eqref{tilde_n_titjtd_Part1} and \eqref{tilde_n_titjtd_Part2} into above equations, then they can be transferred to a set of univariate polynomial equations with respect to $T$, that is, if define
$
    \mathcal{L}_{np} = \{(1+T^2)^2 \tilde{n}_{t_i}, \tilde{d}-(1+T^2)^2 \tilde{n}_{t_i}, (1+T^2)^2 \tilde{n}_{t_j}, \tilde{d}-(1+T^2)^2 \tilde{n}_{t_j}, \epsilon_r^2\tilde{d}-\tilde{n}_t^2\}
$, then
$$
    \varrho_i \in O(T^{\eta_i}),~\forall \varrho_i \in \mathcal{L}_{np}
$$
where $\eta_i$ is the degree of each polynomial equation.
Similar to \eqref{w_ij_non}, the interference parts of the proposed path $\mathbf{P}(T)$ for non-parallel cable segments are formulated by
\begin{align}
    P^{ij}_{np} = \{T: \tilde{d}(T)>0, \varrho_i(T) \geqslant 0, \forall \varrho_i(T) \in \mathcal{L}_{np}\} \nonumber
\end{align}

Then for the case of parallel case, substituting $\mathbf{s}_i$ \eqref{rayBaseds_is_j_T} and $\mathbf{s}_{ij}$  \eqref{rayBasedCrossProd_sisj} into the corresponding interference conditions \eqref{parallel_cond}, yields an univariate polynomial inequality, i.e., $\varrho(T) =\epsilon_r^2\|\textbf{s}_i\|^2- \|\textbf{s}_i \times \mathbf{s}_{ij}\|^2$.
As such, the interference parts of the proposed path $\mathbf{P}(T)$ for parallel cable segments are given as follows,
\begin{align}
     P^{ij}_{p} = \{T: \tilde{d}(T)=0, \varrho(T)  \geqslant 0\} 
\end{align}

The resulting interference parts along the path $\mathbf{P}(T)$ for all $m$ cable segments are 
\begin{align}\label{W_set}
    P_c = \bigcup_{i=2}^{m}\bigcup_{j=1}^{i-1}(P^{ij}_{np}\cup P^{ij}_{p}) 
\end{align}

So from \eqref{wei_sub}, the interference free parts along the proposed trajectory can be defined as
\begin{align}
    P_{IFC} = \{t: t = 2tan(T)^{-1}/\theta, T \notin  P_{c}, t \in [0, 1]\} \nonumber
\end{align}

\section{Simulation and Results}

\begin{figure*}[htbp]
    \centering
    \subfigure[Linear path]{
    \label{ffig:reslutingLinear}
    \includegraphics[width=0.45\textwidth]{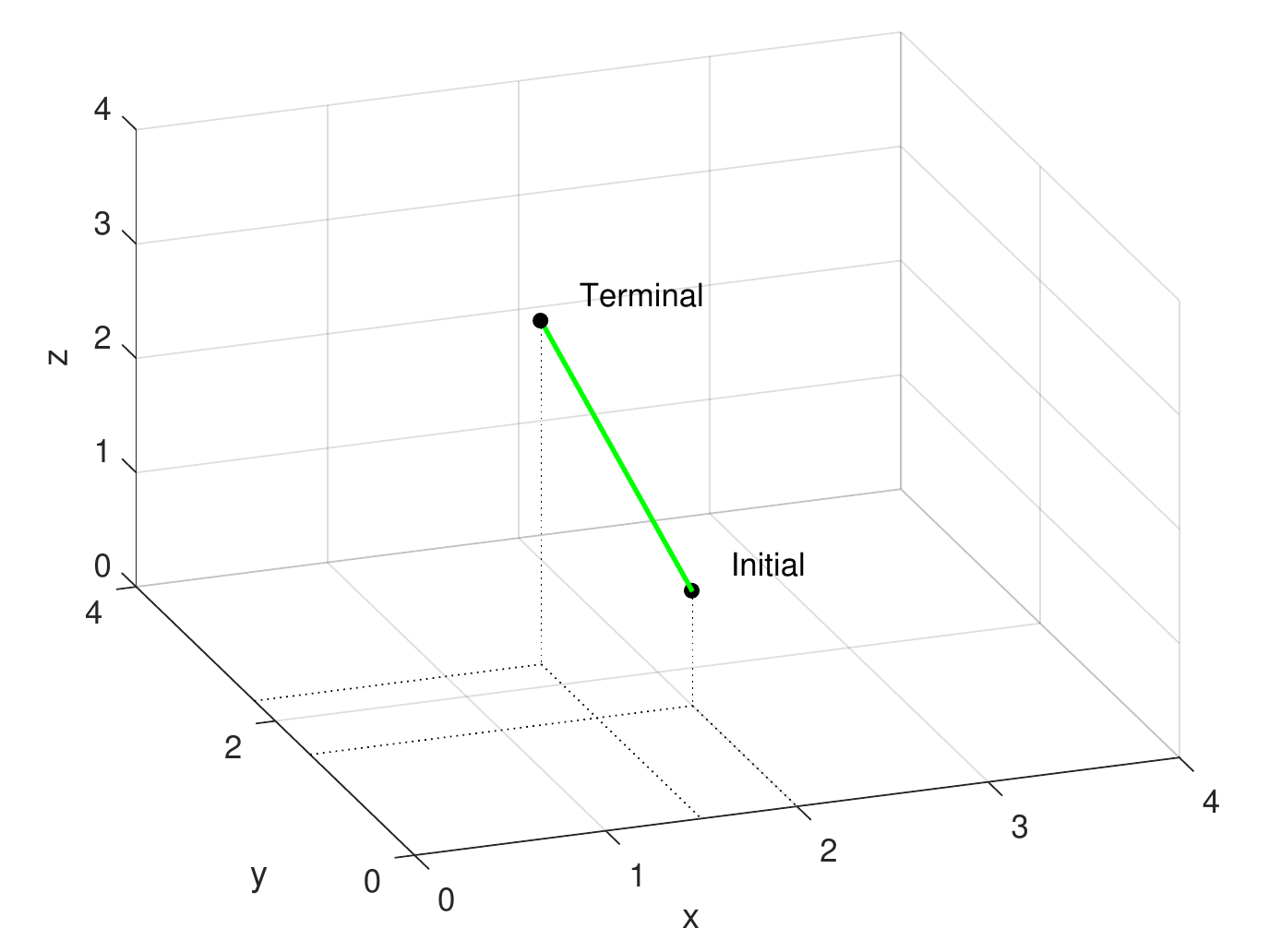}}~\hfil
	\subfigure[Quadratic polynomial path, where the point $(1.2, 1.9, 1.8)$ is given to be passed through]{
    \label{ffig:resultingQuad}
    \includegraphics[width=0.45\textwidth]{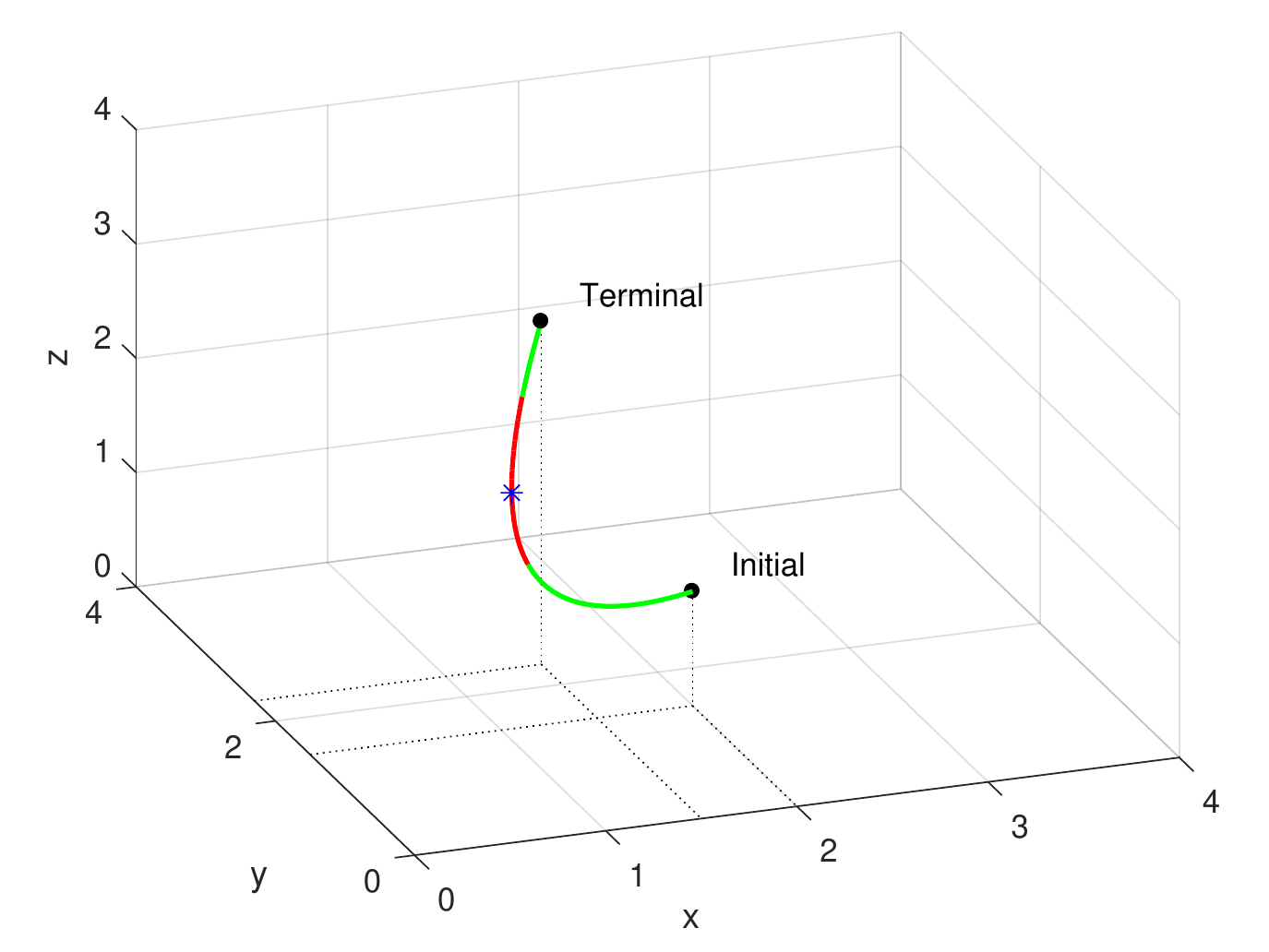}}
    \caption{The resulting trajectories, where the green parts are feasible trajectories while the red parts means infeasible trajectories.}
    \label{fig:results}
\end{figure*}

\begin{figure*}[htbp]
    \centering
    \subfigure[A set of points inserted in the vertical plane]{
    \label{ffig:quadVer}
    \includegraphics[width=0.45\textwidth]{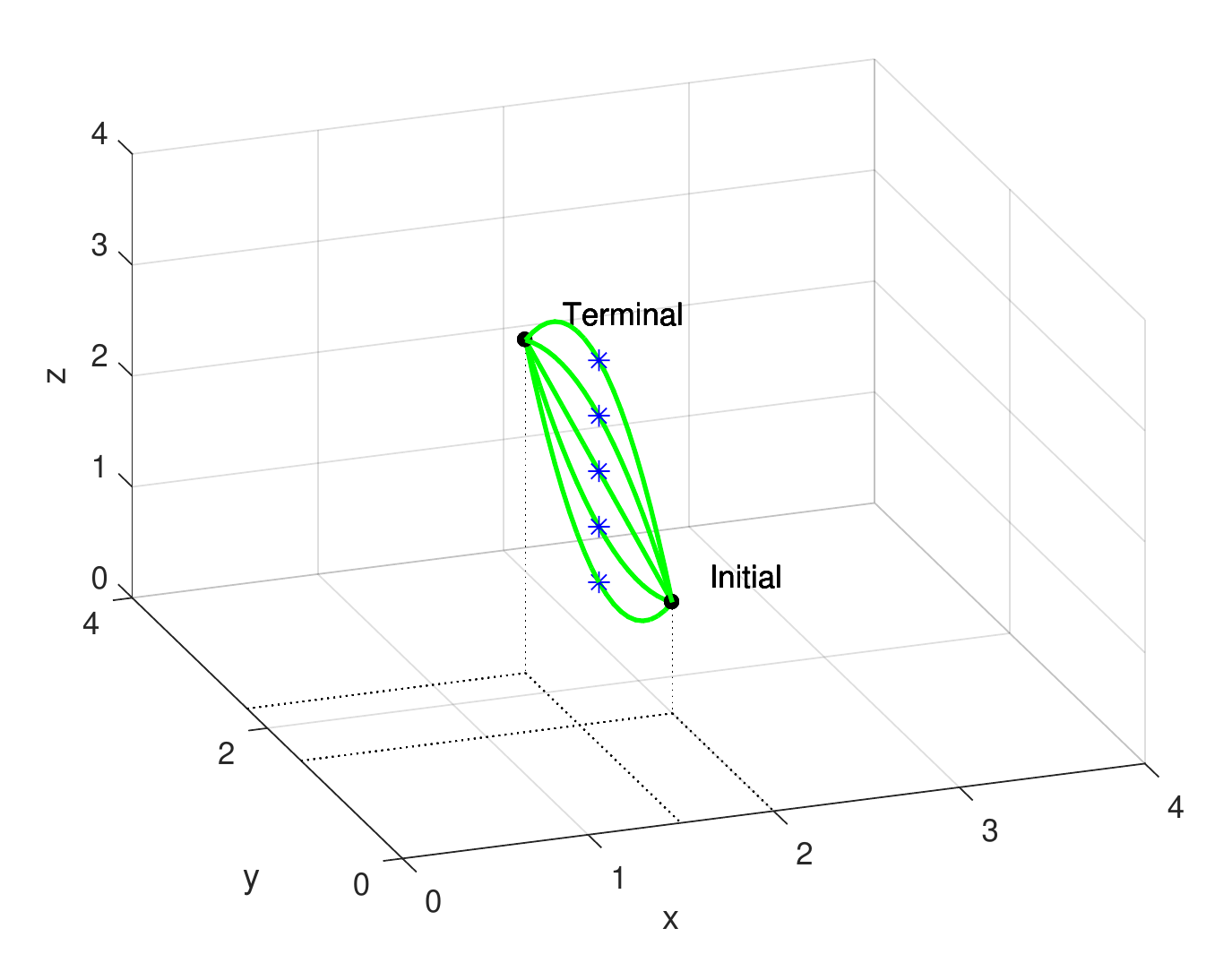}}~\hfil
	\subfigure[A set of points inserted in the horizontal plane]{
    \label{ffig:quadHor}
    \includegraphics[width=0.45\textwidth]{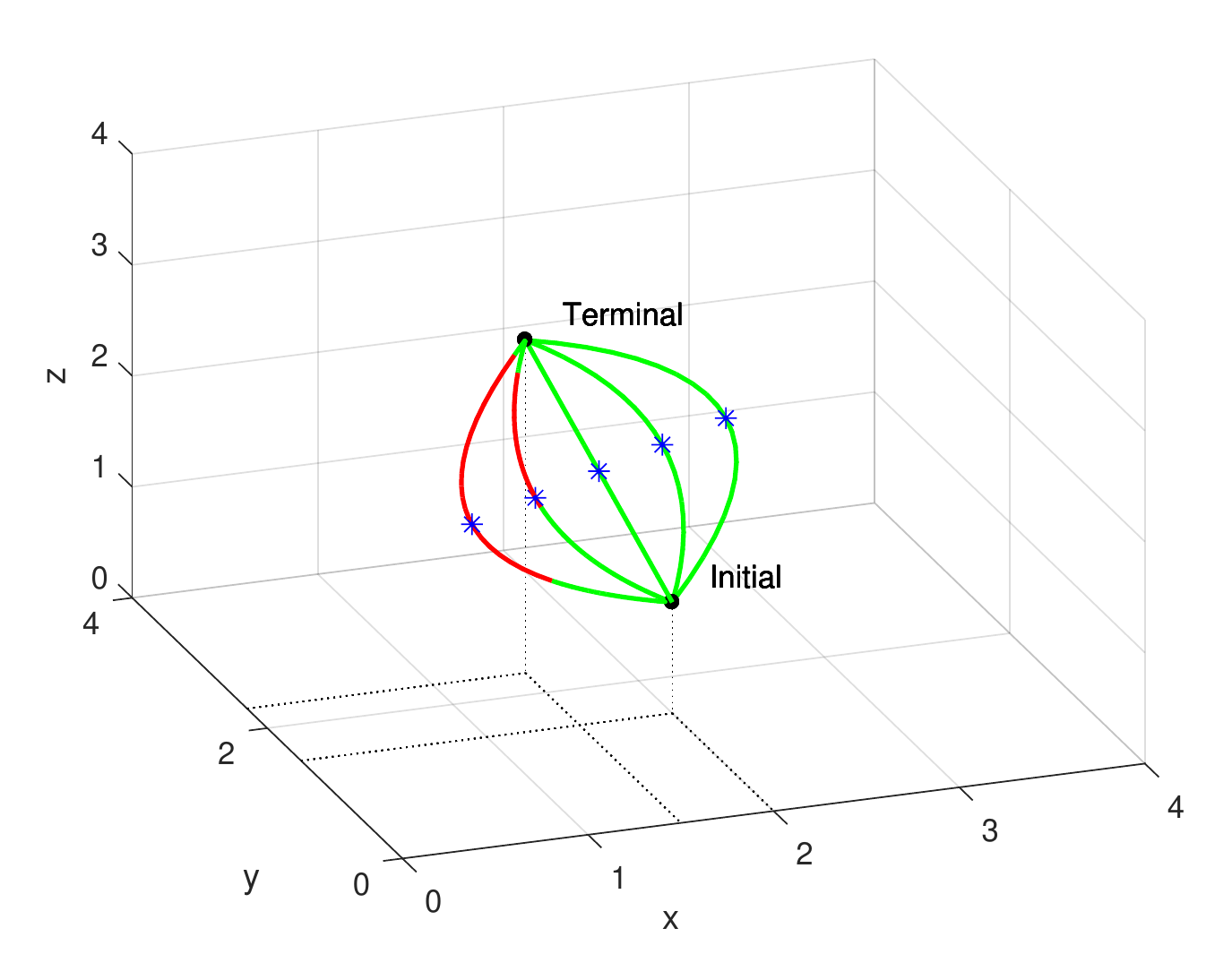}}
    \caption{Multiple quadratic trajectories}
    \label{fig:results2}
\end{figure*}

\begin{figure*}
    \centering
    \subfigure[3D view]{
    \label{ffig:sphere_3d}
    \includegraphics[width=0.7\textwidth]{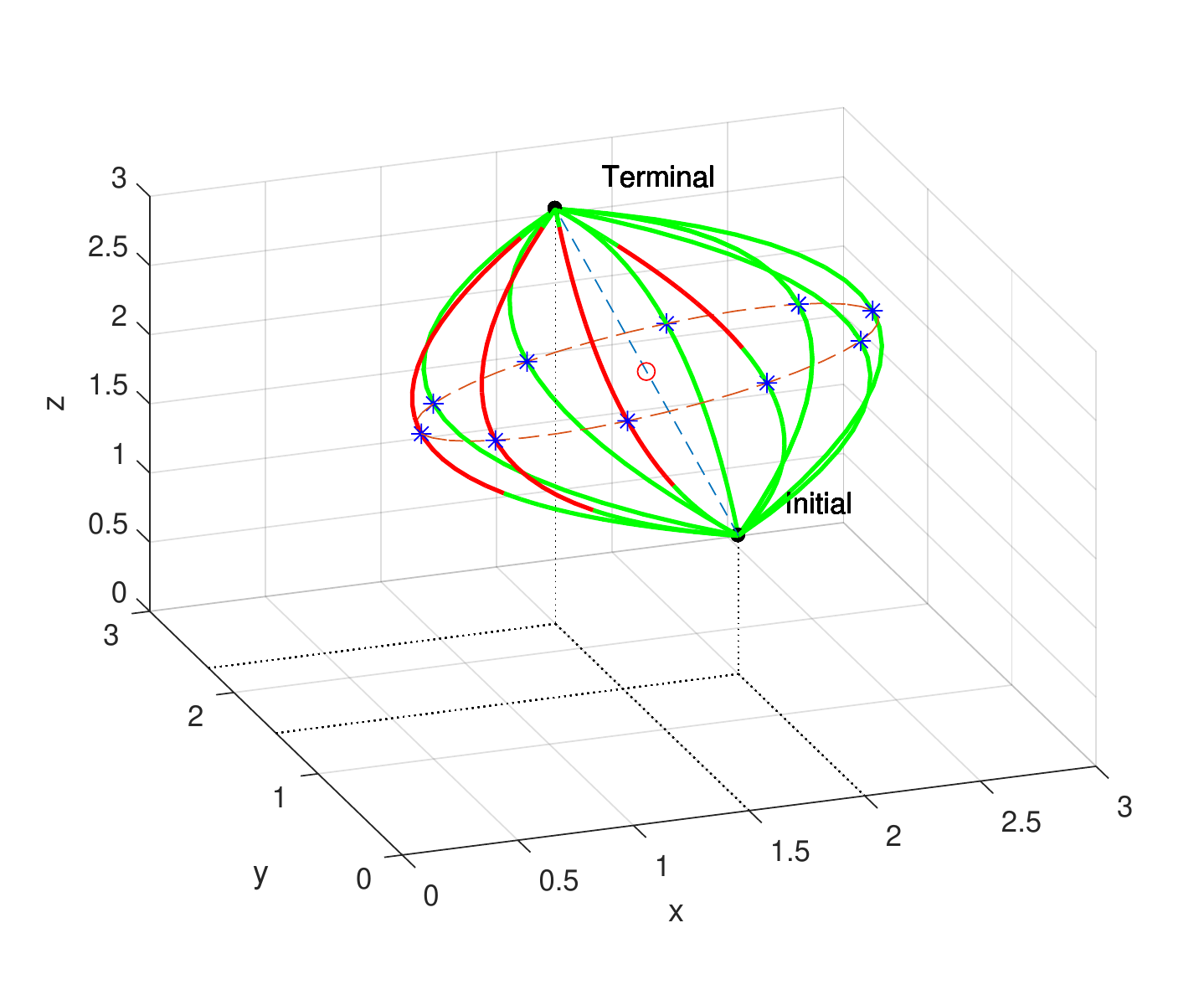}}~\vfil
	\subfigure[Top view $(x-y)$]{
    \label{ffig:sphere_xy}
    \includegraphics[width=0.5\textwidth]{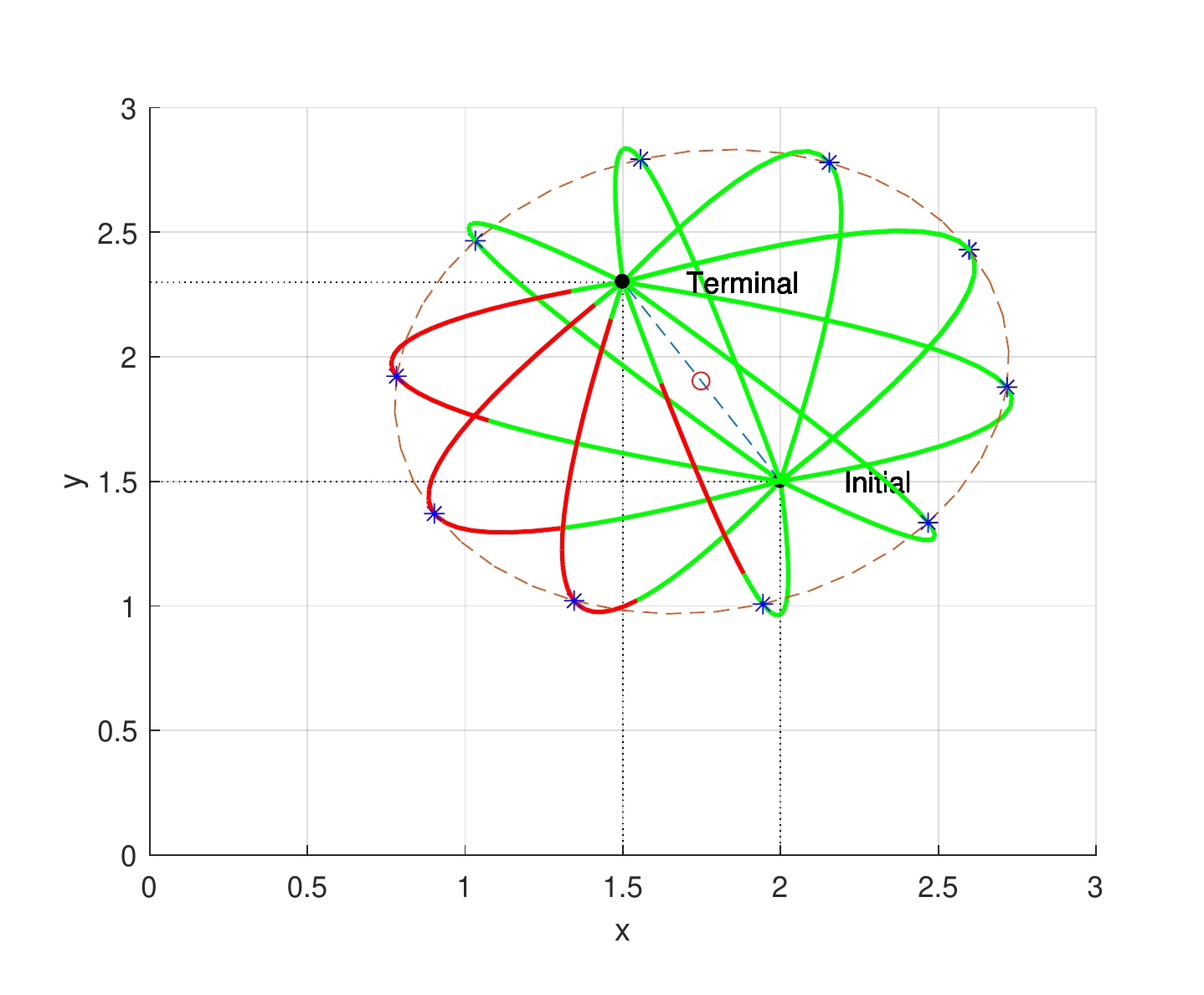}}~\hfil
    \subfigure[Side view $(x-z)$]{
    \label{ffig:sphere_xz}
    \includegraphics[width=0.5\textwidth]{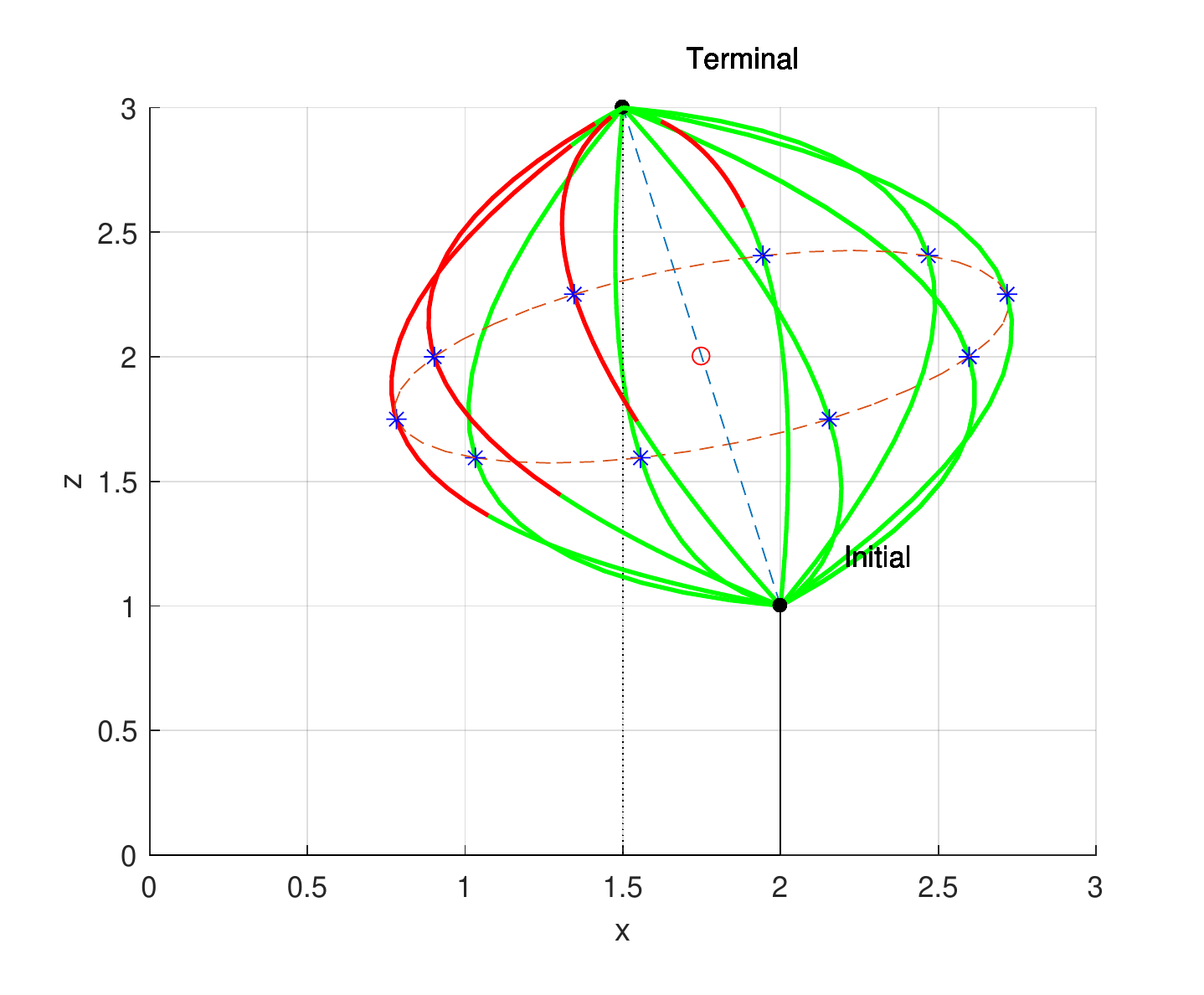}}
    \caption{A family of quadratic trajectories}
    \label{fig:sphere}
\end{figure*}

In this section, a bunch of trajectories, including linear, quadratic and some complex trajectories for the CDPR are verified by the proposed method. The results are shown from Fig.~\ref{fig:results} to Fig.~\ref{fig:sphere}, where the green and red sections represent the feasible and infeasible parts of the trajectories, respectively. From the simulation results, it is shown that the proposed approach can treat the trajectories with varying translation and orientation simultaneously.

The configuration of the CDPR is given in Table~\ref{case1_CDPR}.The starting and terminal positions are $[2, 1.5, 1]$ and $[1.5, 2.3, 3]$ while the orientations are $[0, 0, 30]$ and $[0, 0, 0]$ (in $deg$). Besides, the safe distance between cables is set to be $0.1m$.

First of all, the linear translational trajectory is verified. As such, from \eqref{path_transPoly} the translation trajectory can be rewritten in the linear polynomial form as
\begin{align}\label{case_linear_trans_tau}
    \mathbf{T}(\tau) = 
    \begin{bmatrix}
        x(\tau)\\y(\tau)\\z(\tau)
    \end{bmatrix}=
    \begin{bmatrix}
        -0.5\tau+2\\
        0.8\tau+1.5\\
        2\tau+1
    \end{bmatrix}
    ,~ \tau\in[0,1]
\end{align}

After converting the Euler angles into the unit quaternion form, i.e., $\mathbf{Q}_s=[0.9659, 0, 0, 0.2588]$ and $\mathbf{Q}_e=[1, 0, 0, 0]$, then the orientation trajectory is determined by \eqref{slerpExpress} as
\begin{align}\label{case_linear_orie_t}
    \mathbf{Q}(t) &= slerp([0.9659, 0, 0, 0.2588],[1, 0, 0, 0],t), ~(t \in [0, 1])
\end{align}
Substituting \eqref{relation_tT}
\begin{align}
    \tau = \frac{T}{0.1317}
\end{align}
into \eqref{case_linear_trans_tau}, it yields
\begin{align}\label{case_linear_trans_T}
    \mathbf{T}(T) = 
    \begin{bmatrix}
        -3.7979 T+2\\
        6.0766T+1.5\\
        15.1915+1   
    \end{bmatrix}
\end{align}
Using the Weierstrass substitution \eqref{wei_sub}, the \eqref{case_linear_orie_t} can be converted as
\begin{align}\label{case_linear_orie_T}
    \mathbf{Q}(T) = 
    \frac{1}{1+T^2}
    \begin{bmatrix}
        -0.9659 T^2 + 0.5176 T + 0.9659\\
        0\\
        0\\
        -0.2588 T^2 + -1.9319 T + 0.2588
    \end{bmatrix}
\end{align}

Then the equivalent rotation matrix from Q(T) \eqref{case_linear_orie_T} can be transformed in the polynomial form \eqref{rotationMatrix_T} as
\begin{align}\label{case_linear_rotationMat}
    R(T)= \frac{1}{(1+T^2)^2}
    \begin{bmatrix}
    r_{11}& r_{12}& r_{13}\\
    r_{21}& r_{22}& r_{23}\\
    r_{31}& r_{32}& r_{33}\\
    \end{bmatrix}
\end{align}
where 
\begin{align}
    \begin{split}
        r_{11} &= 0.8660T^4	-2T^3	-5.1962T^2	+2T	+0.8660;\\
        r_{12} &= -0.5T^4   -3.4641T^3   +3T^2    +3.4641T  -0.5;\\
        r_{13} &= 0;\\
        r_{21} &= 0.5T^4    +3.4641T^3   -3.T^2  -3.4641T    +0.5;\\
        r_{22} &= 0.8660T^4   -2.0000T^3   -5.1962T^2   +2T+  0.8660\\
        r_{23} &= 0;\\
        r_{31} &= 0;\\
        r_{32} &= 0;\\
        r_{33} &= T^4 + 2T^2 + 1;\\
    \end{split}
\end{align}

Furthermore, substituting \eqref{case_linear_trans_T} and \eqref{case_linear_rotationMat} into \eqref{s_i_T}, the first cable segment $\mathbf{s}_i$ can be described as follows
\begin{align}
    \begin{split}
        &\mathbf{s}_1 =\\
        &\begin{bmatrix}
            -3.7979T^5    +1.9201T^4   -6.9493T^3    +4.4794T^2   -4.4443T    +1.9201;\\
            6.0766T^5    +0.3384T^4   +11.8336T^3    +1.9696T^2   +6.3962T    +0.3384;\\
            15.1915T^5   +1.3000T^4   +30.3830T^3    +2.6000T^2   +15.1915T   +1.3000;\\
        \end{bmatrix}
    \end{split}
\end{align}

Similarly, other cable segments can be yielded in the form of polynomial equations of $T$ as well. 

Taking cable 1 and cable 2 as an example, it is shown that $\tilde{n}_{t_1}$, $\tilde{n}_{t_2}$, $\tilde{n}_{t}$ and $d$ from \eqref{tilde_n_titjtd_Part1} and \eqref{tilde_n_titjtd_Part2} can be obtained as
\begin{align}
    \begin{split}
        \tilde{n}_{t_1} &= 900T^{14}+319+T^{13}+\cdots+83T+19;\\
        \tilde{n}_{t_2} &= 900T^{14}+317+T^{13}+\cdots+76T+20;\\
        \tilde{n}_{t} &= 3.0383T^{9}+21.3100T^{8}+\cdots+1.2370T+0.2600;\\
        d &= 829T^{18}+535T^{17}+\cdots+65T+18;
    \end{split}
\end{align}

Finally, by solving a set of polynomial inequalities \eqref{interferenceInequ}, it is found that the solution is $\{T: [0, 0.1317]\}$. Converting $T$ to $t$, it is shown that the valid part along this linear trajectory is $P_{IFC} =\{t: [0, 1]\}$, which means the whole trajectory is feasible for this CDPR, as shown in Fig.~\ref{ffig:reslutingLinear}.


Moreover, a quadratic trajectory is verified for the CDPR with same configuration in Table~\ref{case1_CDPR}. The parameters of the trajectory is given as follows,
\begin{align}\label{case_quad_traj}
    \begin{split}
        x(\tau) &= 126.9301 \tau^2  -20.5085\tau+    2.0000;\\
        y(\tau) &= 6.0766\tau+    1.5000;\\
        z(\tau) &= 46.1564\tau^2    +9.1149\tau+    1.0000; \\
        \mathbf{Q}(t) &= slerp([0, 0, 30],[0, 0, 0],t) 
    \end{split}
\end{align}
where $\tau \in [0, 1]$ and $t\in[0,1]$. Then some variables in the polynomial form are listed below.
\begin{align}
    \begin{split}
        \tau &= T/0.1317;\nonumber\\
        \mathbf{s}_1 &=
        \begin{bmatrix}
           126.9301T^6+  \cdots  -21.1549T+ 1.9201;\\
            6.0766T^5+       \cdots    +6.3962T+  0.3384;\\
            46.1564T^6+      \cdots    +9.1149T+ 1.3000 ;\\
        \end{bmatrix}\nonumber\\
        \tilde{n}_{t_1} &= 6.6648\times10^4T^{16} + \cdots -201.8771T+18.8198;\\ 
    \end{split}
\end{align}

Then  the feasible parts along the trajectory are solved, that is $P_{IFC} = \{t: [0, 0.5008] \cup [0.8852, 1]\}$ as shown in Fig.~\ref{ffig:resultingQuad}. It implies that the robot can not move from the initial pose to the terminal pose along the trajectory in \eqref{case_quad_traj}. Furthermore, the invalid trajectory can be pointed out explicitly from the simulation results, that is $0.5008 < t < 0.8852$. Here $t$ is the normalised parameter.

In addition, a set of quadratic trajectories in the vertical plane between two end points are treated by the proposed method. it is shown that all given paths are feasible in Fig.~\ref{ffig:quadVer}. That means the end-effector can move along these green paths without any collisions between its cables. In Fig.~\ref{ffig:quadHor}, several quadratic trajectories are selected in one plane between the initial and terminal points. Then  there exist some infeasible trajectories that can not satisfy the interference conditions, as shown in Fig.~\ref{ffig:quadHor},

Finally, a bunch of quadratic paths are verified, where a set of points lie on a circle (red dash curve) whose center is the middle point between the initial and terminal points (small red circle) and radius is $1 m$. As shown in Fig.~\ref{sphere}, it can be shown that the end-effector of the CDPR has a variety of choices to move between two end points.

The comparison between the proposed method and point-wise method is demonstrated in the Table.~\ref{table:trajVeri_ray_point}, in terms of the result accuracy and time efficiency. It is shown that the result accuracy of the proposed method is higher than point-wise method with $\Delta t = 0.01$, while the time consumption is great less than it. Furthermore, the high numerical accuracy of the proposed method results from solving the polynomial inequalities. Although the time cost of verification for one isolated point is low, it would take more time with the finer discretization step $\Delta$ t. However, since the proposed method checks the trajectory by solving the polynomial inequalities, it can yield the result directly and does not suffer from the discretization step.

\begin{table}[ht]
    \caption{Comparison for the accuracy and time efficiency}
    \centering
    \begin{tabular}{ |c|c|c|c| } 
     \hline
     \multirow{2}{4em}{Method} & Discretization step & 1st Feasible range & Time \\
      & $\Delta$t & t & (sec)  \\ 
     \hline
     \multirow{3}{4em}{Point-wise method} & 0.1 & [0,0.5] & 0.49 \\ 
     & 0.05 & [0,0.50] & 1.42  \\ 
     & 0.01 & [0,0.50] & 4.89  \\ 
     \hline
     Proposed method & N/A & $[0, 0.5008]$ & 0.60  \\ 
     \hline
    \end{tabular}
    \label{table:trajVeri_ray_point}
\end{table}


\chapterend
\pdfoutput=1
%

\chapter{Conclusion and Future Works}

This thesis presents a generalized approach to IFW analysis with cable-cable and cable-object for arbitrary CDRs. The ray-based IFW could be generated for CDPRs and MCDRs with all of DoFs and it is shown that each ray in the workspace is efficiently determined by solving a set of univariate polynomial inequalities resulting from the interference conditions. Additionally, the ray-based IFW provides the continuity information and hence the global understanding of the whole workspace. At the same time, interference between cable segments and different shaped objects could be detected by the proposed method as well while preserving the nature of polynomial equations. Thus, objects could be approximated by polyhedra with triangular faces and treated as a combination with cylinders, ellipsoids and cones. 

Furthermore, a novel method on path planning and verification under IFC and WCC for CDRs is proposed. The path of the robot could be converted into univariate polynomial equations in the quaternion representation as well. Then the feasible range of the trajectory could be found out under IFC and WCC separately. This study provides a new approach to determine the feasible trajectory of the robot compared with the traditional point-wise searching. Cable interference could be occurred by this simulation and then avoided in the operation.

The future works may focus on following aspects:
\begin{itemize}
    \item the study of a wider range of obstacles that can be solved using the ray-based method;
    \item the scheme of path planning and re-planning problem by using the workspace represented by continuous interference free paths.
\end{itemize}

\chapterend




\newpage
\addcontentsline{toc}{chapter}{Bibliography}
\bibliographystyle{ieee}
\bibliography{database}

\end{document}